%% file: iclr2027_conference_arxiv.tex
\documentclass{article} % For LaTeX2e
\usepackage{iclr2027_conference,times}

\input{math_commands.tex}

\usepackage{hyperref}
\usepackage{url}

\usepackage{booktabs}       % professional-quality tables
\usepackage{amsfonts}       % blackboard math symbols
\usepackage{nicefrac}       % compact symbols for 1/2, etc.
\usepackage{microtype}      % microtypography
\usepackage{xcolor}         % colors

\usepackage{xspace}

\usepackage{multirow}
\usepackage{tabularx}

\usepackage{amsmath}
\usepackage{amssymb}

\usepackage{subcaption}

\usepackage{wrapfig}

\usepackage{enumitem}

\usepackage[most]{tcolorbox}

\definecolor{bgcolor}{HTML}{F6F8FA}
\usepackage{float}
\newfloat{listing}{htbp}{lol}
\floatname{listing}{Listing}
\usepackage[capitalize]{cleveref}
\Crefname{equation}{}{}
\Crefname{appendix}{App.}{Apps.}
\Crefname{theorem}{Thm.}{Thms.}
\Crefname{section}{Sec.}{Secs.}

\usepackage{resizegather}

\newcommand{\bolt}{\textsc{BoLT}\xspace}

\title{\bolt: A Benchmark to Democratize Black-box Optimization Research for Expensive LLM Tasks}

\author{Ruth Wan Theng Chew\thanks{Work done while at NUS.} \\
  University of Washington \\
  \texttt{ruthchew@cs.washington.edu}
  \And
  Zhiliang Chen \\
  National University of Singapore \\
  \texttt{chenzhiliang@u.nus.edu}
  \And
  Apivich Hemachandra \\
  Singapore-MIT Alliance for Research and Technology\\
  \texttt{apivich.hemachandra@smart.mit.edu}
  \And
  Bryan Kian Hsiang Low \\
  National University of Singapore \\
  \texttt{dcslkh@nus.edu.sg}
}

\iclrfinalcopy % Uncomment for camera-ready version, but NOT for submission.
\begin{document}

\setcounter{footnote}{1}  % to use dagger for author footnote
\maketitle
\lhead{Preprint.}

\begin{abstract}
Optimization of LLM training and inference configurations, such as hyperparameters, data mixtures, prompts, and parallelism strategies, is critical to performance. However, it is often done heuristically, which can lead to suboptimal outcomes. These tasks are \emph{noisy}, \emph{expensive}, and \emph{derivative-free}, making them a natural fit for sample-efficient Bayesian optimization (BO) and other black-box optimization (BBO) methods.  This direction remains underexplored, largely because LLM training and inference are prohibitively expensive for most BBO researchers, and new methods are often evaluated only on synthetic test functions and small-scale datasets that fail to capture the challenges of LLM optimization. This impedes the development of BBO methods and makes their effectiveness on modern LLM tasks difficult to assess. We introduce \bolt, the first LLM-centric benchmark that democratizes LLM research for the BBO community. \bolt covers a broad set of well-motivated LLM optimization problems, involving multi-fidelity, multi-objective, heteroscedastic noise, high-dimensional search spaces, and black-box constraints. Each problem is grounded in thousands of real LLM experiments, and is fully reproducible and accessible through lightweight emulators or tabular datasets. We benchmark a wide range of BO and BBO methods on \bolt. BO methods generally outperform other baselines, but no single method wins across tasks, and methods remain far from optimal on some problems. These gaps underscore the need for benchmarks grounded in modern LLM tasks. \bolt is released at \url{https://github.com/chewwt/bolt}.
\end{abstract}

\section{Introduction}

Optimizing hyperparameters, data mixtures, prompts, and parallelism strategies for LLMs are critical decisions in any LLM pipeline and often have substantial computational cost. These decisions are often made by heuristics or grid search, which can be inefficient and suboptimal~\citep{albalak2024survey}. BO and other BBO methods offer a principled alternative. They handle noisy observations, require no gradients, and are designed to find good solutions in few evaluations~\citep{frazier2018tutorial,garnett2023bayesian}.

However, researchers in the BBO community often lack computational resources to validate new methods on LLM problems. As a result, many methods are only validated on numerical or classical optimization benchmarks (e.g., BBOB~\citep{hansen2009real}, HPO-B~\citep{arango2hpo}), making it difficult to reliably assess their effectiveness on LLM tasks. This impedes the development of BBO methods and leaves the potential of BBO for LLMs largely untapped.

\begin{figure}
    \begin{minipage}{0.73\textwidth}
        \centering
        \includegraphics[width=\textwidth]{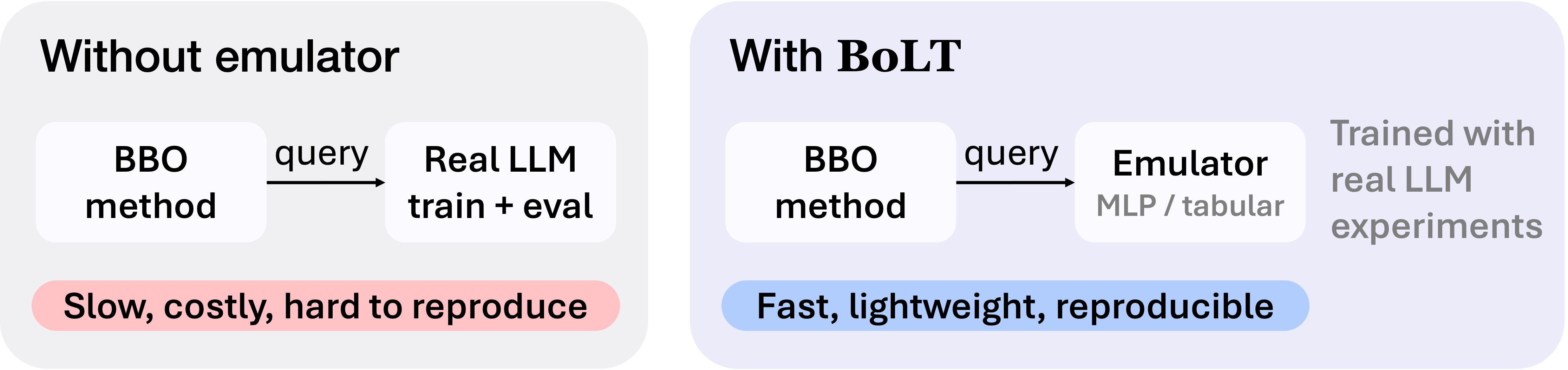}
        \caption{\bolt enables accessible and reproducible evaluation on emulated LLM tasks for BBO methods.}
        \label{fig:bolt}
    \end{minipage}
    \hfill
    \begin{minipage}{0.23\textwidth}
        \centering
        \begin{subfigure}[b]{0.42\textwidth}
            \centering
            \includegraphics[width=\textwidth]{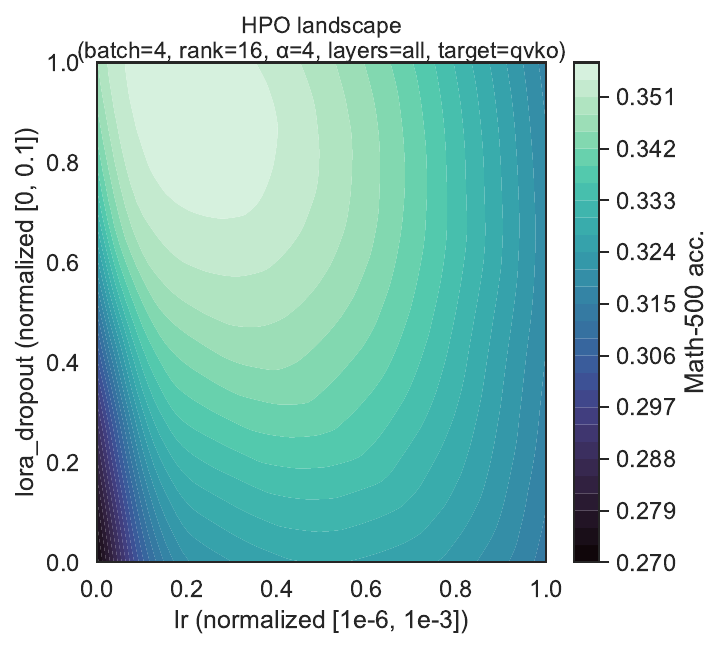}
        \end{subfigure}
        \hfill
        \begin{subfigure}[b]{0.46\textwidth}
            \centering
            \includegraphics[width=\textwidth]{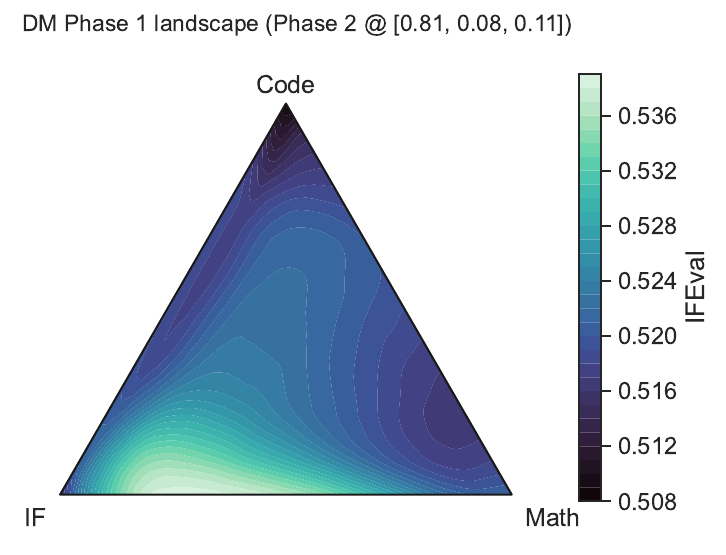}
        \end{subfigure}
        
        \begin{subfigure}[b]{0.46\textwidth}
            \centering
            \includegraphics[width=\textwidth]{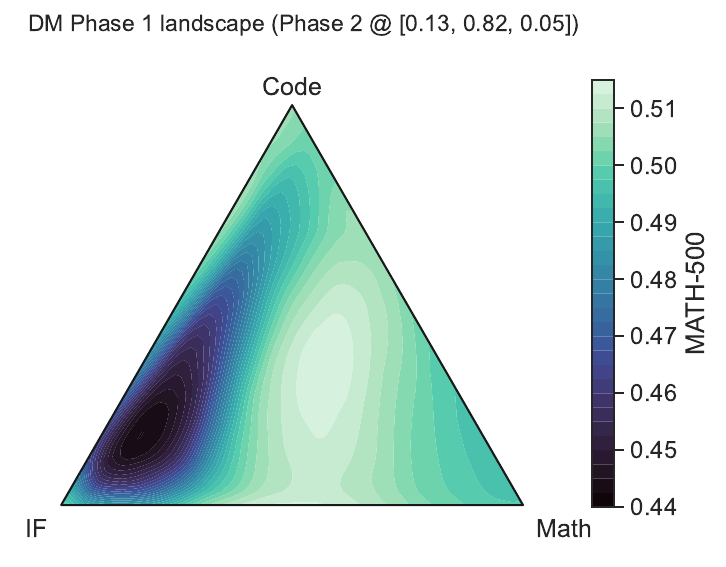}
        \end{subfigure}
        \hfill
        \begin{subfigure}[b]{0.46\textwidth}
            \centering
            \includegraphics[width=\textwidth]{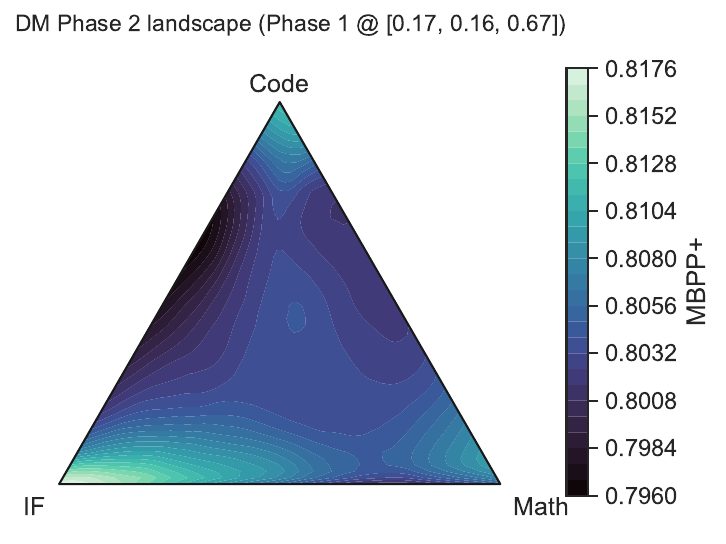}
        \end{subfigure}
        \caption{2D slices of emulator landscapes}
        \label{fig:em_slices}
    \end{minipage}
\end{figure}

We aim to democratize black-box optimization research for modern LLM problems by releasing \bolt, a \underline{\textbf{B}}enchmark for \underline{\textbf{o}}ptimization of expensive \underline{\textbf{L}}LM \underline{\textbf{T}}asks, the first black-box optimization benchmark grounded in real LLM objectives. The key goal of \bolt is to \emph{modernize the research efforts of the BBO community by allowing researchers to validate their methods on LLMs without requiring access to expensive computational resources}. 
\bolt introduces of four task families of LLM optimization tasks (\cref{sec:task-families}): LoRA hyperparameter optimization (HPO), data mixture optimization (DMO), prompt optimization (PO), and parallelism configuration optimization (PCO), all of which are grounded in real LLM objectives and are active areas of research in LLMs~\citep{tribes2024hyperparameter,chen2026chickeneggdilemmacooptimizing,chen2024towards,chen2025duetoptimizingtrainingdata,chen2026efficientlyestimatingoptimalhyperparameter,oppa}. 

Each task family consists of multiple problem instances, framed as derivative-free optimization problems that span key challenges in black-box optimization research, including multi-fidelity, multi-objective, heteroscedastic noise, high-dimensional, and black-box constrained settings. 

\bolt lowers the barrier for BBO community to adopt LLM tasks as standard, reproducible benchmarks via tabular datasets of precomputed evaluations and \emph{emulators}. In particular, \bolt consist of a suite of emulators (\cref{sec:emulators}), which are surrogate models trained on over $22$k real LLM evaluations, with noise levels set to match those measured on the true objectives. Each emulator is validated to ensure that conclusions drawn are consistent with the true objectives. The emulators require \emph{little compute and time to query}, providing access to validated LLM-grounded objectives and enabling reproducible comparisons without large-scale compute.

In \cref{sec:exps}, we perform \emph{extensive empirical evaluation of BO and black-box baselines} on the LLM tasks presented in \bolt. BO methods generally outperform standard HPO and evolutionary baselines, while our LLM tasks expose structural and practical challenges for existing methods. These results suggest that BO is a promising but underexplored direction for LLM optimization, and underscore the importance of principled, accessible, and reproducible benchmarking to realize this potential.

Our benchmark is available at \url{https://github.com/chewwt/bolt}, and emulator models and tabular data are available on the Huggingface Hub at \url{https://huggingface.co/collections/Glow-AI/bolt}. Usage examples for \bolt's API, are provided in \cref{ap:usage}.

\section{Preliminaries and Related Works}

\paragraph{Black-box optimization (BBO).}
Black-box optimization seeks to optimize a black-box objective function $f: \mathcal{X} \to \mathbb{R}^d$, where $d$ is the output dimension, $d = 1$ for single-objective problems and $d > 1$ if there are multiple objectives. Observed values are often noisy: $y = f(x) + \varepsilon$, where $\varepsilon \sim \mathcal{N}(0, \sigma_\text{N}^2)$, and $\sigma_\text{N}$ is the noise standard deviation. For instance, finding the hyperparameter that produces the best LLM performance is one such black-box optimization problem.

\paragraph{BBO for LLM.} A growing body of work frames LLM optimization problems as BBO. \citet{zhou2022large} and \citet{yang2023large} cast prompt optimization as a black-box problem with the LLM itself as optimizer. Subsequent work has proposed surrogate-based approaches, such as BO with Gaussian process or NTK kernels \citep{cheninstructzero,hu2024localized}, neural surrogates over prompt embeddings \citep{lin2024use}, and neural bandits \citep{wu2024prompt}, to improve sample efficiency.

HPO for LLM fine-tuning is another natural application for BO, as evaluation cost makes sample efficiency crucial. BO is well-established as an effective approach to HPO in machine learning \citep{snoek2012practical, shahriari2015taking}, and recent work extends this to LLM settings. \citet{jang2024model} applied multi-objective BO to jointly optimize hyperparameters and fusion weights, while \citet{seong2026efficient} injected LoRA domain knowledge to achieve strong configurations in $30$ iterations.

Beyond prompting and HPO, BO has been applied to data mixture optimization. \citet{yen2025data} proposed a multi-fidelity framework that models uncertainty across mixtures and scales, while \citet{chen2025duetoptimizingtrainingdata} used BO to re-weight post-training mixtures, achieving performance gains. 
Additionally, there are several under-explored use cases for BO in LLM training, such as to maximize the throughput of parallel (multi-GPU) LLM training, as explored by \citet{oppa} who investigated how to combine BO with prior domain knowledge about parallel LLM training.
These works indicate that BO is effective for a range of LLM optimization problems, with its sample efficiency and principled uncertainty quantification. Our benchmark complements these efforts by enabling accessible and reproducible comparison of BBO methods across these emerging problems.

\paragraph{BBO benchmarks.} BBO methods are commonly evaluated on synthetic functions, such as Ackley, Branin-Hoo, and Hartmann-6~\citep{binghamvirtuallib}, or COCO's~\citep{hansen2021coco} 24 BBOB functions~\citep{hansen2009real} and their noisy, large-scale, and bi-objective variants. Synthetic functions are convenient as they are cheap to evaluate, analytically tractable, and have known global optima, but they may not represent real-world BBO challenges well~\citep{palar2019use}.

Several real-world BBO benchmarks exist, each targeting specific problem settings. \citet{liang2021benchmarking} introduced five datasets in the material science domain and \citet{kumar2020test} introduced a suite of constrained optimization problems drawn from real-world applications, though closed-form equations are used rather than simulation or experimental results. EXPObench~\citep{bliek2023benchmarking} provides four expensive real-world problems, such as wind farm layout optimization, with mixed and conditional variables. \citet{dreczkowski2023framework} addresses mixed-variable and combinatorial BO including RNA inverse folding. Our benchmark complements these efforts with a focus on optimization problems arising in LLM development, a setting not addressed by existing BBO benchmarks.

\paragraph{Optimization benchmarks for deep learning tasks.} 
Closer to our setting are benchmark suites for HPO in machine learning. HPOBench \citep{eggensperger2hpobench} and YAHPO Gym \citep{pfisterer2022yahpo} provide surrogate- and tabular-based HPO problems on classical ML models, with support for multi-fidelity and multi-objective settings. HPO-B \citep{arango2hpo} is another black-box HPO benchmark derived from OpenML. These benchmarks have driven meaningful progress in HPO research. However, they target classical or moderately-sized ML models and do not capture the optimization challenges that arise in LLM development, a gap our benchmark is designed to address.

\begin{table}
\centering
\small
\caption{Summary of \bolt optimization problems. MF: multi-fidelity. MO: multi-objective.}
\label{tab:benchmark_summary}
\begin{tabular}{llcl}
\toprule
\textbf{Family} & \textbf{Problem} & \textbf{Dim} & \textbf{Optimization challenges} \\
\midrule
\multirow{3}{*}{Hyperparameters}
  & HPO          & 7  & mixed variables \\
  & HPO-MF-Cont  & 8  & mixed variables, MF (cont.)     \\
  & HPO-MF-Disc  & 8  & mixed variables, MF (disc.)   \\
\midrule
\multirow{3}{*}{Data mixtures}
  & DMO         & 6  & simplex constraint                  \\
  & DMO-MO      & 6  & simplex constraint, MO                 \\
  & DMO-Het  & 6  & simplex constraint, het. noise            \\
\midrule
Prompts
  & PO-\{128,256,512,768\} & 128/256/512/768 & high-dimensional \\
\midrule
Parallelism config.
  & PCO-\{16,32,64\} & 8 & black-box constraint \\
\bottomrule
\end{tabular}
\end{table}

\section{\bolt}

\bolt provides a suite of LLM problems for reproducible and accessible benchmarking of BO and BBO methods. We introduce four task families, which include optimization of \textbf{LoRA hyperparameters}, \textbf{data mixtures}, \textbf{prompts}, and \textbf{parallelism configurations}. An overview is in \cref{tab:benchmark_summary}. We provide detailed descriptions, problem formulation, and problem variants for each task family in \cref{sec:task-families}, and more details about the emulators in \cref{sec:emulators}.

\bolt is designed around three principles. It is \textbf{grounded in real LLM development workflows}, where each task family corresponds to a critical decision in the LLM pipeline. Objective landscapes are obtained from real LLM experiments, so comparisons on \bolt are meaningful for practical LLM research. \bolt also offers \textbf{structured coverage of BO challenges}, including mixed continuous-discrete-categorical search spaces, multi-fidelity observations, multi-objective trade-offs, heteroscedastic noise, constraints, and high dimensions. Finally, \bolt enables \textbf{reproducibility through emulators and tabular data}. Running BBO experiments with LLM training or inference in the loop is expensive and often irreproducible due to under-specified details. Thus, every black-box function is backed by a deterministic \emph{emulator}, a surrogate model trained on real experimental runs, or by tabular data, so BBO methods can be evaluated at negligible cost without large-scale compute.

\subsection{Task Families}
\label{sec:task-families}

\paragraph{Hyperparameter optimization (HPO).}
\label{sec:hpo}
Hyperparameters such as learning rate and batch size strongly affect LLM training outcomes~\citep{li2025predictablescaleistep}. Our HPO task optimizes LLM performance over a moderate-dimensional search space of LoRA fine-tuning hyperparameters (rank, alpha, target modules), learning rate, and batch size.

We collect datapoints from real LLM training runs with different hyperparameters. Each run fine-tunes a model on a fixed $100$k subset of OpenMathInstruct-2~\citep{toshniwalopenmathinstruct} using LoRA~\citep{hulora}, and the fine-tuned model is evaluated on MATH-500~\citep{hendrycks2021measuring,lightman2023let} using the \texttt{lm-eval} harness framework \cite{eval-harness}. 
In total, this yielded around $8$k observations per emulator across all runs and checkpoints. We trained 2 emulators (HPO-4B and HPO-8B), one for each model size (Qwen3-4B and Qwen3-8B~\citep{yang2025qwen3}). More details are in \cref{ap:hpo}.

\underline{Problem formulation}: 
The base problem search space is a $7$-dimensional \emph{mixed}-variable space, comprising continuous variables (learning rate, LoRA dropout), integer variables (batch size, LoRA rank, LoRA alpha, LoRA layers), and categorical variables (LoRA target modules). Full search space specification in \cref{tab:hpo_space}.

\underline{Problem variants}:
The base \emph{HPO} problem is single-fidelity at the full training budget, with a $7$-dimensional search space. The other two add a fidelity parameter indicating the compute budget (training tokens or model size) at which the LLM performance is observed.
\emph{HPO with continuous token fidelity} (HPO-MF-Cont) varies the number of training tokens seen, while \emph{HPO with discrete model-size fidelity} (HPO-MF-Disc) switches between a 4B low-fidelity and 8B high-fidelity model. Both are $8$-dimensional, due to additional fidelity parameter. Evaluating at a higher fidelity costs more optimization budget, but yields more accurate observations.

\paragraph{Data mixture optimization (DMO).}
\label{sec:dm}
Optimization over data mixtures can improve performance for both pretraining \citep{xie2023doremi, liuregmix} and supervised fine-tuning (SFT)~\citep{chen2025duetoptimizingtrainingdata, li2025data}. Further gains are possible by adjusting mixtures across training stages~\citep{olmo2025olmo,zhang2026beyond}, at the cost of a larger search space, or by optimizing over multiple, possibly competing, objectives. Our DMO problem settings span all of these research areas.

Data points for the emulator were collected from real LLM training runs across different training data mixtures. Each training data mixture contains 10k samples randomly drawn from three TULU-3 SFT data domains (instruction following, math, and code)~\citep{lamberttulu} at a specified mixing proportion. We perform SFT on a pretrained Qwen3-4B model and evaluate on three objectives:  IFEval~\citep{zhou2023instruction}, MATH-500 (4-shot), and MBPP+~\citep{austin2021program, liu2023your}. The DMO-4B emulator is trained on $5931$ datapoints, spanning diverse mixture configurations, and predicts all three objective scores given a data mixture curriculum. See \cref{ap:dmo} for more training details.

\underline{Problem formulation}: 
The search space is $\mathcal{X}_\text{DMO} \triangleq \big\{ x \in \mathbb{R}^6_{\geq 0} \mid \sum_{i=1}^3 x_i = 1,\ \sum_{j=4}^6 x_j = 1 \big\}$, where $x_{1:3}$ and $x_{4:6}$ are the mixture proportions over the three data sources at stage~1 (tokens $0$--$5$M) and stage~2 (tokens $5$M--$10$M), respectively (full space in \cref{tab:dmo_space}). Evaluations comprise IFEval,
MATH-500, and MBPP+, which are combined into different objectives for each problem variant.

\underline{Problem variants}: 
All three variants share $\mathcal{X}_\text{DMO}$ as the search space and a single multi-output emulator, differing only in objective definition and noise structure. The base \emph{DMO} problem defines a single objective as the unweighted mean of the three scores. \emph{Multi-objective DMO} (DMO-MO) instead returns the three scores directly via $f_{\text{DMO-MO}}: \mathcal{X}_\text{DMO} \to [0,1]^3$, targeting Pareto-front methods such as multi-objective BO~\citep{daulton2021parallel}. \emph{DMO with heteroscedastic noise} (DMO-Het) uses the same unweighted mean objective as DMO, but adds noise $\varepsilon \sim \mathcal{N}(0, \sigma_\text{DMO}^2(x))$ to the MATH-500 objective, with $\sigma_\text{DMO}(x)$ predicted from the noise emulator, making it a testbed for noise-aware methods~\citep{kersting2007most,lazaro2011variational}.

\begin{table}
\centering
\caption{Example prompt candidates and corresponding scores}
\label{tab:prompt_eg}
\begin{tabular}{lc}
\toprule
\textbf{Prompt} & \textbf{Score} \\
\midrule
Write the solution in a concise, technical style typical of advanced mathematics. & 0.810 \\
Break down the problem into smaller subproblems and solve each one sequentially.  & 0.716 \\
Provide only the final answer without any explanation or intermediate steps.       & 0.374 \\
\bottomrule
\end{tabular}
\end{table}

\paragraph{Prompt optimization (PO).}
Well-chosen prompts can substantially improve performance at inference time~\citep{zhou2022large}, but the prompt space is hard to search. The text space is discrete and unstructured, while its embedding space is high dimensional. Prompt selection at test time is also time-sensitive, and repeated long-context inferences are computationally costly. Our PO dataset makes it assesible for researchers to evaluate PO methods without repeated, expensive inference.

We pre-generated a corpus of $5{,}014$ candidate prompts for mathematical reasoning. Each prompt is embedded using EmbeddingGemma~\citep{embedding_gemma_2025}, which supports Matryoshka Representation Learning (MRL)~\citep{kusupati2022matryoshka} for meaningful representations at varying dimensionalities. Candidates are evaluated on MATH-500 (0-shot) to isolate the prompt's influence from few-shot demonstrations. Example prompts and scores are in \cref{tab:prompt_eg}. We use tabular data rather than a learned emulator here as interpolated embeddings cannot be decoded back into natural language prompts, and selecting from a candidate pool is standard in prompt optimization~\citep{zhou2022large,hu2024localized,lin2024use,shi2024efficient}. Further details are in \cref{ap:po}.

\underline{Problem formulation and variants}: 
The search space is $\mathcal{X}_\text{PO} \triangleq \{ z_i \}_{i=1}^{5014} \subset \mathbb{R}^d$, where each point $z_i$ is the $d$-dim embedding of a candidate prompt $i$. The objective $f_\text{PO}: \mathcal{X}_\text{PO} \to [0, 1]$ is the MATH-500 accuracy under prompt $i$ from Qwen3-14B. Acquisition functions optimize over the discrete candidate set.
\emph{PO-128}, \emph{PO-256}, \emph{PO-512}, and \emph{PO-768} correspond to MRL truncations at $d \in \{128, 256, 512\}$ and the full embedding dimension $d = 768$, respectively.

\paragraph{Parallelism configuration optimization (PCO).}
The leap in LLM performance is partially due to the ability to scale up LLM training by parallelizing it across many GPUs \citep{kuchaiev_nemo_2019,huang_gpipe_2019,rasley_deepspeed_2020,zhao_pytorch_2023}. 
Hyperparameters in a parallelism configuration (PC) controls how the LLM training is parallelized, and affects the training throughput (i.e., number of training steps ran per second) and the maximum GPU memory usage, and can be difficult to optimize. Methods for PCO have been introduced, and generally involves adaptive querying of PCs \citep{deepspeed_autotuning}, a surrogate model based on knowledge about parallel LLM training \citep{li_amp_2022,zheng_alpa_2022,lin_nnscaler_2024}, or both \citep{oppa}. 

We consider PCO for parallel training of LLMs on the \textsc{NeMo} framework \citep{kuchaiev_nemo_2019}, where each problem variant involves training on different hardware setups, LLMs, batch sizes and sequence lengths. 
Since the search space is tabular and discrete, we define a candidate set of PCs and performed an exhaustive run to measure the training throughput and the maximum GPU memory usage of all PCs in the candidate set on real multi-GPU hardware. Further details are in \cref{ap:dto}.

\underline{Problem formulation and variants}:
The search space is $\mathcal{X}_\text{PCO} \subset \mathbb{Z}^d$, where each point $x \in \mathcal{X}_\text{PCO}$ is a $d$-dim vector of integers representing a PC, with $|\mathcal{X}_\text{PCO}|$ in the order of $10^2$. 
The objective $f_\text{PCO}: \mathcal{X}_\text{PCO} \to (0, \infty)$ is the training throughput for a given PC.
The black-box constraint $c_\text{PCO}: \mathcal{X}_\text{PCO} \to [-1, 0) \cup \{ m_\text{OOM} \}$ gives the maximum GPU memory usage normalized to be between $-1$ and $0$, or gives value $m_\text{OOM} > 0$ if the training would result in an OOM error.
If $x$ results in an OOM error, its throughput is never measured, so $f_\text{PCO}(x)$ is estimated as the throughput of its nearest feasible neighbor.
Acquisition functions optimize over the discrete candidate set, and each query for $x$ returns both values of $f_\text{PCO}(x)$ and $c_\text{PCO}(x)$. 
PCO is defined as a constrained optimization problem where the goal is to maximize $f_\text{PCO}(x)$ such that $c_\text{PCO}(x) < 0$.
\emph{PCO-16}, \emph{PCO-32} and \emph{PCO-64} correspond to PCO problem variants for training on 16, 32 and 64 GPUs respectively, with different training setups. In this task, $d = 8$.

\subsection{Emulators}
\label{sec:emulators}

\begin{table}[t]
\centering
\caption{Validation of objective function emulators using Spearman's rank correlation $\rho_s$ between
emulator predictions and true evaluations. HF: high fidelity. LF: low fidelity.}
\label{tab:emulator-validation}
\scriptsize
\begin{tabular}{lll ccc}
\toprule
\textbf{Emulator} & \textbf{Problems} & \textbf{Objective} & \textbf{$\rho_\text{train}$}\,/\,\textbf{$\rho_\text{test}$} & \textbf{$R^2_\text{train}$}\,/\,\textbf{$R^2_\text{test}$} & \textbf{$n_{\text{train}}$}\,/\,\textbf{$n_{\text{test}}$} \\
\midrule
HPO-8B & HPO, HPO-MF-Cont, HPO-MF-Disc (HF) & \multirow{2}{*}{MATH-500} & $0.972$\,/\,$0.940$ & $0.946$\,/\,$0.926$ & $8513$\,/\,$785$ \\
HPO-4B & HPO-MF-Disc (LF)  & & $0.956$\,/\,$0.907$ & $0.936$\,/\,$0.786$ & $7970$\,/\,$819$ \\
\midrule
\multirow{3}{*}{DMO-4B} & \multirow{3}{*}{DMO, DMO-MO, DMO-Het}
  & IFEval & $ 0.776$\,/\,$0.762$ & $ 0.713$\,/\,$0.643$ & \multirow{3}{*}{$5931$\,/\,$788$} \\
  & & MATH-500 & $0.819$\,/\,$0.835$ & $0.642$\,/\,$0.664$ \\
  & & MBPP+ & $ 0.793$\,/\,$0.761$ & $ 0.670$\,/\,$0.638$ \\
\midrule
DMO-Noise & DMO-Het & MATH-500 $\sigma$ & $0.826$\,/\,-- & $0.611$\,/\,-- & $100$\,/\,-- \\
\bottomrule
\end{tabular}
\end{table}

Using emulators in place of expensive oracles is well established in optimization benchmarking~\citep{hase2021olympus}. HPOBench and YAHPO Gym both offer surrogate models, and YAHPO Gym shows surrogate-based benchmarks match real functions more closely than tabular ones.  Data points for the HPO and DMO objective emulators were selected via random sampling and adaptive strategies targeting high-variation regions and high-performing configurations.

Emulators are validated on a Sobol-sampled held-out test set. We report Spearman's rank correlation $\rho$ and $R^2$ to assess candidate ranking and calibration in \cref{tab:emulator-validation}. HPO-4B/8B are $3$-layer, $1024$-width MLPs and DMO-4B is a $3$-layer $512$-width MLP. The MLPs were chosen via sweep in \cref{ap:emulator_sweep}. Train and test metrics are close, indicating little overfitting. HPO emulators achieve high $\rho$ and $R^2$, while DMO's emulators are moderate but still reasonable, reflecting its more complex landscape (\cref{fig:em_slices}). We further validate emulator fidelity on regions where BBO methods query in \cref{ap:emulators_bbo_val}.

For DMO-Het, the noise emulator is a non-parametric kernel regression model fitted on $100$ points, each estimated from $5$ repeated evaluations. The noise model architecture was selected via $10$-fold cross-validation. More details and analysis on emulators are in \cref{ap:emulators}.

\section{Experiments}
\label{sec:exps}

We benchmark a range of common BO and BBO methods on \bolt. Noise levels $\sigma_N$ are obtained empirically for HPO and DMO problems (see \cref{ap:noise}), except for the hetereoscedastic problem DMO-Het, where the math objective uses a noise model $\sigma_\text{DMO}(x)$. PO and PCO experiments set $\sigma_N$ as $0.001$. For single-objective problems, we plot the log simple regret, $\log(f^* - \max_{t'\leq t} f(x_{t'}))$, where $x_{t'}$ is the observed point at iteration $t'$. For multi-objective problems, we plot the log hypervolume difference, $\log(\text{HV}(\text{PF}_t) - \text{HV}^*)$, where $\text{PF}_t$ is the noiseless Pareto frontier at iteration $t$. Both are standard metrics in the optimization literature \citep{srinivas2009gaussian, daulton2021parallel}. Each method has $10$ initial observations. GP-based methods are implemented in BoTorch~\citep{balandat2020botorch} with an RBF kernel and dimensionality-scaled lengthscale prior~\citep{hvarfner2024vanilla}. HPO, DMO, and PCO experiments are averaged over $10$ runs, and PO over $5$. $95\%$ confidence intervals are shaded. Further details are in \cref{ap:exp}.

\begin{figure}
    \centering
    
    % Row 1: two subfigures, second spans remaining width
    \hspace{3pt}
    \begin{subfigure}[t]{0.32\textwidth}
        \centering
        \includegraphics[width=\textwidth]{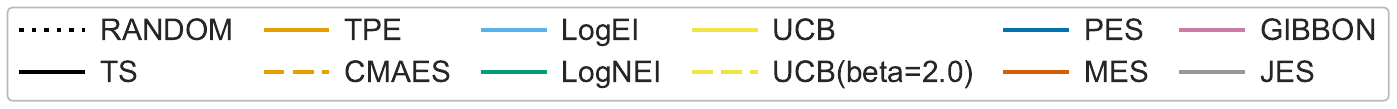}
    \end{subfigure}
    \hspace{23pt}
    \begin{subfigure}[t]{0.22\textwidth}
        \centering
        \includegraphics[width=\textwidth]{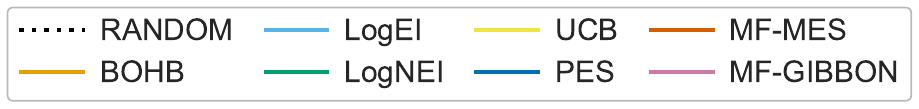}
    \end{subfigure}
    \hfill
    \begin{subfigure}[t]{0.22\textwidth}
        \centering
        \includegraphics[width=\textwidth]{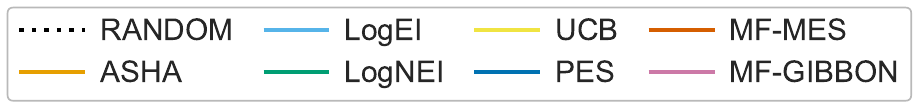}
    \end{subfigure}
    \hspace{11pt}
    
    % Row 2: three subfigures
    \begin{subfigure}[b]{0.32\textwidth}
        \centering
        \includegraphics[width=\textwidth]{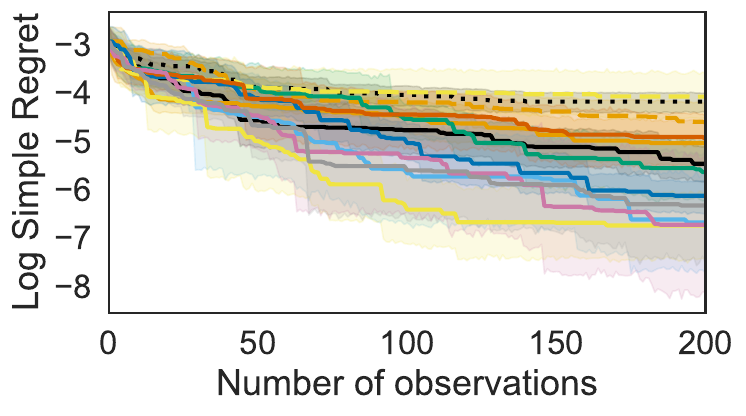}
        \caption{HPO}
        \label{fig:hpo_res}
    \end{subfigure}
    \hfill
    \begin{subfigure}[b]{0.32\textwidth}
        \centering
        \includegraphics[width=\textwidth]{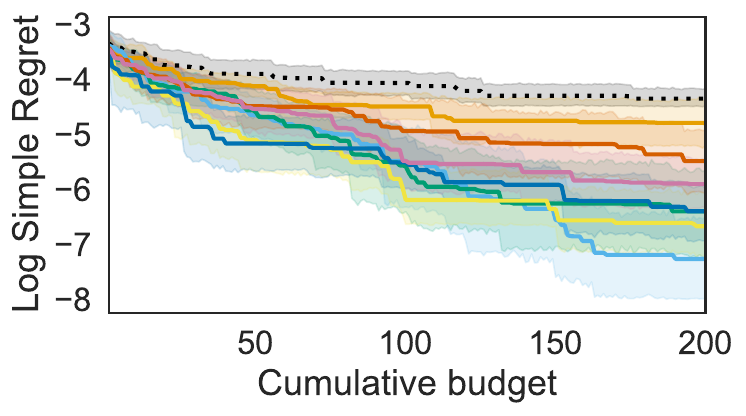}
        \caption{HPO-MF-Cont}
        \label{fig:hpo_mf_c_res}
    \end{subfigure}
    \hfill
    \begin{subfigure}[b]{0.32\textwidth}
        \centering
        \includegraphics[width=\textwidth]{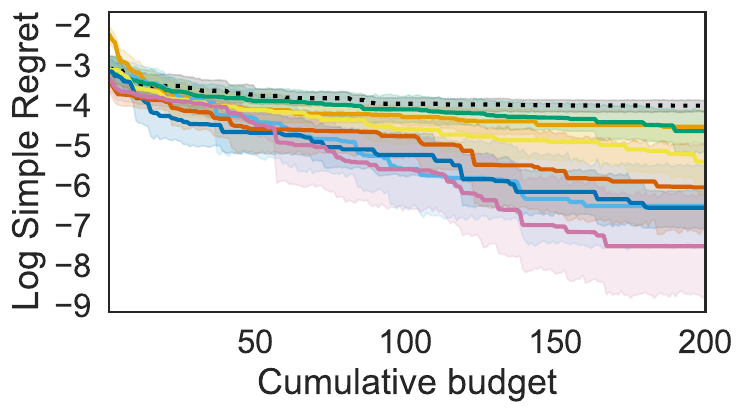}
        \caption{HPO-MF-Disc}
        \label{fig:hpo_mf_d_res}
    \end{subfigure}
    \caption{Results on \bolt's HPO problems. MF variants are plotted against cumulative budget, where each observation's cost reflects compute cost, normalized so the highest fidelity's cost is $1$.}
    \label{fig:HPO_exp}
\end{figure}

\paragraph{HPO experiments and analysis.}
We compare Thompson sampling (TS), expected improvement (EI, NEI), upper confidence bound (UCB), knowledge-gradients (KG), entropy-search methods (PES, MES, GIBBON), and common HPO baselines (TPE, CMA-ES, BOHB, ASHA). For multi-fidelity problems, we adapt acquisition functions to be cost-aware \cite{frazier2009knowledge} and report results against cumulative budget.  The cost of each multi-fidelity evaluation is $(1-\gamma)\cdot\mathrm{fid}(x) + \gamma$, where $\mathrm{fid}(x) \in [0,1]$ is the fidelity parameter and $\gamma$ is the estimated relative compute cost of the lowest fidelity. More details are in \cref{ap:hpo_exp}. Results are in \cref{fig:HPO_exp}, and FLOPs-based results in \cref{fig:HPO_flops}.

\emph{GP-based BO methods generally outperform non-GP-based HPO baselines.} (MF-)GIBBON, UCB, and LogEI are generally the stronger performers across all three settings at $200$ iterations. While non-GP baselines have substantially faster wall-clock times (\cref{tab:wall_clock_hpo_dmo}), this overhead is negligible compared to the far costlier, repeated LLM trainings with different hyperparameters at every iteration.

\emph{Evaluation budget is an important consideration for acquisition function selection.} In the standard HPO setting (\cref{fig:hpo_res}), TPE outperforms most methods under $30$ iterations but is subsequently overtaken while UCB ($\beta = 2)$ outperforms until around iteration $10$. We further analyze UCB's budget dependency through its $\beta$ parameter that balances exploration and exploitation in \cref{ap:ucb}.

\emph{MF methods are sensitive to cost scale.} Multi-fidelity acquisition functions incorporate a cost utility to discount evaluations at lower fidelities, but there is no principled way to set this. We select cost scales by comparing early-stopped runs, but this choice can significantly affect performance and fidelity allocation across methods, affecting the transferability to new problems. Analysis of different cost scales on both MF problems is in \cref{ap:cost_scale}, and fidelity allocation in~\cref{ap:fidelity}.

\begin{figure}
    \begin{minipage}{0.75\textwidth}
        \centering
    
        % Row 1: two subfigures, second spans remaining width
        \begin{subfigure}[t]{0.34\textwidth}
            \centering
            \includegraphics[width=\textwidth]{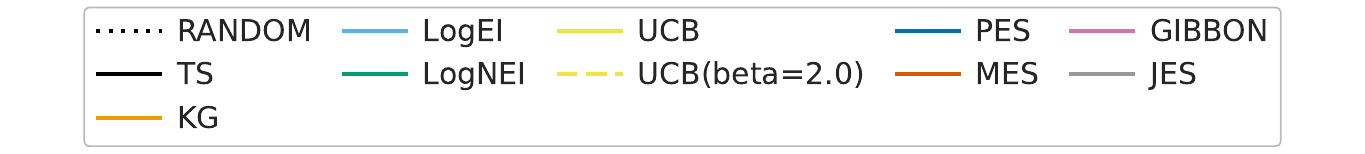}
        \end{subfigure}
        \hspace{18pt}
        \begin{subfigure}[t]{0.24\textwidth}
            \centering
            \includegraphics[width=\textwidth]{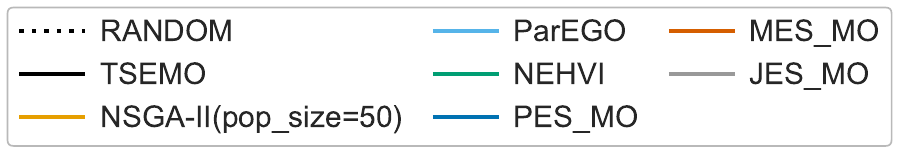}
        \end{subfigure}
        \hfill
        \begin{subfigure}[t]{0.29\textwidth}
            \centering
            \includegraphics[width=\textwidth]{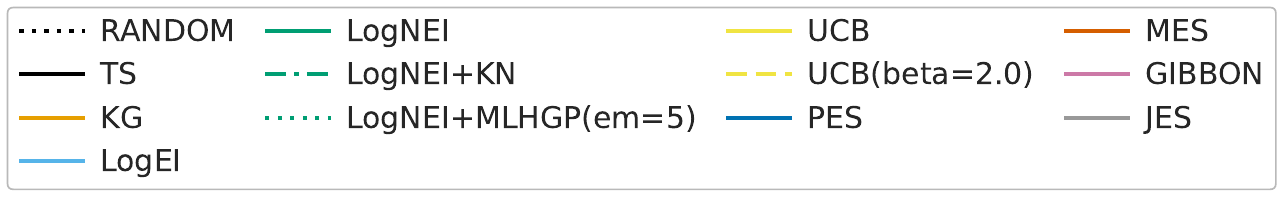}
        \end{subfigure}
        
        % Row 2: three subfigures
        \begin{subfigure}[t]{0.32\textwidth}
            \centering
            \includegraphics[width=\textwidth]{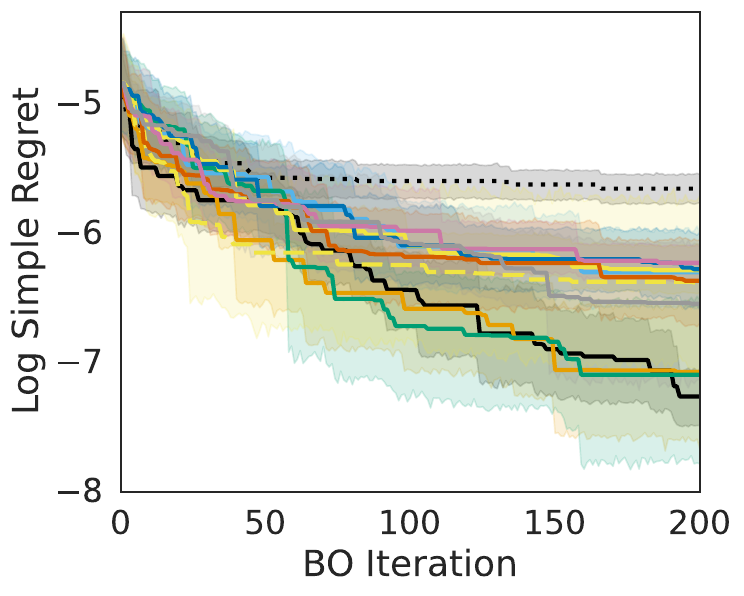}
            \caption{DMO}
            \label{fig:dmo_res}
        \end{subfigure}
        \hfill
        \begin{subfigure}[t]{0.32\textwidth}
            \centering
            \includegraphics[width=\textwidth]{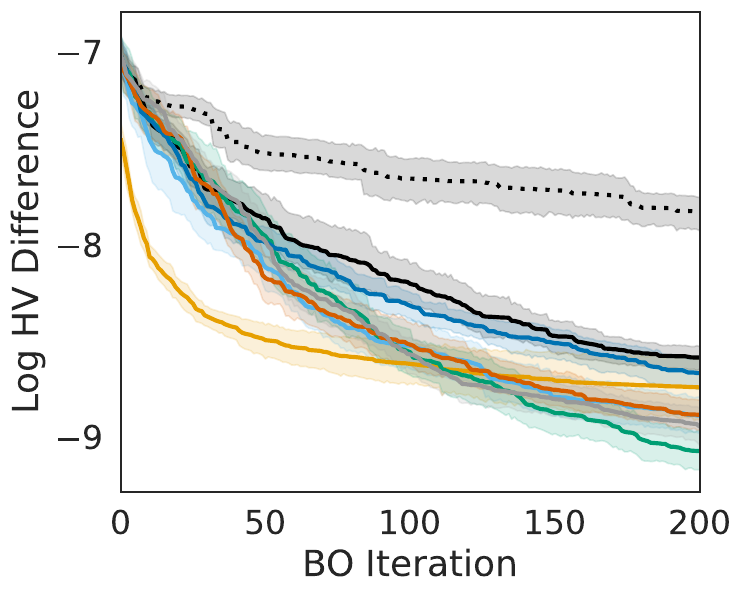}
            \caption{DMO-MO}
            \label{fig:dmo_mo_res}
        \end{subfigure}
        \hfill
        \begin{subfigure}[t]{0.32\textwidth}
            \centering
            \includegraphics[width=\textwidth]{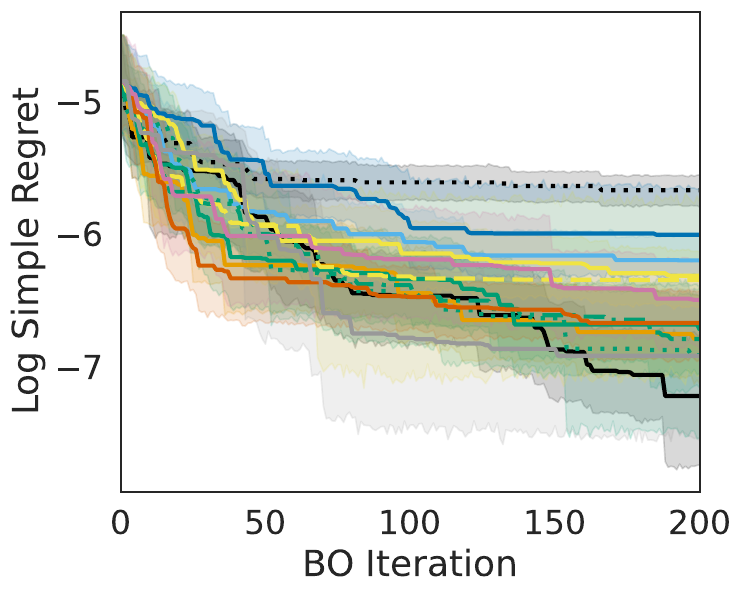}
            \caption{DMO-Het}
            \label{fig:dmo_het_res}
        \end{subfigure}
        \caption{Results on \bolt's DMO problems. NSGA2 has $50$ observations per iteration, while all other methods use only $1$ observation per iteration.}
        \label{fig:DM_exp}
    \end{minipage}
    \hfill
    \begin{minipage}{0.22\textwidth}
        % batch dm
        \centering
        
        % Row 1: 
        \begin{subfigure}[t]{\textwidth}
            \centering
            \includegraphics[width=0.7\textwidth]{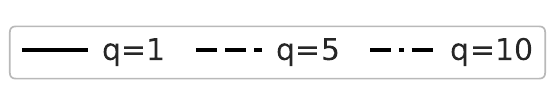}
        \end{subfigure}
        \vspace{-1em}
        % Row 2: 
        \begin{subfigure}[t]{\textwidth}
            \centering
            \includegraphics[width=\textwidth]{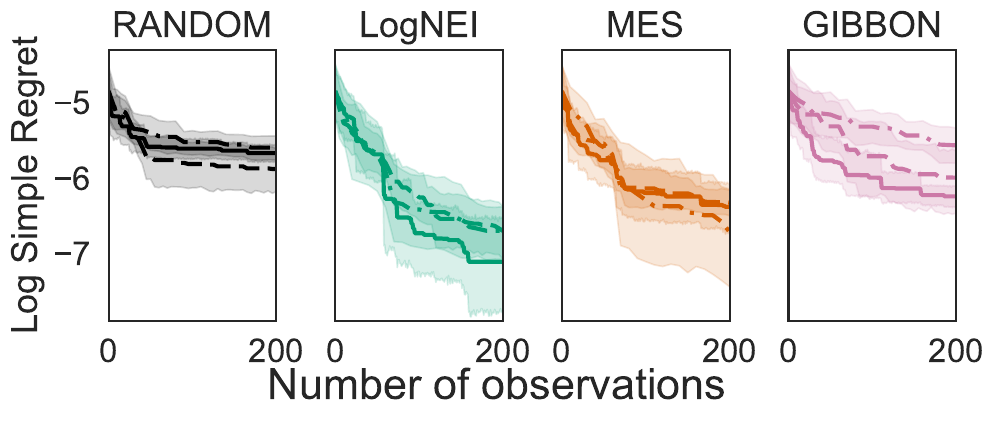}
            \label{fig:dmo_batch}
        \end{subfigure}
        \vspace{-1.5em}
        % Row 3: 
        \begin{subfigure}[t]{\textwidth}
            \centering
            \includegraphics[width=\textwidth]{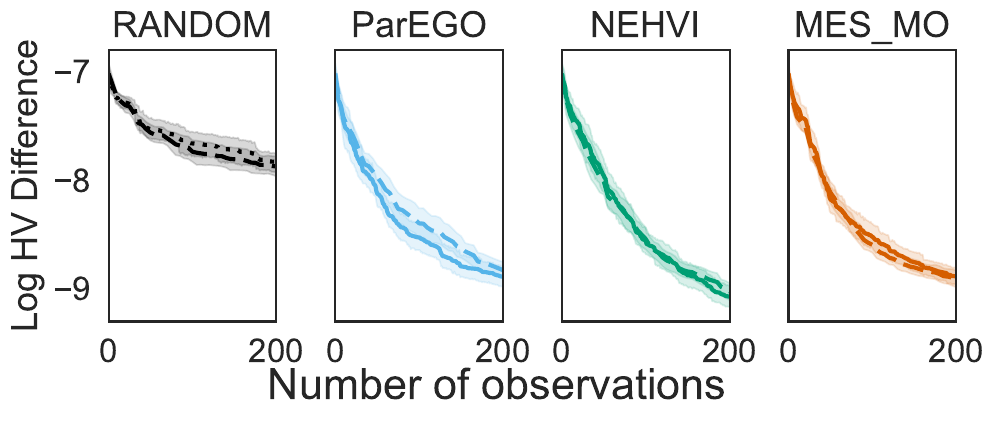}
            \label{fig:dmo_mo_batch}
        \end{subfigure}
        \caption{Different batch sizes on DMO (top) and DMO-MO (bottom).}
        \label{fig:DM_batch_exp}
    \end{minipage}
\end{figure}

\paragraph{DMO experiments and analysis.}
On DMO, we evaluate single- and multi-objective BO methods, and an evolutionary baseline NSGA-II. Entropy-based methods (MES, PES, JES) approximate the optimum's distribution over a candidate set, so we replace the standard box-uniform grid with Dirichlet-sampled points to respect the simplex constraints. For DMO-Het, we add the noise-aware MLHGP which learns a noise model with EM~\citep{kersting2007most}, and a GP with known ground-truth noise. More details are in \cref{ap:dmo_exps} and results are in \cref{fig:DM_exp}. 
  
\emph{A mix of winners.} TS leads single-objective problems at iteration $200$. KG and LogNEI also performed well on DMO, and JES on DMO-Het. For multi-objective DMO-MO, NSGA-II(pop\_size=$50$) leads until iteration $100$, after which BO methods (NEHVI, JES\_MO, ParEGO, MES\_MO), overtake it, likely due to more informative sampling despite using $50\times$ fewer function evaluations per iteration.

\emph{Noise-aware methods yield modest improvements under heteroscedastic noise.} When noise is known (LogNEI+KN) or modelled (LogNEI+MLHGP), performance is better than homoscedastic LogNEI in $200$ iterations, indicating that good noise modelling is beneficial. MLHGP takes around $100$ iterations to learn a good noise model, which we analyze in \cref{ap:mlhgp_noise}. 

\emph{Parallel batched candidates can reduce time cost with little performance loss.} Since LLM finetuning and evaluation are often conducted in parallel, we compared single versus batched observations per iteration for selected acquisition functions in \cref{fig:DM_batch_exp} on both DMO and DMO-MO, with the same observation budget of $200$. Batched BO candidates are selected via sequential greedy conditioning~\citep{balandat2020botorch}. We observed that using a batch size of $5$ or $10$ to run evaluations in parallel is often comparable to single sequential observations, other than for GIBBON.

\begin{figure}
    \centering
    \begin{minipage}{0.75\textwidth}
        \centering
        % Row 1: two subfigures, second spans remaining width
        \begin{subfigure}[t]{0.75\textwidth}
            \centering
            \includegraphics[width=\textwidth]{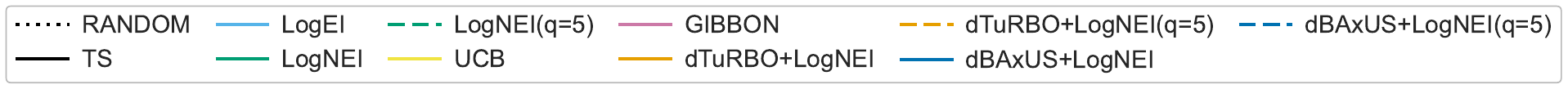}
        \end{subfigure}
        
        % Row 2: four subfigures
        \begin{subfigure}[t]{0.24\textwidth}
            \centering
            \includegraphics[width=\textwidth]{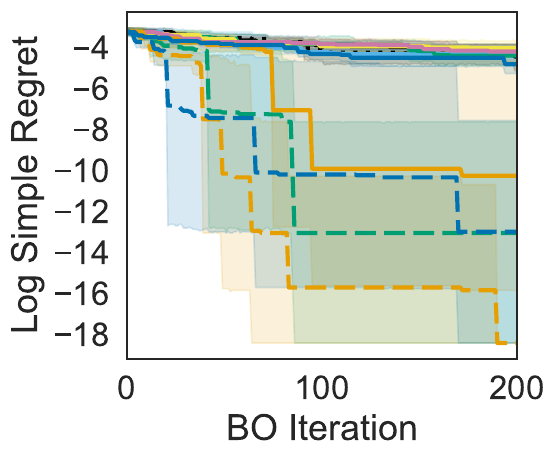}
            \caption{PO-128}
        \end{subfigure}
        \hfill
        \begin{subfigure}[t]{0.24\textwidth}
            \centering
            \includegraphics[width=\textwidth]{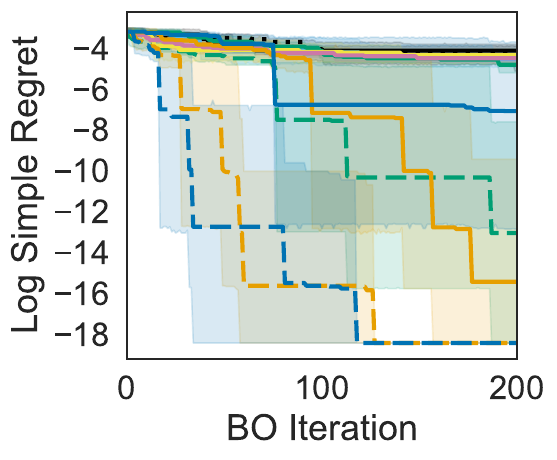}
            \caption{PO-256}
        \end{subfigure}
        \hfill
        \begin{subfigure}[t]{0.24\textwidth}
            \centering
            \includegraphics[width=\textwidth]{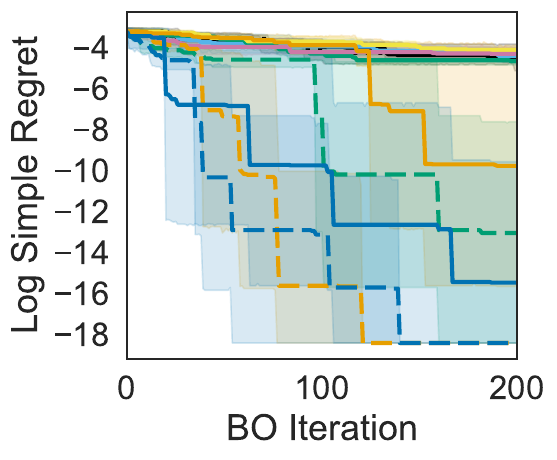}
            \caption{PO-512}
        \end{subfigure}
        \hfill
        \begin{subfigure}[t]{0.24\textwidth}
            \centering
            \includegraphics[width=\textwidth]{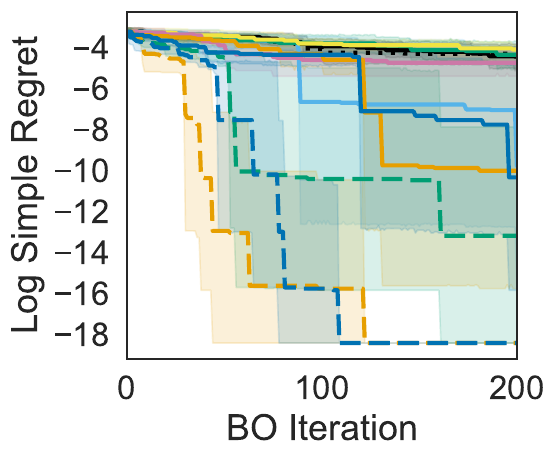}
            \caption{PO-768}
        \end{subfigure}
        \caption{Results on \bolt's PO problems. ($q=5$) indicates batch size $5$. dTURBO and dBAxUS are our adapted methods for a discretized space.}
        \label{fig:PO_exp}
    \end{minipage}
    \hfill
    \begin{minipage}{0.21\textwidth}
        \centering
        \includegraphics[width=\textwidth]{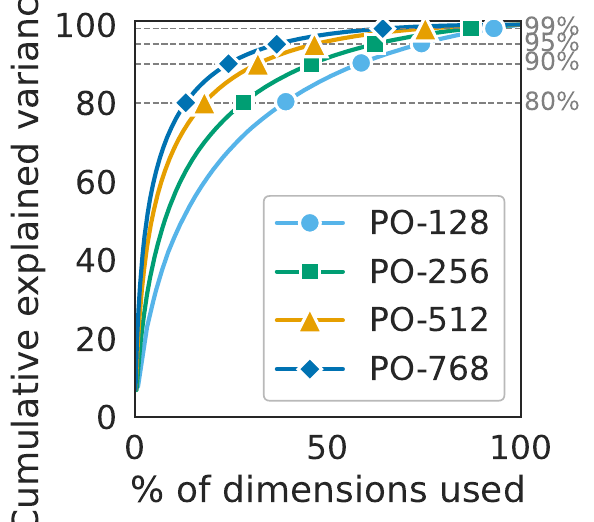}
        \caption{PCA explained variance for PO.}
        \label{fig:po_pca}
    \end{minipage}
\end{figure}

\paragraph{PO experiments and analysis.}
PO is high-dimensional, so we include trust-region and subspace-reduction methods, TuRBO~\citep{eriksson2019scalable} and BAxUS~\citep{papenmeier2022increasing}. We adapt them for discrete candidate sets as {dTuRBO} and {dBAxUS}, where the trust-region hyper-rectangle is replaced with an adaptive $k$-nearest-neighbors set. More details are in~\cref{ap:po_exps}. Batch size $5$ for selected acquitision functions are also evaluated for parallel settings. Results are in \cref{fig:PO_exp}.

\emph{Standard acquisition functions fail to scale.} Their regret plateau between $-3$ and $-5$ within $50$ iterations, with negligible improvement thereafter, and are indistinguishable from random. A notable exception is LogEI at PO-768, which descends to around $-7$, likely benefiting from richer expressivity of the full embedding dimensionality, though it remains far behind trust-region methods.

\emph{Batching helps even without trust regions.} Log simple regret of LogNEI($q=5$) reaches approximately $-12.5$ for all problems without any local search constraints and far exceeds its sequential counterpart, showing that parallel acquisition provides meaningful gains independent of trust regions. 

\emph{Local search methods lead to substantial performance gains.} Across all dimensions, dTuRBO+LogNEI and dBAxUS+LogNEI reach log simple regret of $-10$ to $-15$, showing the $k$-NN adaptation preserves the effectiveness of local search in discretized search spaces. Combined with batch evaluation, dTuRBO+LogNEI($q=5$) and dBAxUS+LogNEI($q=5$) achieve the lowest regret, approaching the optimal solution within $200$ iterations across almost all PO problems.

\emph{Linear subspace reduction may be unnecessary for LLM embedding spaces.} At PO-128, dBAxUS underperforms dTuRBO, as PO-128 embeddings need $74.2\%$ of principal components to explain $95\%$ of variance (\cref{fig:po_pca}), violating BAxUS's low-dimensional linear active-subspace assumption \citep{papenmeier2022increasing}. At higher dimensions, variance concentrates in fewer components and dBAxUS recovers. This suggests subspace reduction is most effective when the embedding space has dimensional redundancy, but LLM embeddings may not always have this property.

\begin{figure}
    \centering
    \begin{subfigure}[t]{0.9\textwidth}
        \centering
        \includegraphics[width=\textwidth]{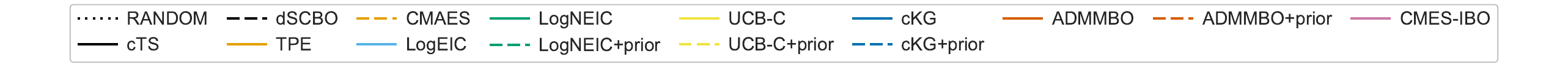}
    \end{subfigure}
    
    \begin{subfigure}[t]{0.32\textwidth}
        \centering
        \includegraphics[width=\textwidth]{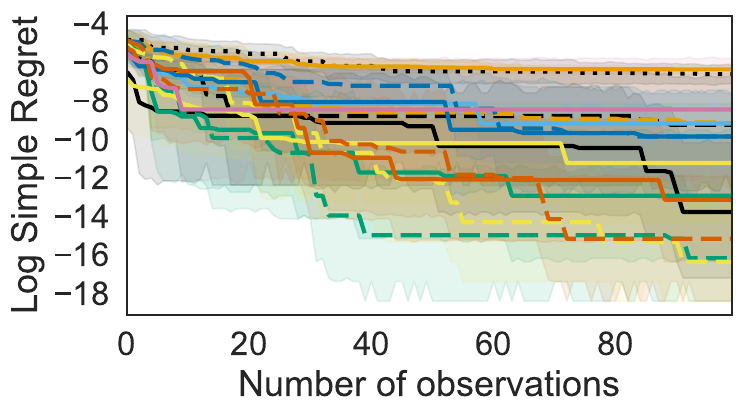}
        \caption{PCO-16}
    \end{subfigure}
    \hfill
    \begin{subfigure}[t]{0.32\textwidth}
        \centering
        \includegraphics[width=\textwidth]{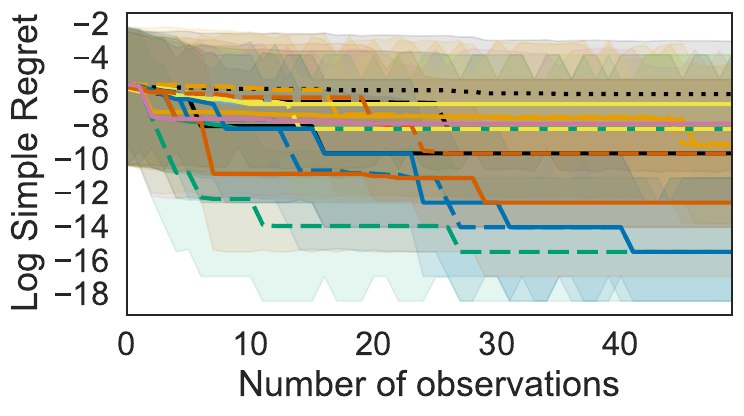}
        \caption{PCO-32}
    \end{subfigure}
    \hfill
    \begin{subfigure}[t]{0.32\textwidth}
        \centering
        \includegraphics[width=\textwidth]{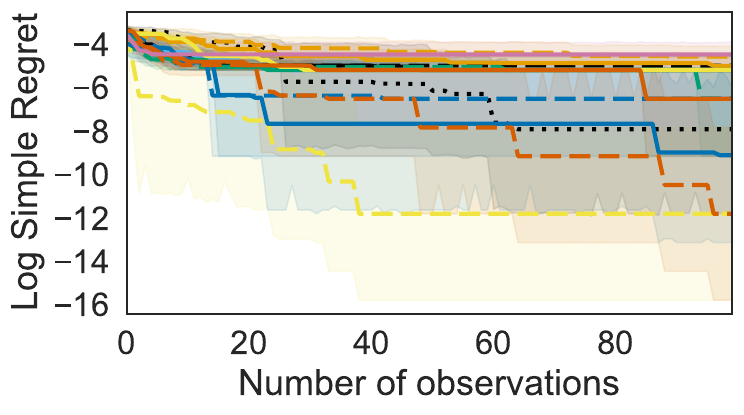}
        \caption{PCO-64}
    \end{subfigure}
    \caption{Results on \bolt's PCO problems.}
    \label{fig:PCO_exp}
\end{figure}

\paragraph{PCO experiments and analysis.}
\label{sec:dto}
We evaluated constrained BO methods: LogEIC~\citep{pmlr-v32-gardner14}, LogNEIC~\citep{letham2019cnei}, UCB-C~\citep{nguyen2024optimistic}, dSCBO~\citep{eriksson2021scalable} (discretized adaptation), cTS, CMES-IBO~\citep{takeno2022sequential}, cKG~\citep{ungredda2024bayesian},  ADMMBO~\citep{ariafar2019admmbo}, and BBO baselines (TPE, CMA-ES). For LogNEIC, UCB-C, cKG, and ADMMBO, we include a variant with GP prior means designed from PC knowledge~\citep{oppa}. More details in \cref{ap:pco_exps}. Regret plots are in \cref{fig:PCO_exp}, and \cref{fig:PCO_exp_outcomes} shows how close each method's final recommendation is to the optimum across seeds.

\emph{The parallelism prior generally helps.} On PCO-16, it raises how often UCB-C, LogNEIC, and ADMMBO find the optimum (3$\to$8, 5$\to$8, 5$\to$7 of 10 seeds),  giving the three lowest-regret methods. On PCO-64, it raises UCB-C and ADMMBO from 0 and 1 to 5/10 seeds. On PCO-32, LogNEIC+prior leads for the first $\sim$40 iterations and finds the optimum in 8/10 seeds, up from 3.

\emph{Performance depends on the landscape.} PCO-32 has a sharp optimum, while PCO-16 and PCO-64 have near-optimal plateaus (\cref{ap:PCO_landscape}), where many methods stall within $2.5\%$ of the optimum without reaching it. On PCO-64, most are even outperformed by random search. No method is best on all three problems, but a prior variant is best or tied for best on each: LogNEIC+prior (with cKG) on PCO-32, and UCB-C+prior on PCO-16 and PCO-64.

\begin{figure}
    \begin{minipage}{0.22\textwidth}
        \centering
        \begin{subfigure}[t]{0.96\textwidth}
            \centering
            \includegraphics[width=\textwidth]{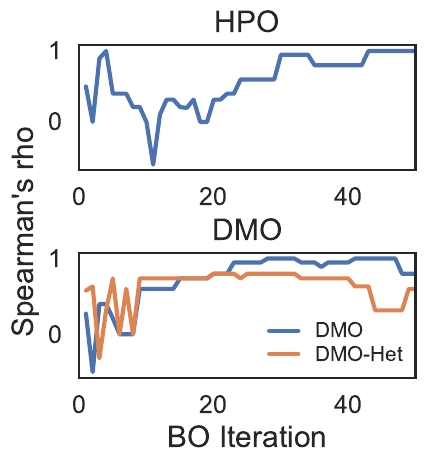}
        \end{subfigure}
        \caption{$\rho$ on mean ranks between real and emulated runs.}
        \label{fig:real_emu_rho}
    \end{minipage}
    \hfill
    \begin{minipage}{0.28\textwidth}
        \centering
        \begin{subfigure}[t]{\textwidth}
            \centering
            \includegraphics[width=\textwidth]{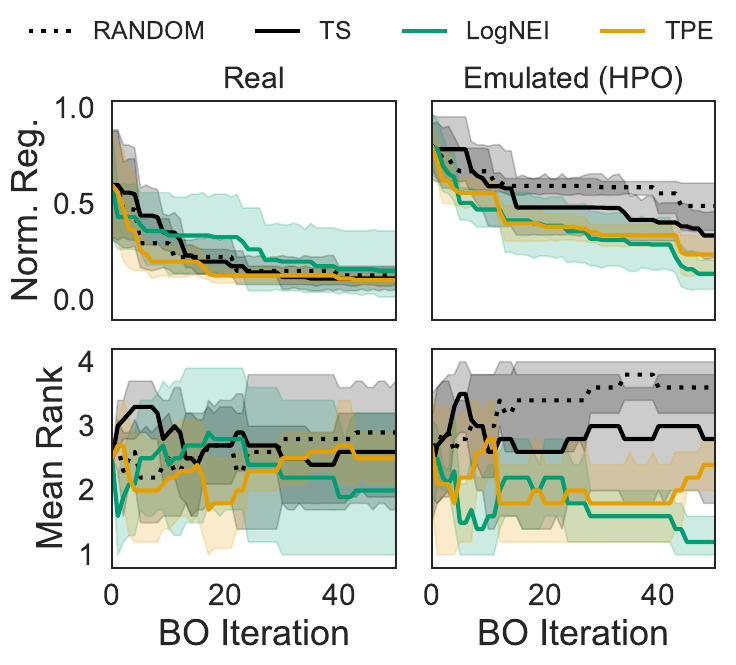}
        \end{subfigure}
        \caption{Real versus emulated for HPO.}
        \label{fig:HPO_real_emu}
    \end{minipage}
    \hfill
    \begin{minipage}{0.42\textwidth}
        \centering
        \begin{subfigure}[t]{\textwidth}
            \centering
            \includegraphics[width=\textwidth]{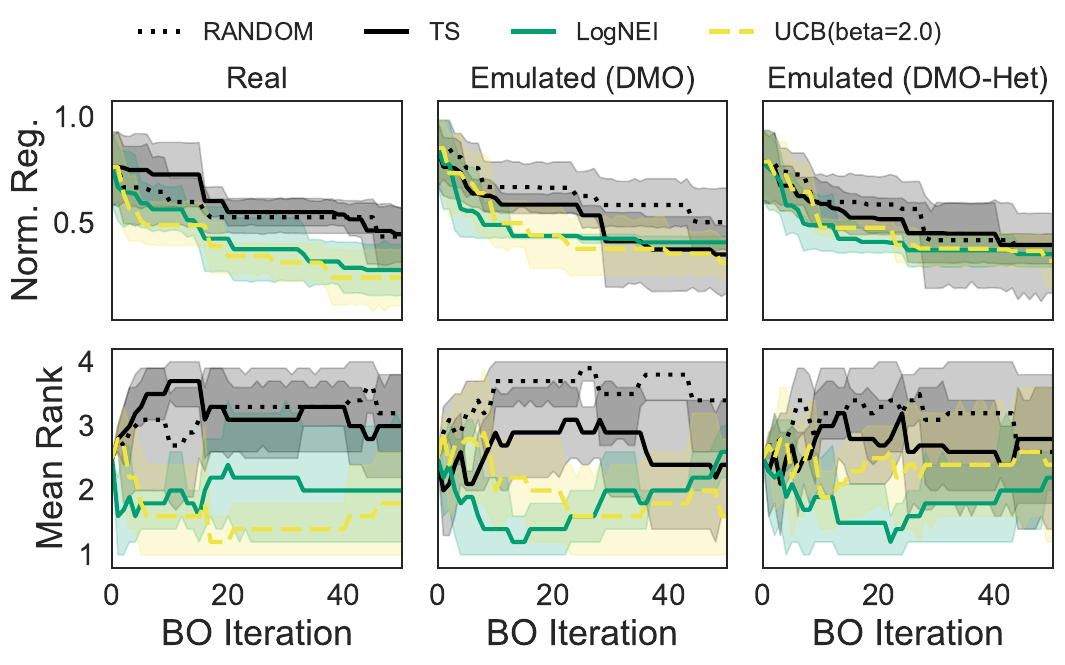}
        \end{subfigure}
        \caption{Real versus the emulated DMO and DMO-Het for data mixtures.}
        \label{fig:DMO_real_emu}
    \end{minipage}
\end{figure}

\paragraph{Transferability to real LLM training.}
\label{sec:transfer}
We run selected methods on HPO, DMO, and DMO-Het for $50$ iterations over $5$ seeds with both real LLM evaluation in the loop and with emulator outputs, to assess how well emulator rankings transfer to real LLM training. Methods are ranked by normalized regret, $(y_\text{best} - y_\text{min}) / (y_\text{max} - y_\text{min})$, as in HPO libraries like YAHPO Gym, because real runs only yield observed values, not the noiseless objective required for simple regret. We do not use normalized regret in previous sections as it is less stable across seeds than simple regret (\cref{ap:regrets}).

As seen in \cref{fig:real_emu_rho}, rankings stabilize and largely agree after around $30$ iterations for HPO and $20$ for DMO. For HPO in \cref{fig:HPO_real_emu}, random performs worst and LogNEI best in both settings, while TPE and TS rankings are tighter in the real runs. LogNEI's worse mean normalized regret on the real experiments stems from a single bad seed, and it still achieves the best mean ranking at $50$ iterations. DMO and DMO-Het results are in \cref{fig:DMO_real_emu}. DMO's rankings match the real runs more closely than DMO-Het's, with both agreeing on LogNEI $>$ UCB ($\beta=2$) $>$ TS $>$ Random after iteration $20$. DMO-Het's main discrepancy is underestimating UCB, likely because its exploration is more sensitive to the noise model, with noise errors amplified in its confidence bounds.

\section{Conclusion}

Black-box optimization methods hold significant promise for automating and accelerating LLM pipeline optimization, but progress has been hampered by the lack of accessible and reproducible evaluations grounded in real LLM experiments. 
\bolt is critical to unlocking this progress, by providing emulators and tabular datasets built on real LLM experiments, enabling fast, reproducible benchmarking without large-scale compute.
Our evaluation shows that mature BO methods transfer to LLM tasks, but no single method wins across tasks, and strong performance often requires adapting to LLM-specific structure, such as simplex-constrained mixtures, high-dimensional discrete prompt spaces, and near-optimal plateaus. Domain knowledge also helps, as parallelism priors improve several constrained BO methods. These challenges reflect the structure of real LLM pipeline optimization problems and highlight gaps for the BBO community to tackle.

\paragraph{Limitations and future extensions.}
To collect sufficient data to fit reliable emulators for \bolt, we introduced measures such as smaller training token budgets, reduced maximum token limits, and the exclusion of chain-of-thought reasoning in HPO tasks. This may not represent the best achievable performance of the evaluated models under unconstrained inference. Nevertheless, the optimization challenges \bolt is designed to benchmark (mixed-variable search spaces, simplex-constraints, etc.) are intrinsic to the structure of LLM tasks themselves and remain consistent across inference and training configurations. We also plan to update the benchmark with new LLM architectures in the future, in an effort to continue modernizing optimization benchmarks.

In practice, \bolt is extensible to more optimization problems, for example contextual  optimization, as new optimization settings can be constructed using our publicly available emulators and datasets. There are also many other LLM-centric tasks that fit the BBO framework, such as decoding configuration and RLHF reward weighting, all of which can be future extensions of \bolt.

% \clearpage

\section*{AI use statement}

In this work, we used generative AI tools for assisting the implementation of data collection code and methods benchmarked, refining data collection methodology, formatting of datasets, and some data analysis.
We have not used generative AI tools for creating synthetic datasets,
and mathematical claims, proofs, hypotheses, and translation are not applicable to this work.
Additionally, we used generative AI tools for  identifying relevant literature and editing for readability. We have reviewed all AI-assisted work. For example, we verified the methods' implementation against released codes, if available, and looked through the codes manually. We take responsibility for the final content of this work,
including text, claims or artifacts produced with the aid of generative AI.

\section*{Ethics statement}

Our dataset and emulators are trained on our own generated LLM results, and does not have privacy, consent, copyright issues. It also does not have potential harmful consequences, beyond those from standard BBO and LLM research, to the best of our knowledge.

\section*{Reproducibility statement}

We provide extensive implementation details for each set of experiments, including training and inference configurations, kernel choices, constants for noise, cost scale, and more, especially in our appendix. Our codes, emulators, and datasets are also publicly available.



\bibliography{refs}
\bibliographystyle{iclr2027_conference}

\clearpage
\appendix

\section{More details on \bolt's codes and usage}
\label{ap:usage}

\subsection{\bolt usage examples}

\begin{listing}[H]
\begin{lstlisting}
from bolt import HPO
import torch

# Instantiate a problem with default noise_std
problem = HPO()

# BoTorch-compatible metadata
problem.bounds           # (2, 7) tensor of lower/upper bounds
problem.dim              # search-space dimensionality
problem.continuous_inds  # indices of continuous variables
problem.discrete_inds    # indices of integer variables
problem.categorical_inds # indices of categorical variables

# Sample random candidates and round integer/categorical variables
lb, ub = problem.bounds
X = lb + torch.rand(10, problem.dim) * (ub - lb)
X[:, problem.discrete_inds]    = X[:, problem.discrete_inds].round()
X[:, problem.categorical_inds] = X[:, problem.categorical_inds].round()
y = problem(X)  # (10, 1)
\end{lstlisting}
\caption{\bolt usage example.}
\end{listing}

\section{Additional emulator details and analysis}
\label{ap:emulators}

\subsection{Emulator Data Points}

For the HPO and DMO objective emulators, data points are selected via random sampling and two adaptive sampling strategies: one maximizes emulator gradient magnitudes to target high-variation regions, and one maximizes predicted objectives to anchor high-performing configurations. Random sampling provides broad coverage while the adaptive strategies concentrate effort where it matters most.

\subsection{Emulator MLP Sweep and Training}
\label{ap:emulator_sweep}

\textbf{Emulator sweep.} The MLP configuration for each HPO and DMO objective was selected via a sweep of MLP configurations on depth and width, shown in \cref{tab:arch_abl_hpo} and \cref{tab:arch_abl_dmo}. The final selected MLP configurations are 3-layer, 1024-width MLPs for HPO-4B and HPO-8B emulators, and 3-layer, 512-width MLP for DMO-4B.

\textbf{Emulator training.} The training recipes are optimized for each emulator. HPO-4B is trained for $4$k steps, with cosine scheduler, HPO-8B for $20$k steps with no scheduler, both at learning rate $10^{-4}$. For DMO-4B, it was trained for $50$k steps, with a learning rate of $3\times10^{-4}$ and a cosine scheduler.

\begin{table}[t]
\centering
\small
\setlength{\tabcolsep}{3.5pt}
\caption{MLP sweep for HPO emulators. 5 seeds per configuration. The $R^2$ and Spearman's $\rho$ are calculated on a held-out test set (819 for HPO-4B, 785 for HPO-8B). $\rho_{25}$: Spearman over the top quartile of the true objective. $\rho_{\mathrm{mt}}$: Spearman at the full token budget. Best value per column, per panel is bolded and second-best underlined. Final selected MLP configuration is also bolded.}
\label{tab:arch_abl_hpo}
\begin{tabular}{ll rrrr}
\toprule
Depth & Width & $R^2$ & $\rho$ & $\rho_{25}$ & $\rho_{\mathrm{mt}}$ \\
\midrule
\multicolumn{6}{l}{\emph{HPO-4B}} \\
2 & 256            & .785$\pm$.021 & .898$\pm$.009 & .538$\pm$.035 & .934$\pm$.016 \\
2 & 512            & .803$\pm$.012 & .904$\pm$.002 & \underline{.634}$\pm$.016 & .933$\pm$.016 \\
2 & 1024           & .802$\pm$.009 & \underline{.908}$\pm$.004 & \textbf{.660}$\pm$.006 & .938$\pm$.012 \\
3 & 256            & .812$\pm$.015 & .902$\pm$.004 & .555$\pm$.008 & .926$\pm$.027 \\
3 & 512            & .821$\pm$.016 & .905$\pm$.003 & .618$\pm$.017 & .928$\pm$.007 \\
\textbf{3} & \textbf{1024}  & .835$\pm$.007 & \underline{.908}$\pm$.004 & \underline{.634}$\pm$.006 & \textbf{.940}$\pm$.009 \\
4 & 256            & .814$\pm$.011 & .904$\pm$.007 & .544$\pm$.043 & .928$\pm$.035 \\
4 & 512            & .820$\pm$.011 & .906$\pm$.006 & .611$\pm$.028 & .936$\pm$.009 \\
4 & 1024           & \textbf{.841}$\pm$.008 & \underline{.908}$\pm$.004 & .631$\pm$.011 & .935$\pm$.005 \\
5 & 256            & .799$\pm$.009 & .899$\pm$.012 & .533$\pm$.032 & .932$\pm$.031 \\
5 & 512            & .817$\pm$.010 & .907$\pm$.003 & .593$\pm$.034 & \underline{.939}$\pm$.007 \\
5 & 1024           & \underline{.836}$\pm$.005 & \textbf{.909}$\pm$.002 & .604$\pm$.026 & .935$\pm$.004 \\
\midrule
\multicolumn{6}{l}{\emph{HPO-8B}} \\
2 & 256            & .914$\pm$.015 & .933$\pm$.006 & .772$\pm$.024 & .881$\pm$.018 \\
2 & 512            & .916$\pm$.012 & .941$\pm$.007 & \textbf{.810}$\pm$.024 & .885$\pm$.005 \\
2 & 1024           & .912$\pm$.014 & \underline{.944}$\pm$.003 & \underline{.808}$\pm$.012 & .887$\pm$.004 \\
3 & 256            & .911$\pm$.024 & .934$\pm$.009 & .762$\pm$.037 & .873$\pm$.018 \\
3 & 512            & .920$\pm$.010 & .940$\pm$.007 & .781$\pm$.014 & .881$\pm$.009 \\
\textbf{3} & \textbf{1024}  & \textbf{.927}$\pm$.006 & \textbf{.946}$\pm$.005 & .804$\pm$.030 & \textbf{.893}$\pm$.009 \\
4 & 256            & .911$\pm$.017 & .937$\pm$.008 & .732$\pm$.044 & .883$\pm$.029 \\
4 & 512            & .920$\pm$.012 & .939$\pm$.007 & .774$\pm$.038 & \textbf{.893}$\pm$.013 \\
4 & 1024           & \underline{.926}$\pm$.007 & .942$\pm$.007 & .789$\pm$.016 & .884$\pm$.008 \\
5 & 256            & .905$\pm$.005 & .935$\pm$.007 & .709$\pm$.048 & .876$\pm$.011 \\
5 & 512            & .920$\pm$.010 & .938$\pm$.007 & .777$\pm$.033 & .876$\pm$.010 \\
5 & 1024           & .924$\pm$.009 & .941$\pm$.008 & .767$\pm$.031 & \underline{.888}$\pm$.008 \\
\bottomrule
\end{tabular}
\end{table}

%  dmo emus
\begin{table}[t]
\centering
\small
\setlength{\tabcolsep}{4pt}
\caption{MLP sweep for the DMO-4B emulator. 5 seeds per configuation, on 788 held-out Sobol points. $R^2$ / $\rho$ / $\rho_{25}$ are unweighted means over ifeval / math / code. Best value per column bolded and second-best underlined. Selected MLP configuration is also in bold. }
\label{tab:arch_abl_dmo}
\begin{tabular}{ll rrr}
\toprule
Depth & Width & $R^2$ & $\rho$ & $\rho_{25}$ \\
\midrule
2 & 256           & .644$\pm$.001 & .785$\pm$.001 & .446$\pm$.002 \\
2 & 512           & \underline{.646}$\pm$.001 & \textbf{.786}$\pm$.000 & \underline{.447}$\pm$.003 \\
2 & 1024          & \underline{.646}$\pm$.000 & \textbf{.786}$\pm$.000 & .446$\pm$.001 \\
\addlinespace
3 & 256           & .645$\pm$.001 & .784$\pm$.001 & .446$\pm$.003 \\
\textbf{3} & \textbf{512} & \textbf{.647}$\pm$.001 & \textbf{.786}$\pm$.000 & \textbf{.448}$\pm$.003 \\
3 & 1024          & \textbf{.647}$\pm$.001 & \underline{.785}$\pm$.000 & .444$\pm$.001 \\
\addlinespace
4 & 256           & .642$\pm$.001 & .782$\pm$.001 & .441$\pm$.003 \\
4 & 512           & \textbf{.647}$\pm$.001 & \underline{.785}$\pm$.000 & .445$\pm$.000 \\
4 & 1024          & \textbf{.647}$\pm$.001 & \underline{.785}$\pm$.000 & .444$\pm$.003 \\
\addlinespace
5 & 256           & .639$\pm$.000 & .781$\pm$.001 & .439$\pm$.002 \\
5 & 512           & .645$\pm$.000 & .784$\pm$.000 & .442$\pm$.004 \\
\bottomrule
\end{tabular}
\end{table}

\subsection{Emulator analysis}

\begin{figure}
    \centering
    \includegraphics[width=\textwidth]{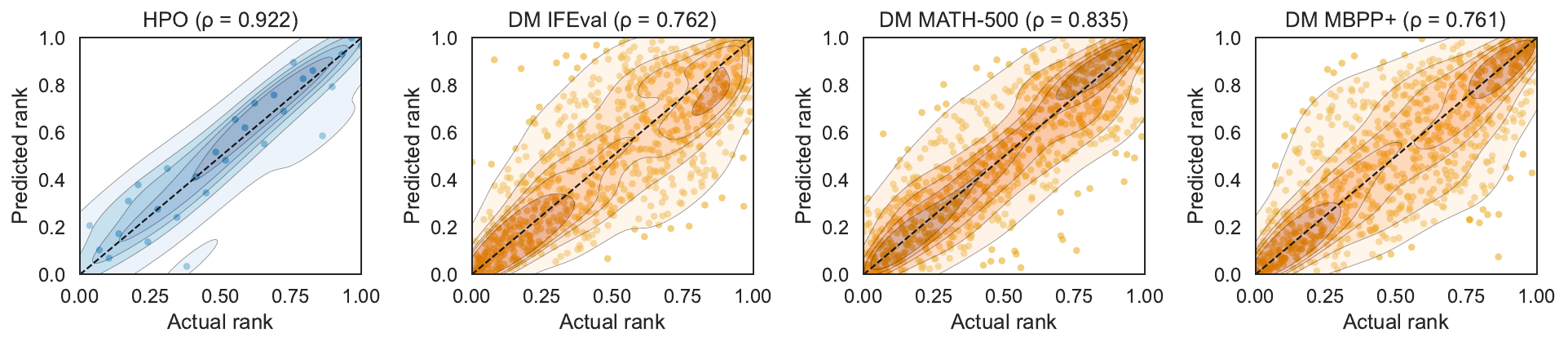}
    \caption{Real and predicted ranks on test set.}
    \label{fig:em_rank}
\end{figure}

\begin{figure}
    \centering
    \includegraphics[width=0.5\textwidth]{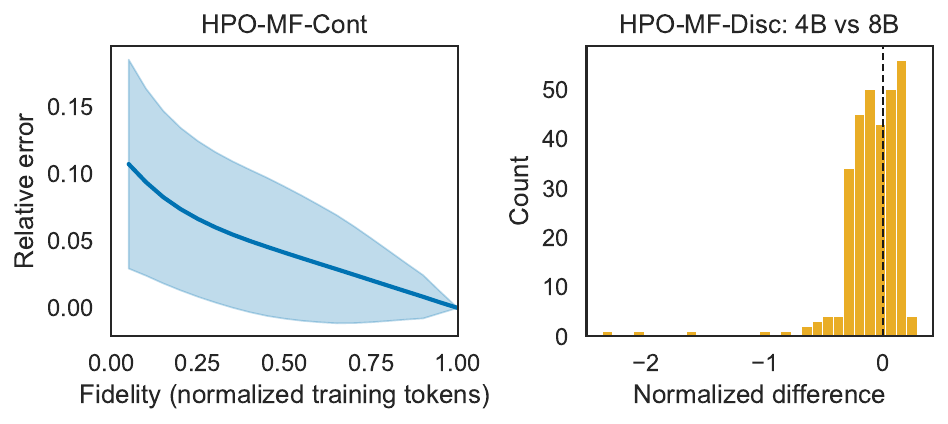}
    \caption{Differences in emulated objective at different fidelities}
    \label{fig:em_mf}
\end{figure}

\begin{figure}
    \centering
    \includegraphics[width=\textwidth]{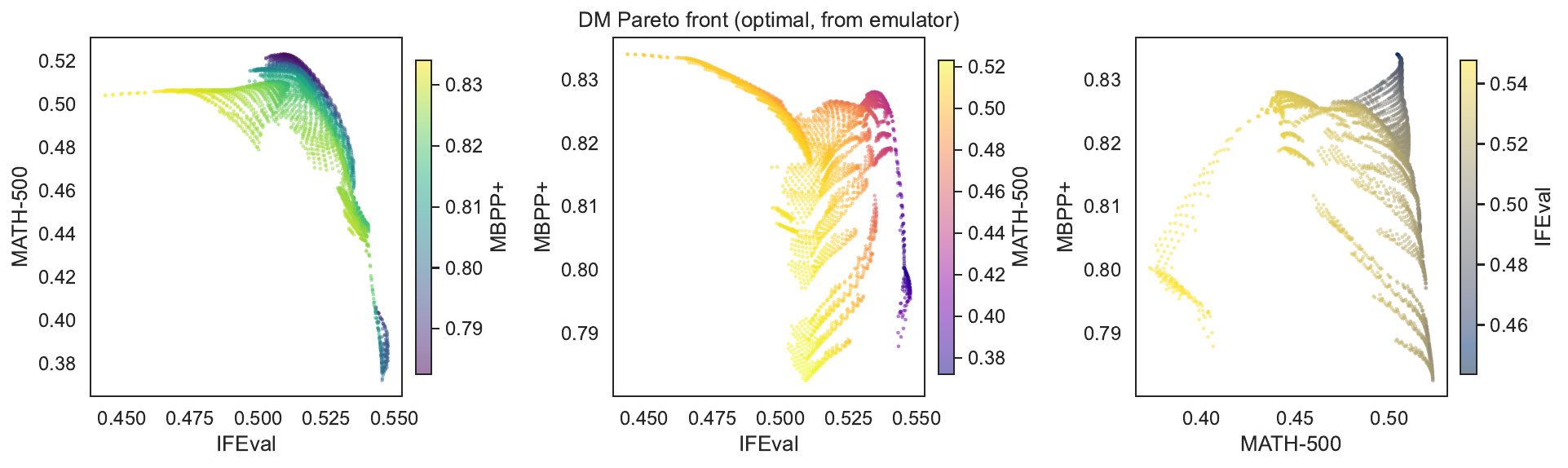}
    \caption{Estimated optimal Pareto front, as 2D slices of a 3D space}
    \label{fig:em_mo}
\end{figure}

\begin{figure}
    \centering
    \begin{subfigure}[b]{0.66\textwidth}
        \centering
        \includegraphics[width=\textwidth]{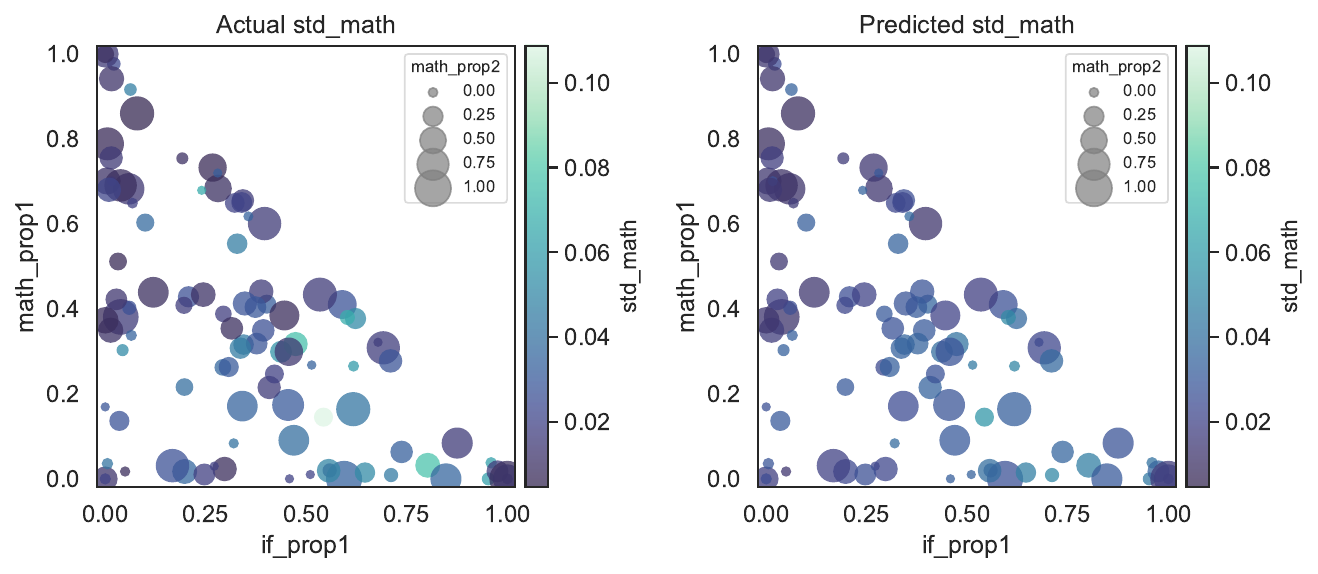}
    \end{subfigure}
    \hfill
    \begin{subfigure}[b]{0.29\textwidth}
        \centering
        \includegraphics[width=\textwidth]{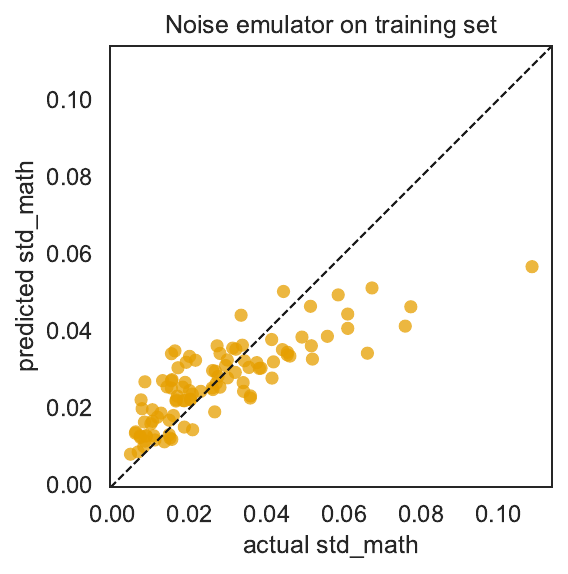}
    \end{subfigure}
    
    \caption{Noise emulator predictions for MATH-500 standard deviation. if\_prop1 indicates mixture proportion of information following data at stage 1 of the curriculum, while math\_prop1 and math\_prop2 represent math data proportions at stage 1 and 2 respectively.}
    \label{fig:em_noise}
\end{figure}

\textbf{Actual and predicted rank comparison} of data points in the test set is shown \cref{fig:em_rank}, where predicted ranks largely align with actual ranks. For the DMO emulator, lower- and higher-ranked points are generally better aligned than intermediate ones, which could be due to mixture compositions in the intermediate range being more similar and harder to discriminate.

\textbf{Multi-fidelity settings.} We analyze the HPO emulators under multi-fidelity settings in \cref{fig:em_mf}. For continuous fidelity, as expected, error relative to the highest fidelity is largest at lower fidelities. For discrete fidelity, differences largely cluster around $-0.6$ to $0.2$, indicating that the 4B model serves as a reliable proxy for the 8B model's performance.

\textbf{Multi-objective setting.} We visualize the optimal Pareto frontier of the DMO-4B emulator in \cref{fig:em_mo}, shown as three 2D slices of the joint IFEval--MATH-500--MBPP+ objective space. A clear trade-off is visible between IFEval and MATH-500, forming a concave frontier in that projection. MBPP+ varies more smoothly across the frontier, suggesting code performance is less in tension with the other two objectives. Overall, the frontier spans a diverse set of Pareto-optimal mixtures, reflecting meaningful flexibility across objective priorities.

\textbf{For the noise model,} hyperparameters such as kernel choice were selected via $k$-fold cross-validation with $k=10$ and the Spearman'$\rho$ metric in \cref{tab:noise-model-ablation}. The final model was trained on all $100$ data samples using the selected hyperparameters: a Laplacian kernel with $\gamma = 0.5$. This model choice was motivated by the limited number of samples, as each data point requires multiple training runs and evaluations, making large-scale collection computationally expensive. 

\begin{table}[t]
\centering
\begin{tabular}{lcccc}
\toprule
Method & CV $R^2$ & CV $\rho$ & Train $R^2$ & Train $\rho$ \\
\midrule
Ridge (Linear)             & 0.269 & 0.518 & 0.339 & 0.579 \\
Ridge (Quadratic)          & \textbf{0.338} & 0.549 & 0.471 & 0.680 \\
\textbf{Kernel ridge (Laplacian)}   & 0.298 & \textbf{0.598} &  \textbf{0.611} &  \textbf{0.826} \\
Kernel ridge (RBF)         & 0.283 & 0.575 & 0.449 & 0.694 \\
Kernel ridge (Quadratic)   & 0.233 & 0.540 & 0.344 & 0.639 \\
GP (Mat\'ern-5/2)          & 0.242 & 0.564 & 0.549 & 0.774 \\
GP (RBF)                   & 0.233 & 0.551 & 0.494 & 0.743 \\
\bottomrule
\end{tabular}
\caption{Noise model ablation to model Math-500 objective's standard deviation on the 6 mixture-proportion features (100 configs, 5 seeds each). Cross-validation (CV) columns report the mean of $R^2$/Spearman $\rho$ over random 10-fold (90/10) splits. Train columns are fit on all 100 configs.}
\label{tab:noise-model-ablation}
\end{table}

\textbf{Visualization of the noise model predictions} are shown in \cref{fig:em_noise} on 3 dimensions. The predicted noise distribution closely mirrors the actual, with higher noise concentrated at low if\_prop1 (stage 1 information following proportion) and high math\_prop1 (stage 1 math proportion) values. The predicted vs. actual plot on the training set also shows points tracking closely along the diagonal, confirming that the noise emulator fits the training data well. Some underestimation is visible at higher std\_math values, which is expected given the limited number of high-noise samples in the $100$-point training set.

\subsection{Emulator validation on real BBO runs}
\label{ap:emulators_bbo_val}

We also validated the emulators in regions where BBO algorithms query, based on the real BBO runs from experiments in \cref{fig:real_emu_rho,fig:HPO_real_emu,fig:DMO_real_emu}. We report the $R^2$, Spearman's $\rho$, and RSME between the real and emulated objectives, of the configurations visited by BBO methods on real LLM runs, in \cref{tab:real_vs_emulated_bbo_fit}. BO methods tend to concentrate their evaluations near the optimum, and scores are often similar, so these points can be hard to differentiate. Even so, the emulators explain 50\% to 68\% of the variance in real scores ($R^2$), rank them with Spearman $\rho$ of $ 0.66$ to $0.76$, and show near-zero bias, with RMSE of at most $0.022$ on DMO and $0.044$ on HPO.

\begin{table}
\centering
\caption{Emulator fit on the points visited by the real BBO runs (DMO: TS, LogNEI, UCB($\beta=2.0$); HPO: TS, LogNEI, TPE). Brackets are 95\% bootstrap CIs.}
\label{tab:real_vs_emulated_bbo_fit}
\begin{tabular}{llrccr}
\toprule
\textbf{Task} & \textbf{Objective} & $n$ & $R^2$ & Spearman $\rho$ & RMSE \\
\midrule
DMO & IFEval & 798 & $0.68$ {\scriptsize$[0.62, 0.72]$} & $0.70$ {\scriptsize$[0.65, 0.74]$} & $0.013$ \\
 & MATH-500 & 798 & $0.64$ {\scriptsize$[0.59, 0.69]$} & $0.70$ {\scriptsize$[0.66, 0.74]$} & $0.022$ \\
 & MBPP+ & 798 & $0.61$ {\scriptsize$[0.56, 0.65]$} & $0.76$ {\scriptsize$[0.72, 0.79]$} & $0.009$ \\
 & Mean & 798 & $0.60$ {\scriptsize$[0.55, 0.63]$} & $0.73$ {\scriptsize$[0.70, 0.77]$} & $0.009$ \\
\midrule
HPO & MATH-500 & 800 & $0.50$ {\scriptsize$[0.40, 0.58]$} & $0.66$ {\scriptsize$[0.61, 0.70]$} & $0.044$ \\
\bottomrule
\end{tabular}
\end{table}

\section{Details on data generation for emulators}

\subsection{Hyperparameter optimization}
\label{ap:hpo}

OpenMathInstruct-2~\citep{toshniwalopenmathinstruct} was selected as the training data. We used only the answer for training, due to time and compute constraint. Long reasoning chains increase the training and evaluation time substantially. Evaluation was similarly done by prompting for only an answer without reasoning, on MATH-500.

We fine-tune from Qwen3 Base models, using 2 H200 for each training run. Checkpoints are saved every 200 steps for evaluation. We evaluate these checkpoints to collect multiple observations for each training run, and used early stopping to increase coverage of the hyperparameter space within a fixed compute budget.

HPO's mixed-variable search space is specified in \cref{tab:hpo_space}.

\begin{table}[t]
    \centering
    \caption{Mixed-variable search space for HPO.}
    \label{tab:hpo_space}
    \begin{tabular}{lll} % l=left, r=right alignment
        \toprule
        \textbf{Parameter} & \textbf{Type} & \textbf{Range} \\
        \midrule
        Learning rate        & Cont. & $[10^{-6}, 10^{-3}]$ normalized to $[0, 1]$ \\
        Per device batch size ($2^n$)   & Int.  & $[2, 4]$ (effective batch size is $2 \times$)\\
        LoRA rank ($2^n$)    & Int.  & $[2, 5]$ \\
        LoRA alpha ($2^n$)   & Int.  & $[2, 5]$ \\
        LoRA dropout ($2^n$) & Cont. & $[0, 0.1]$ normalized to $[0, 1]$ \\
        LoRA layers          & Int.  & $[1, 30]$ \\
        \multirow{4}{*}{LoRA target modules} & \multirow{4}{*}{Cat.} & $0$: \texttt{(q\_proj, v\_proj)} \\
        & & $1$: \texttt{(q\_proj, v\_proj, k\_proj, o\_proj)} \\
        & & $2$: \texttt{(gate\_proj, up\_proj, down\_proj)} \\
        & & $3$: \texttt{(all-linear)} \\
        {[HPO-MF-Cont]} Token fidelity & Cont. & $[10^5, 9.1 \times 10^6]$ normalized to $[0, 1]$ \\
        \multirow{2}{*}{[HPO-MF-Disc] Model fidelity} & \multirow{2}{*}{Cat.} & $0$: HPO-4B emulator \\
        & & $1$: HPO-8B emulator \\
        
        \bottomrule
    \end{tabular}
\end{table}

\subsection{Data mixture optimization}
\label{ap:dmo}

We construct training mixtures from three datasets spanning instruction following, math, and coding: \texttt{tulu-3-sft-personas-instruction-following}, \texttt{tulu-3-sft-personas-math}, and \texttt{tulu-3-sft-personas-code}~\citep{lamberttulu}. Each training run has two stages: stage 1 (tokens $0$-$5$M) and stage 2 (tokens $5$M-$10$M). At each stage, $10$k samples are drawn from the three datasets following a mixing proportion. This roughly corresponds to $5$M tokens or less for each $1$ epoch. Training hyperparameters are: learning rate $10^{-4}$, per device batch size $16$ (effective batch size $32$), lora rank $8$, lora alpha $8$, lora dropout $0.05$, lora layers $30$, and lora target modules \texttt{q\_proj,v\_proj}.

Evaluations were done with lm-eval \citep{eval-harness}. We use the \texttt{inst\_level\_strict\_acc} metric for IFEval, \texttt{math\_verify} metric for MATH-500 (4-shot, minerva format), and \texttt{pass@1} for MBPP+.

\begin{table}[ht]
\centering
\caption{Simplex-constrained search space for DMO.}
\label{tab:dmo_space}
\begin{tabular}{llll}
\toprule
\textbf{Parameter} & \textbf{Type} & \textbf{Range} & \textbf{Constraint} \\
\midrule
$x_1$ & IF proportion (stage 1) & $[0,1]$ & $x_1 \geq 0,\;\sum_{i=1}^{3} x_i = 1$\\
$x_2$ & Math proportion (stage 1) & $[0,1]$ & $x_2 \geq 0,\;\sum_{i=1}^{3} x_i = 1$\\
$x_3$ & Code proportion (stage 1) & $[0,1]$ & $x_3 \geq 0,\;\sum_{i=1}^{3} x_i = 1$\\
$x_4$ & IF proportion (stage 2) & $[0,1]$ & $x_4 \geq 0,\;\sum_{j=4}^{6} x_j = 1$\\
$x_5$ & Math proportion (stage 2) & $[0,1]$ & $x_5 \geq 0,\;\sum_{j=4}^{6} x_j = 1$\\
$x_6$ & Code proportion (stage 2) & $[0,1]$ & $x_6 \geq 0,\;\sum_{j=4}^{6} x_j = 1$\\
\midrule
\multicolumn{4}{l}{$\mathcal{X}_\text{DMO} \triangleq \{ x \in \mathbb{R}^6_{\geq 0} \mid \sum_{i=1}^{3} x_i = 1,\; \sum_{j=4}^{6} x_j = 1 \}$.\; Effective free parameters: 4.} \\
\bottomrule
\end{tabular}
\end{table}

\subsection{Prompt optimization}
\label{ap:po}

\subsubsection{Prompt generation}

Qwen3-14B was used to generate a set of prompts for the candidate set, using multiple meta-prompts to ensure a diverse set of generated prompts. Prompts are embedding using EmbeddingGemma~\citep{embedding_gemma_2025}. Duplicate prompts were removed via exact text match and thresholding on embeddings distance. In addition to generated prompts, common prompts selected by prompt optimization papers in \cref{tab:prompts} were also added to the candidate set. Meta-prompts used to generate the rest of the prompts are shown in \cref{tab:meta_prompts}.

\begin{table}[ht]
\centering
\small
\caption{Prompts from prior prompt optimization methods.}
\label{tab:prompts}
\begin{tabular}{@{}p{0.3\linewidth} p{0.7\linewidth}@{}}
\toprule
\textbf{Method} & \textbf{Prompt} \\
\midrule
Zero-Shot CoT \citep{kojima2022large}
  & ``Let's think step by step.'' \\
  \addlinespace[4pt]
APE \citep{zhou2022large}
  & ``Let's work this out in a step by step way to be sure we have the right answer.'' \\
  \addlinespace[4pt]
OPRO \citep{yang2023large}  & ``Take a deep breath and work through this problem step-by-step.'' \\
  & ``Break this down.'' \\
  & ``A little bit of arithmetic and a logical approach will help us quickly arrive at the solution to this problem.'' \\
  & ``Let's combine our numerical command and clear thinking to quickly and accurately decipher the answer.'' \\
  & ``Let's work through this problem step-by-step:'' \\
  \addlinespace[4pt]
PromptBreeder \citep{fernando2024promptbreeder}
  & ``SOLUTION:'' \\
  \addlinespace[4pt]
ZOPO \citep{hu2024localized}
  & ``Let's find the solution by using the given information.'' \\
  \addlinespace[4pt]
INSTINCT \citep{lin2024use}
  & ``Let's think about it.'' \\
  \addlinespace[4pt]
InstructZero \citep{cheninstructzero}
  & ``Let's use the instruction to solve the problem.'' \\
  \addlinespace[4pt]
Plan-and-Solve+ \citep{wang2023plan}
  & ``Let's first understand the problem, extract relevant variables and their corresponding numerals, and make a complete plan. Then, let's carry out the plan, calculate intermediate variables (pay attention to correct numerical calculation and commonsense), solve the problem step by step, and show the answer.'' \\
\bottomrule
\end{tabular}
\end{table}

\begin{table}[ht]
\small
\centering
\caption{Meta-prompt categories and generation prompts. All meta-prompts follow the template: \textit{``Generate [N] diverse and distinct short instructions (1--4 sentences each) that [DESCRIPTION]. Each instruction should be phrased as a direct directive to the solver. Output one instruction per line with no numbering or extra formatting.''}\ The table shows only the [DESCRIPTION] component and count $N$ for each category.}
\label{tab:meta_prompts}
\resizebox{\textwidth}{!}{%
\begin{tabular}{p{0.15\linewidth}p{0.75\linewidth}r}
\toprule
\textbf{Category} & \textbf{Description} & \textbf{Count} \\
\midrule
Reasoning Style &
tell a math solver to use a specific \textit{reasoning style} (e.g., step-by-step breakdown, chain-of-thought, working backwards, proof by contradiction, elimination of wrong choices, estimation first, drawing a diagram, considering special cases) & 1000 \\[4pt]
Verbosity &
tell a math solver to operate at a specific \textit{level of detail or verbosity} (e.g., give only the final numerical answer, briefly sketch the key steps, show every algebraic manipulation, explain each line as if teaching a beginner, include intermediate checks) & 1000 \\[4pt]
Persona &
adopt a specific \textit{voice or role} for solving math problems (e.g., address the reader as a student being guided, write as a peer collaborator, explain as a strict grader, narrate as if thinking aloud, use the tone of a competition preparation coach) & 1000 \\[4pt]
Format &
specify a \textit{notation or output format} for math solutions (e.g., write all equations in LaTeX, box the final answer, use numbered lines for each step, present results as a table when possible, state assumptions explicitly before solving) & 1000 \\[4pt]
Subject-Specific &
guide a math solver to \textit{identify and leverage the relevant mathematical structure} of a problem before solving (e.g., identify whether the problem is algebraic, geometric, or combinatorial, reframe the problem in the most natural mathematical language, check whether the problem reduces to a known result) & 1000 \\[4pt]
Common Heuristics &
are common, general-purpose \textit{problem-solving heuristics} applicable to any math problem (e.g., think step-by-step, verify your answer at the end, re-read the problem carefully before starting, check your solution against special cases, estimate the answer before computing) & 1000 \\[4pt]
Adversarial &
are \textit{unhelpful or counterproductive} for solving math problems (e.g., vague directives with no clear method, instructions to skip showing work, contradictory guidance, focus on irrelevant topics, overly restrictive rules that prevent flexible reasoning) & 1000 \\
CoT Triggers &
serve as \textit{zero-shot chain-of-thought triggers} --- brief directives that prompt a solver to reason carefully before giving the final answer, as alternatives to ``Let's think step by step'' (e.g., ``Take a deep breath and work through this carefully.'', ``Reason slowly and methodically.'', ``Walk through your logic before concluding.'') & 200 \\[4pt]
CoT Structured &
impose an \textit{explicit reasoning structure} on the solving process (e.g., plan your approach before computing, write out a scratchpad of intermediate values, number each logical sub-step, annotate each line with the rule or property used, decompose the problem into clearly labelled stages) & 200 \\[4pt]
CoT Verification &
ask the solver to \textit{verify or sanity-check} during or after solving (e.g., substitute the answer back into the original equation, estimate the order of magnitude and check against it, consider a simple special case to test your formula, check each step for arithmetic or sign errors) & 200 \\[4pt]
CoT Decomposition &
guide the solver to \textit{decompose the problem} into manageable parts before solving (e.g., identify all given quantities and unknowns, break into sub-problems and solve each in turn, reduce to a simpler analogous case first, separate into cases and handle each independently) & 200 \\[4pt]
CoT Reflection &
encourage \textit{metacognitive self-monitoring} during solving --- prompting the solver to reflect on their own reasoning process (e.g., pause after each step and ask whether it follows logically, flag uncertain steps and revisit them, backtrack if a contradiction is reached, ask ``does this answer make intuitive sense?'') & 200 \\
\bottomrule
\end{tabular}%
}
\end{table}

\subsubsection{Prompt evaluation}

We evaluate candidate instruction prompts on the MATH-500 benchmark~\citep{hendrycks2021measuring} using the \texttt{lm-eval} library~\citep{eval-harness}, on Qwen3-14B in non-thinking mode. The candidate instruction is placed in the system instructions. All evaluations use greedy decoding (temperature $T = 0$), 4-shot prompting, and \texttt{math\_verify} as the scoring metric. We cap generation at $1024$ tokens.  

The best instruction prompt recovered by our BO search is: ``Write the solution in a concise, technical style typical of advanced mathematics.'', achieving $81.0\%$ accuracy on MATH-500. Note that the optimal instruction is specific to our generation configuration, and may differ especially under a different token budget. Our cap of $1024$ tokens is substantially smaller than those used in most frontier model papers, for example, the evaluation results in Qwen3 use a max output length of $32,768$~\citep{yang2025qwen3}. We chose a smaller budget to keep evaluation cost tractable across $5014$ prompt candidates evaluations. Qwen3-14B in non-thinking mode is also able to produce concise, direct answers that fit comfortably within $1024$ tokens.

\subsection{Parallelism configuration optimization}
\label{ap:dto}

In \cref{tab:pco-problem}, we list the training configuration of each problem variant, including the hardware setup, LLM, batch size and sequence length. 

\begin{table}[t]
\centering
\caption{Training setup for each PCO problem variant.}
\label{tab:pco-problem}
\begin{tabular}{lccc}
\toprule
\textbf{Problem variant} & PCO-16 & PCO-32 & PCO-64 \\
\midrule
\textbf{GPU type} & NVIDIA H100 & NVIDIA GH200 & NVIDIA GH200\\
\textbf{\# GPUs} (\texttt{ngpus}) & 16 & 32 & 64 \\
\textbf{\# Nodes} & 2 & 8 & 16 \\
\midrule
\textbf{LLM} & Llama-2-7B & Qwen3-14B & Qwen3-32B \\
\textbf{Batch size} & 512 & 512 & 512 \\
\textbf{Sequence length} & 2048 & 8192 & 8192 \\
\midrule
\textbf{Size of $\mathcal{X}_\text{PCO}$} & 919 & 379 & 780 \\
\bottomrule
\end{tabular}
\end{table}

In \cref{tab:pco-hyperparams}, we list the hyperparameters considered in the parallelism configuration (PC) for the PCO problem. 
Note that we apply logarithm to several of the hyperparameters due to their value generally being considered on an exponential scaling. In practice, we require several constraints to be met for the PC, namely (a) $\texttt{ldp} + \texttt{ltp} + \texttt{lpp} + \texttt{lcp} = \lg \texttt{ngpus}$ so that the GPUs are fully utilized; (b) if $\texttt{ldp} = 0$ then $\texttt{fsdp} = 0$ as a default value; and (c) if $\texttt{lpp} = 0$ then $\texttt{lvp} = 0$ as a default value. Additionally, we apply additional constraints to further prune the candidate space, namely (a) setting $\texttt{ldp} > 0$ and $\texttt{ltp} \leq 2$ to encourage some DP and discourage too much TP; (b) setting $\texttt{lcp} = 1$ for PCO-16 due to shorter sequence length; and (c) setting $\texttt{lacc} = 1$ for PCO-32 and PCO-64. Since the candidate set is pre-generated, these constraints are automatically incorporated into the optimization problem and do not need to explicitly considered when using the benchmark.

\textbf{OOM configurations.} Configurations that run out of memory have no measured throughput. We set $c_\text{PCO}(x) = m_\text{OOM} = 0.2$, and $f_\text{PCO}(x)$ to the throughput of their nearest feasible configuration, or the mean throughput if there are ties.

\begin{table}[t]
\centering
\caption{Hyperparameters in the parallelism configuration for the PCO problem. Here, all logarithms are of base 2, and we use $\lg x$ to denote log base 2 of $x$. Note that \texttt{fsdp} hyperparameter encodes discrete setting for FSDP (which are equivalent to the ZeRO stages); setting \texttt{fsdp} to $0$, $1$, $2$ or $3$ is equivalent to using the flags \texttt{no\_shard}, \texttt{optim}, \texttt{optim\_grads} and \texttt{optim\_grads\_params} for the sharding strategy respectively (however $\texttt{fsdp} = 1$ is omitted). Some of the hyperparameters (e.g., \texttt{lacc}, \texttt{lbsz}) may have larger search ranges, however is restricted in our setting to values that likely give rise to an optimal PC.}
\label{tab:pco-hyperparams}
\begin{tabular}{lll}
\toprule
\textbf{Parameter} & \textbf{Description} & \textbf{Range} \\
\midrule
\texttt{ldp} & Log of data parallelism (DP) size & $[0, \lg \texttt{ngpus}]$ \\
\texttt{ltp} & Log of tensor parallelism (TP) size & $[0, \lg \texttt{ngpus}]$\\
\texttt{lpp} & Log of pipeline parallelism (PP) size & $[0, \lg \texttt{ngpus}]$ \\
\texttt{lcp} & Log of context parallelism (CP) size & $[0, \lg \texttt{ngpus}]$ \\
\texttt{lsp} & Log of sequence parallelism (SP) size & $[0, \texttt{ltp}]$ \\
\midrule
\texttt{lbsz} & Log of \# DP bucket size (in MB) & $\{6, 7, 8, 9\}$ \\
\texttt{fsdp} & FSDP stage & $\{0, 2, 3\}$ \\
\texttt{lvp} & Log number of virtual pipeline chunks for PP & $\{0, 1, 2\}$ \\
\texttt{lacc} & Log of \# of gradient accumulation steps & $\{0, 1, 2\}$ \\
\bottomrule
\end{tabular}
\end{table}

\section{More experiment details}
\label{ap:exp}

Optimal values $f*$ and $\text{HV}^*$ are found via exhaustive evaluation for tabular data, or gradient ascent of $200$ best points found via grid evaluation for MLP emulators. Monte Carlo (MC)-based implementations were used for LogNEI, PES, MES, GIBBON, JES, and KG acquisition functions, with $128$ quasi-MC samples using Sobol sequences per iteration. Initial observations are not included in the regret plots.

\subsection{Empiricial noise levels}
\label{ap:noise}

\begin{table}[h]
\small
\centering
\caption{Empirically measured noise standard deviations $\sigma_N$ for \bolt problems, estimated from $5$ repeated evaluations at matched configurations.}
\label{tab:measured_noise_std}
\begin{tabular}{llcc}
\toprule
\textbf{Family} & \textbf{Problem} & \textbf{Measured $\sigma_N$} & \textbf{\# Configs ($\times 5$ seeds)} \\
\midrule
\multirow{3}{*}{Hyperparameters}
  & HPO          & 0.0290 & 30 \\
  & HPO-MF-Cont  & 0.0220 & 90 \\
  & HPO-MF-Disc  & 0.0290 & 30 \\
\midrule
\multirow{5}{*}{Data mixtures}
  & DMO         & 0.0114 & 100 \\
  & DMO-MO      & (0.0131, 0.0274, 0.0101) & 100 \\
  & \multirow{3}{*}{DMO-Het} & $\sqrt{\sigma_\text{if}^2 + \sigma_\text{math}(x)^2 + \sigma_\text{code}^2}/3$ & \multirow{3}{*}{100} \\
  & & $\sigma_\text{if}=0.0131$, $\sigma_\text{code}=0.0101$ (measured) & \\
  & & $\sigma_\text{math}(x)$ from fitted noise model & \\
\bottomrule
\end{tabular}
\end{table}

Noise levels $\sigma_N$ for the HPO and DMO experiments were empirically found, via $5$ repeated evaluations at matched configurations. Details in \cref{tab:measured_noise_std}. We use empirically measured noise levels to better emulate real LLM problems. 

\subsection{HPO experiments}
\label{ap:hpo_exp}

For the single-fidelity HPO problem, we compare random sampling, Thompson sampling (TS)~\citep{chapelle2011empirical}, log expected improvement (LogEI)~\citep{jones1998efficient, ament2023unexpected}, log noisy expected improvement (LogNEI)~\citep{ament2023unexpected}, upper confidence bounds (UCB)~\citep{srinivas2009gaussian}, predictive entropy search (PES)~\citep{hernandez2014predictive}, max-value entropy search (MES)~\citep{wang2017max}, GIBBON~\citep{moss2021gibbon}, and joint entropy search (JES)~\citep{hvarfner2022joint}. Knowledge gradient (KG)~\citep{frazier2009knowledge} was excluded as it was prohibitively slow for mixed-variable search spaces. These methods were implemented using the BoTorch library~\citep{balandat2020botorch}. 

To handle the mixed-variable space, we use a sum-and-product composite kernel comprising a Hamming-distance kernel for categorical variables and an RBF kernel for continuous and discrete ordinal variables. Acquisition function optimization alternates between local search over discrete and categorical dimensions, and gradient-based optimization over continuous dimensions, with continuous relaxation and rounding applied to high-cardinality discrete variables (e.g., number of LoRA layers, with $30$ discrete values)~\citep{balandat2020botorch}.

We also include dedicated HPO baselines from popular HPO libraries: tree-structured Parzen estimator (TPE)~\citep{bergstra2011algorithms} and covariance matrix adaptation evolution strategy (CMA-ES)~\citep{nomura2021warm,hansen2006cma} from Optuna~\citep{akiba2019optuna} for the single-fidelity setting problem, BOHB~\citep{falkner2018bohb} from SMAC3~\citep{lindauer2022smac3} for the continuous-fidelity setting, and asynchronous successive halving algorithm (ASHA)~\citep{li2020system} from Optuna for the discrete-fidelity setting. Note that CMA-ES cannot handle categorical variables and will fall back to random sampling for the categorical LoRA target module variable.

\textbf{For multi-fidelity problems,} each evaluation consumes budget $\lambda(x) = (1-\gamma)\cdot\mathrm{fid}(x) + \gamma$, normalized so that the highest fidelity costs $1$. We set $\gamma$ from estimated training compute. For HPO-MF-Cont, all fidelities fine-tune the same model (Qwen3-8B), so training FLOPs grow linearly with the number of training tokens, at roughly $30$ GFLOP/token for LoRA fine-tuning (about twice the forward-pass cost). The intercept $\gamma = 0.05$ accounts for per-run fixed overhead (model loading and the MATH-500 evaluation pass), which we estimate at about $5\%$ of a full-fidelity run. For HPO-MF-Disc, both fidelities train on the same token budget and differ only in model size, thus the cost ratio matches the ratio of per-token training FLOPs. Counting non-embedding parameters and the LM head, Qwen3-4B and Qwen3-8B have $4.02$B and $7.57$B parameters, which leads to about $16.1$ and $30.3$ GFLOP/token respectively ($4$ FLOPs per parameter per token), and $\gamma = 16.1/30.3 \approx 0.53$.

We compare MF-MES~\citep{takeno2020multi}, MF-GIBBON, as well as cost-scaled variants of single-fidelity acquisition functions (LogEI, LogNEI, UCB, and PES). In cost-scaled acquisition functions, acquisition values are divided by an affine cost model $c(x) = \alpha \cdot \mathrm{fid}(x) + 1$, where $\mathrm{fid}(x)$ is the fidelity parameter and $\alpha$ controls the cost sensitivity. For log-valued acquisition functions, $\log(c(x))$ is subtracted from the acquisition values instead. For HPO-MF-Cont, we set $\alpha = 0.005$ for NEI, PES, and MF-GIBBON, $\alpha = 0.01$ for EI, and $\alpha = 0.05$ for UCB and MF-MES. For HPO-MF-Disc, we set $\alpha = 0.005$ for PES and MF-GIBBON, $\alpha = 0.01$ for NEI, UCB, and MF-MES, and $\alpha = 0.05$ for EI. A sensitivity analysis over $\alpha$ is in \cref{ap:cost_scale}.

\begin{figure}
    \centering
    
    % Row 1: legends
    \hspace{3pt}
    \begin{subfigure}[t]{0.32\textwidth}
        \centering
        \includegraphics[width=\textwidth]{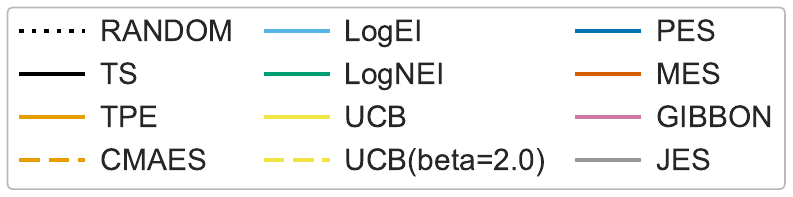}
    \end{subfigure}
    \hfill
    \begin{subfigure}[t]{0.32\textwidth}
        \centering
        \includegraphics[width=\textwidth]{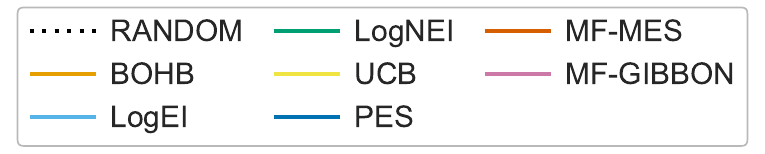}
    \end{subfigure}
    \hfill
    \begin{subfigure}[t]{0.32\textwidth}
        \centering
        \includegraphics[width=\textwidth]{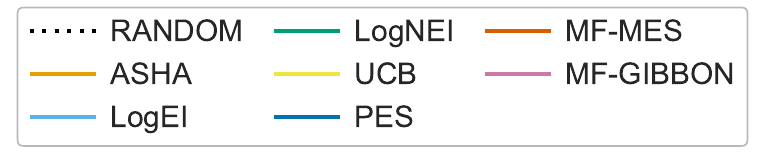}
    \end{subfigure}
    
    % Row 2: three subfigures
    \begin{subfigure}[b]{0.32\textwidth}
        \centering
        \includegraphics[width=\textwidth]{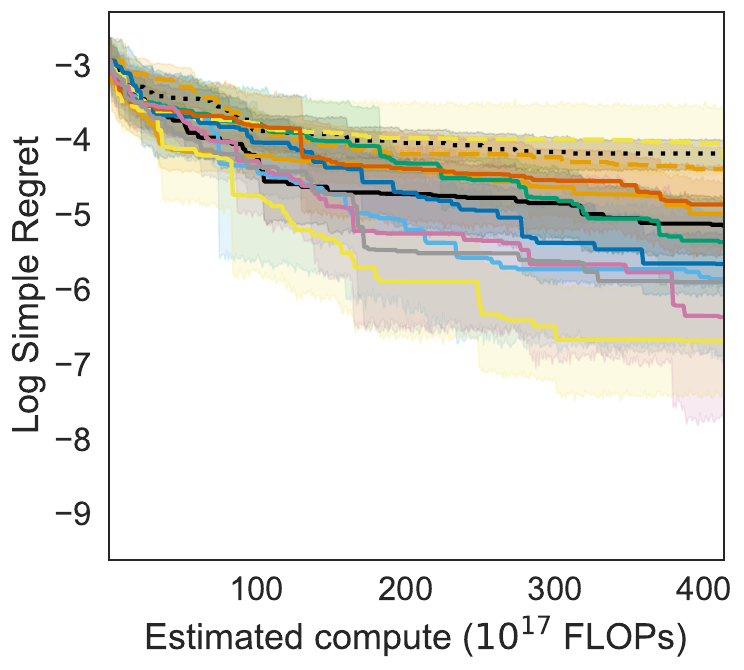}
        \caption{HPO}
    \end{subfigure}
    \hfill
    \begin{subfigure}[b]{0.32\textwidth}
        \centering
        \includegraphics[width=\textwidth]{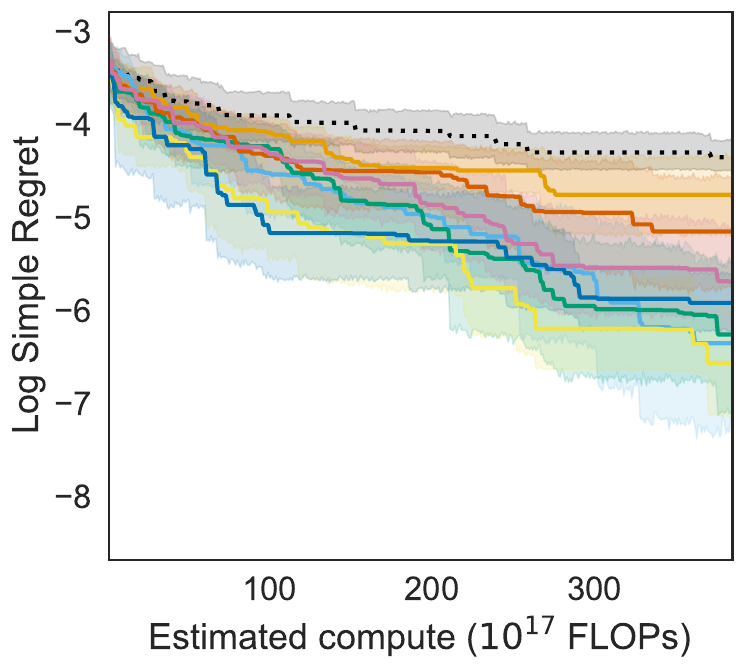}
        \caption{HPO-MF-Cont}
    \end{subfigure}
    \hfill
    \begin{subfigure}[b]{0.32\textwidth}
        \centering
        \includegraphics[width=\textwidth]{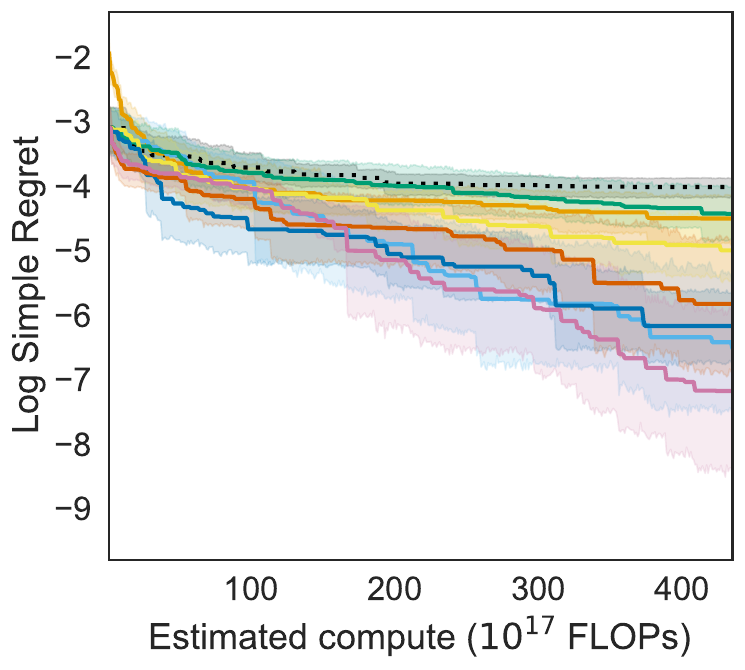}
        \caption{HPO-MF-Disc}
    \end{subfigure}
    \caption{Results on \bolt's HPO problems, against FLOPs}
    \label{fig:HPO_flops}
\end{figure}

\textbf{FLOPs-based results} are additionally shown in \cref{fig:HPO_flops}, where the x-axis is measured against the estimated compute budget rather than BO iterations or MF budget. FLOPs as the budget are common for HPO practitioners. For the FLOPs plots, each query's compute cost is estimated by an analytical estimate of its LoRA fine-tuning FLOPs, which accounts for model size, training tokens and LoRA depth. Unlike the budget $\lambda(x)$, it counts training compute only and excludes fixed overhead such as model loading and evaluation. The differences are minor and the analysis on the simple regret plots measured against BO iterations hold.

\subsection{Discussion on UCB's exploration parameter}
\label{ap:ucb}

\begin{figure}
    \centering
    % Row 1: legend only
    \begin{subfigure}[t]{\textwidth}
        \centering
        \includegraphics[width=0.5\textwidth]{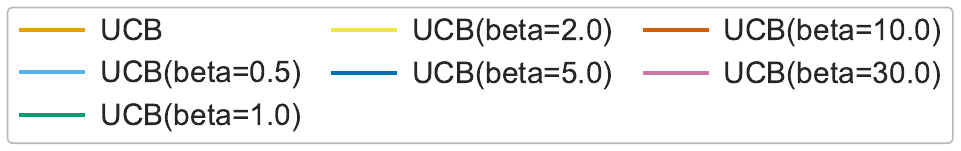}
    \end{subfigure}
    % Row 2
    \begin{subfigure}[t]{0.35\textwidth}
        \centering
        \includegraphics[width=\textwidth]{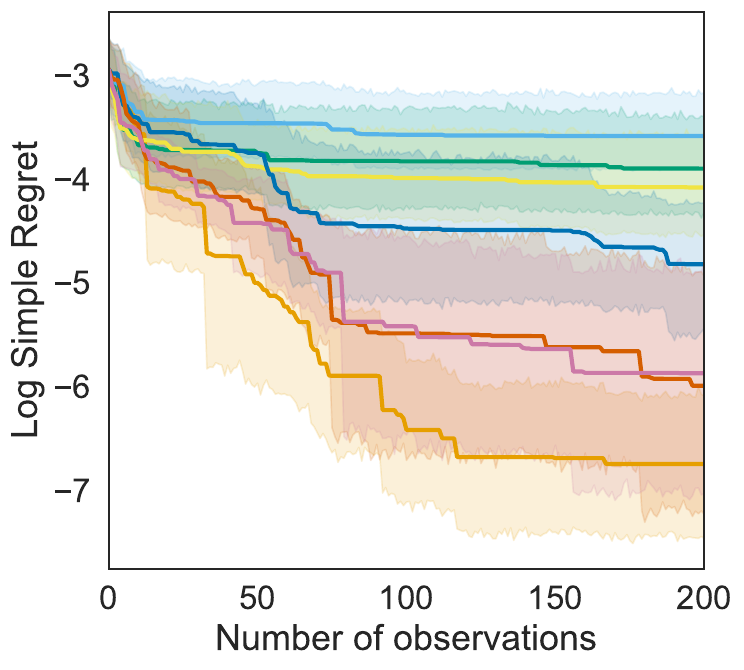}
        \caption{HPO}
    \end{subfigure}
    \begin{subfigure}[t]{0.35\textwidth}
        \centering
        \includegraphics[width=\textwidth]{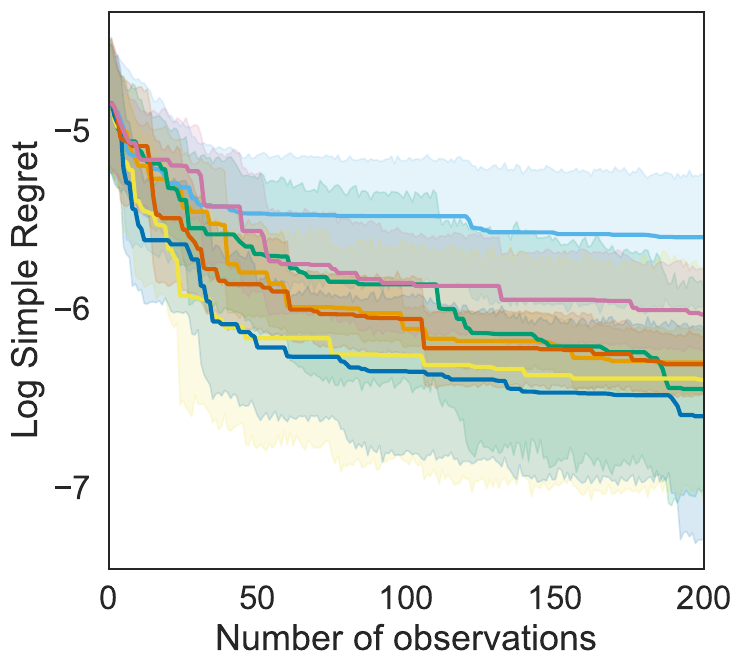}
        \caption{DMO}
    \end{subfigure}
    \caption{Comparison of $\beta$ for UCB on HPO and DMO base problems. "UCB" refers to using scheduled $\beta_t$ from \citet{srinivas2009gaussian}}
    \label{fig:ucb}
\end{figure}

\cref{fig:ucb} compares UCB with a fixed $\beta$ against the scheduled $\beta_t$ \citep{srinivas2009gaussian}. Our implementation differs slightly, where we use the input dimension $d$ in place of the number of candidate points for $\beta_t$, which makes it slightly smaller.

A fixed $\beta$ allows direct control over the exploration-exploitation tradeoff and thus the convergence behavior. Smaller $\beta$ reduces exploration, which can speed up early progress but tends to stall at worse solutions in both problems, as seen for $\beta=1.0$ in both HPO and DMO plots. Conversely, larger $\beta$ values converge more slowly but tend to yield better final solutions. This suggests that if a fixed $\beta$ is preferred, $\beta$ should be chosen with the compute budget in mind, to balance exploration and exploitation effectively.

On HPO, the schedule outperforms every fixed $\beta$, while on DMO, differences are small and a moderate fixed $\beta$ ($2$--$5$) is competitive. We include UCB ($\beta=2$) in the main regret plots for reference.  $\beta$ of 2 is smaller than scheduled $\beta_t$ so it tends to be better in early iterations.

\subsection{Cost scale sensitivity}
\label{ap:cost_scale}

\begin{figure}
    \centering
    % Row 1
    \begin{subfigure}[t]{\textwidth}
        \centering
        \includegraphics[width=\textwidth]{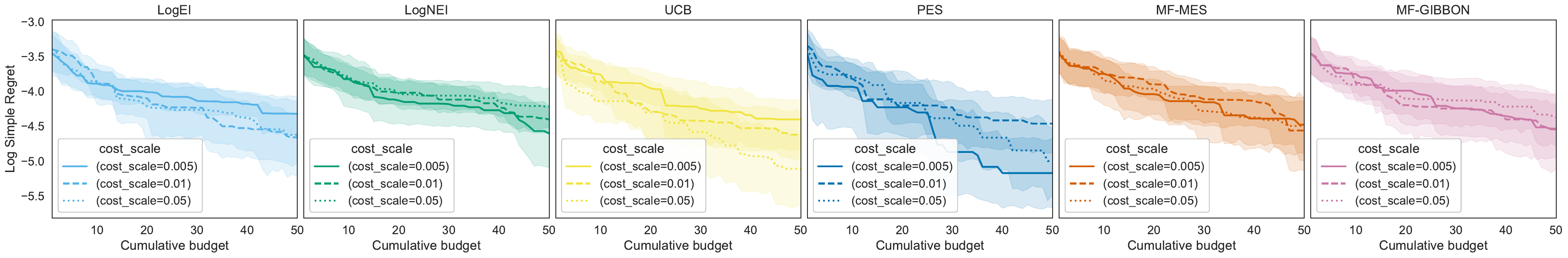}
        \caption{HPO-MF-Cont}
    \end{subfigure}
    
    % Row 2
    \begin{subfigure}[t]{\textwidth}
        \centering
        \includegraphics[width=\textwidth]{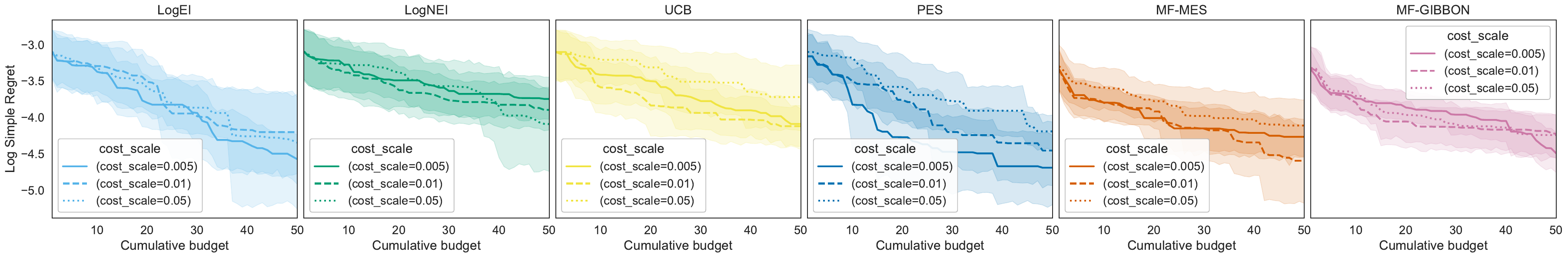}
        \caption{HPO-MF-Disc}
    \end{subfigure}
    \caption{Performance on MF HPO problems on different cost scale.}
    \label{fig:cost_scale}
\end{figure}

In cost-scaled acquisition functions, acquisition values are divided by an affine cost model $c(x) = \alpha \cdot \mathrm{fid}(x) + 1$, where $\mathrm{fid}(x)$ is the fidelity parameter and $\alpha$ controls the cost sensitivity. Larger $\alpha$ penalizes high-fidelity queries more heavily in their acquisition scores, encouraging more low-fidelity exploration. We evaluate sensitivity to the cost scale parameter $\alpha$, comparing $\alpha \in \{0.005, 0.01, 0.05\}$ on both multi-fidelity HPO variants (continuous token-based fidelity and discrete model-size fidelity) in \cref{fig:cost_scale}. The degree of sensitivity varies across acquisition functions and problems. The effect of cost scale $\alpha$ is method-dependent, and no single value is best across methods. On the continuous-fidelity problem (HPO-MF-Cont), PES and UCB are the most sensitive, and at their best $\alpha$ they reach the lowest final regret of all methods. Their preferred values differ, however, with $\alpha=0.005$ for PES and $\alpha=0.05$ for UCB. On the discrete-fidelity problem (HPO-MF-Disc), all methods show moderate sensitivity, with differences across $\alpha$ largely within the confidence intervals. Since the best cost scale varies by method and problem, $\alpha$ should be treated as a hyperparameter to tune rather than fixed a priori.

\subsection{Fidelity proportions over iterations}
\label{ap:fidelity}

\begin{figure}[t]
    \centering
    % Row 1
    \begin{subfigure}[t]{0.9\textwidth}
        \centering
        \includegraphics[width=\textwidth]{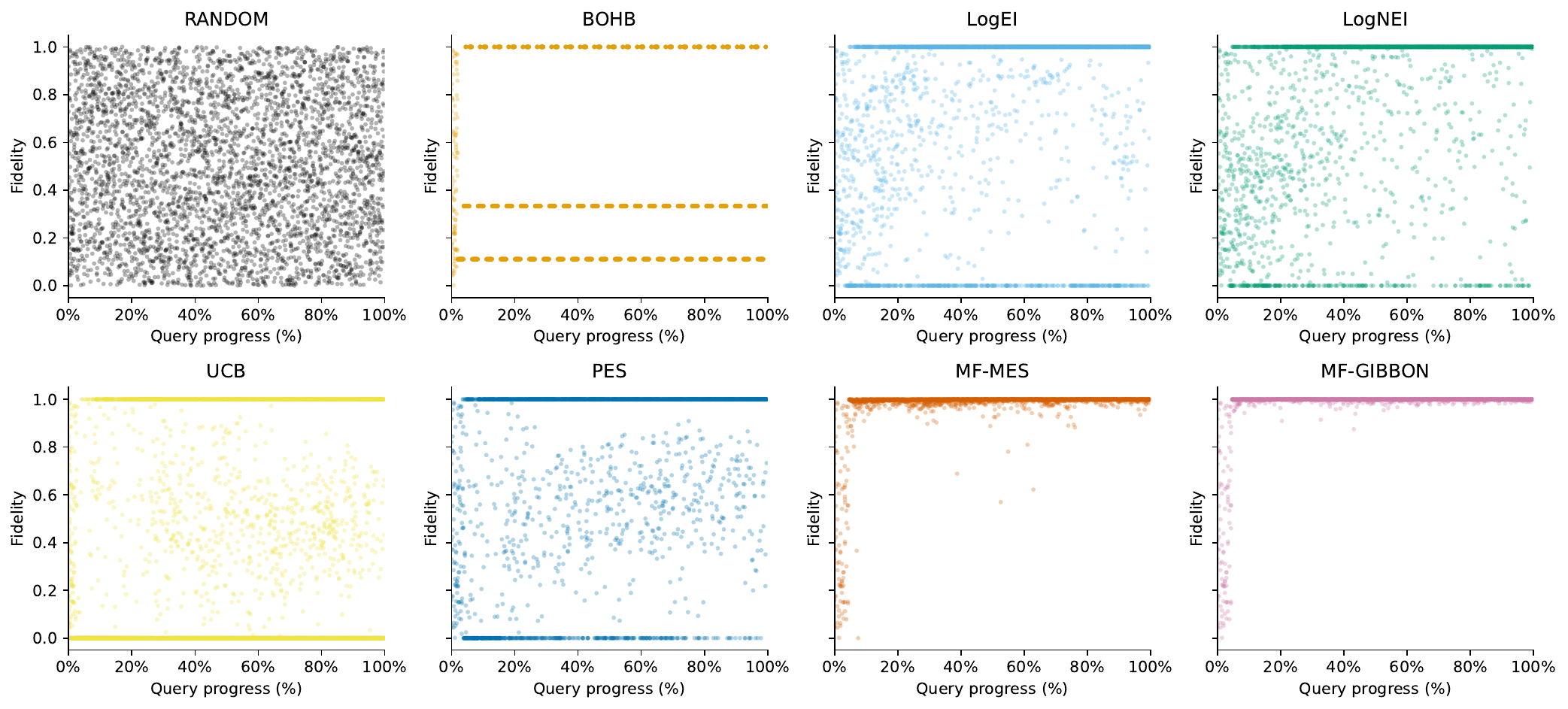}
        \caption{HPO-MF-Cont}
    \end{subfigure}

    % Row 2
    \begin{subfigure}[t]{0.9\textwidth}
        \centering
        \includegraphics[width=\textwidth]{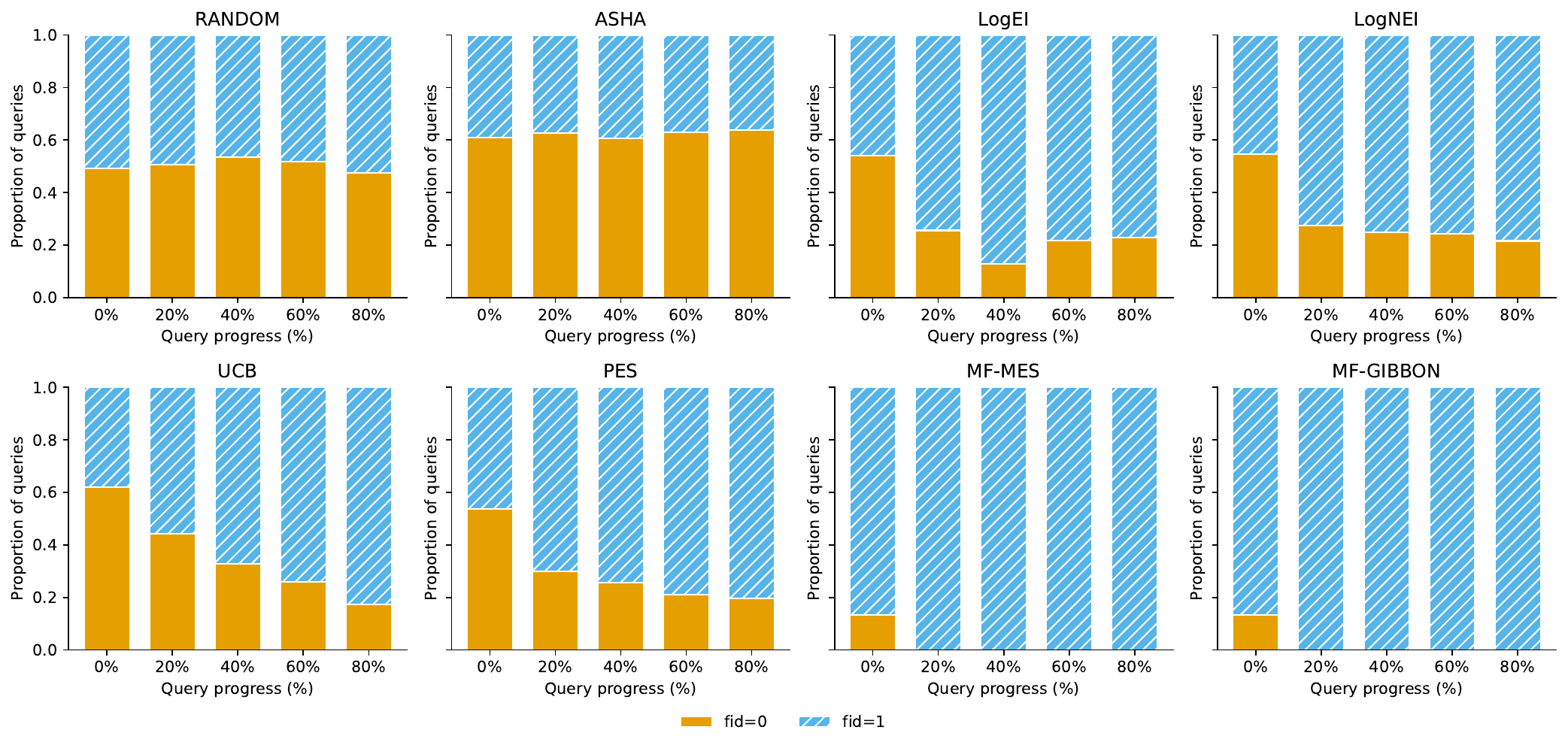}
        \caption{HPO-MF-Disc}
    \end{subfigure}
    \caption{Fidelity proportions of observations on MF HPO problems.}
    \label{fig:fid}
\end{figure}

Figure~\ref{fig:fid} shows how each method allocates queries across fidelity levels over the course of optimization. Recall the acquisition functions are all cost scaled. For HPO-MF-Cont, $\alpha = 0.005$ for NEI, PES, and MF-GIBBON, $\alpha = 0.01$ for EI, and $\alpha = 0.05$ for UCB and MF-MES. For HPO-MF-Disc, $\alpha = 0.005$ for PES and MF-GIBBON, $\alpha = 0.01$ for NEI, UCB, and MF-MES, and $\alpha = 0.05$ for EI.

For the continuous-fidelity setting (HPO-MF-Cont), BOHB concentrates queries at a few discrete fidelity levels set by its bracket schedule. MF-MES and MF-GIBBON query almost exclusively at fidelity $1$ after initialization. In contrast, cost-scaled LogEI, LogNEI, UCB, and PES maintain a persistent cluster at the lowest fidelity alongside scattered intermediate-fidelity queries, most pronounced for UCB. These mixed-fidelity methods achieve the lowest regret on this task (\cref{fig:hpo_mf_c_res}).

For the discrete-fidelity setting (HPO-MF-Disc), cost-scaled acquisition functions shift toward high-fidelity queries as optimization progresses, while MF-MES and MF-GIBBON query only the high fidelity after initialization. PES achieves the lowest regret, followed by MF-GIBBON (\cref{fig:hpo_mf_d_res}).

Given that methods with vastly different fidelity allocation strategies can perform well, these results suggest a complex relationship between fidelity allocation and final optimization quality, one that likely depends on the specific configurations evaluated at each fidelity level rather than the allocation pattern alone.

\subsection{DMO experiments}
\label{ap:dmo_exps}

For the single-objective problems, we compare random sampling, TS, LogEI, LogNEI, KG, UCB, PES, MES, GIBBON, and JES. For the multi-objective problem, we benchmarked TSEMO~\citep{bradford2018efficient}, NSGA2~\citep{deb2002fast}, ParEGO~\citep{knowles2006parego},  NEHVI~\citep{daulton2021parallel}, PESMO~\citep{hernandez2016predictive}
MESMO~\citep{belakaria2019max}, and JESMO~\citep{tu2022joint}. 

Information-theoretic acquisition functions (PES, MES, GIBBON, JES, and their multi-objective variants) require approximations of the global optimum or Pareto-optimal set. Since standard box-constrained sampling cannot satisfy the simplex equality constraints of the data-mixture search space, we instead draw feasible candidates via symmetric Dirichlet sampling and select approximate optima by Thompson sampling and taking the argmax of each posterior draw over the candidate set. For JES, optimal inputs and values are drawn from the same posterior sample to ensure consistency. For multi-objective entropy-search methods, we draw multiple joint posterior samples over feasible candidates, extract the non-dominated set from each sample as an approximate Pareto front.     

The reference HV point $\text{HV}^*$ was set per objective as $10\%$ below the observed minimum, with slack scaled to each objective's range. Minima were computed from surrogate predictions over a dense regular grid on the mixture-ratio simplex.

\subsection{MLHGP noise analysis}
\label{ap:mlhgp_noise}

\begin{figure}
    \centering
    \includegraphics[width=\textwidth]{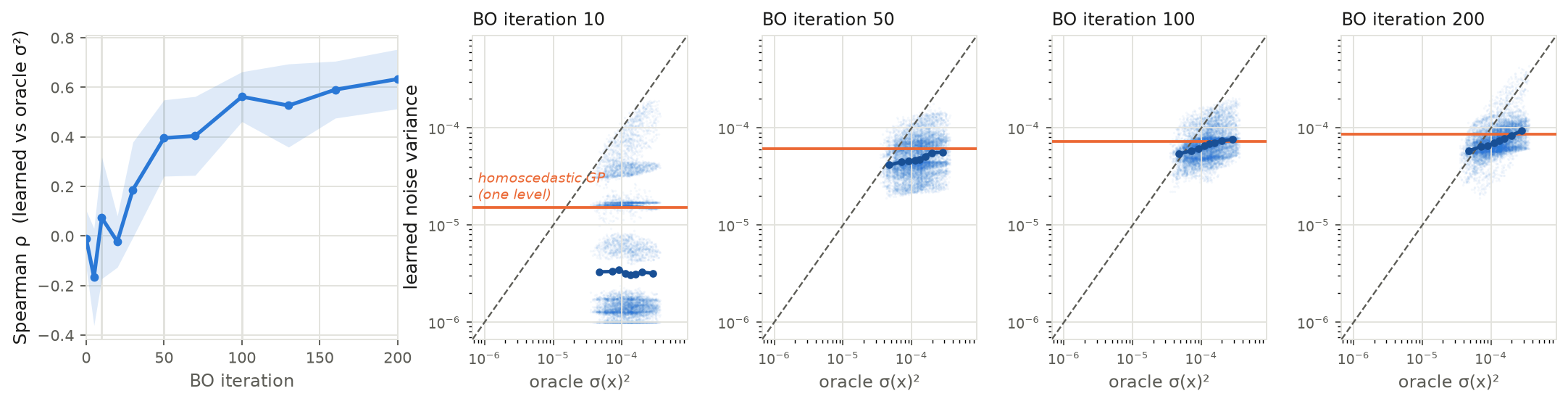}
    \caption{MLHGP noise model over BO iterations}
    \label{fig:mlhgp_noise}
\end{figure}

We evaluate the fit of MLHGP's learned noise model over BO iterations in \cref{fig:mlhgp_noise}, using a held-out set of $1000$ sampled test points. The noise model's Spearman $\rho$ against the oracle noise variance improves steadily and becomes reasonably accurate by around iteration $100$. This is consistent with LogNEI+MLHGP overtaking LogNEI in regret at roughly the same point. Unlike a homoscedastic noise estimate, which fits a single noise level everywhere, MLHGP correctly captures the varying noise levels across the input space. It does, however, tend to underestimate the noise magnitude, which likely pushes LogNEI toward more exploitative behavior.

\subsection{PO experiments}
\label{ap:po_exps}

For PO experiments, we compare random sampling, TS, LogEI, LogNEI, UCB, and GIBBON, as well as batched, dTuRBO, and dBAxUS variants with LogNEI. dTuRBO and dBAxUS are our discretized adaptations of TuRBO and BAxUS.  We omit SAASBO~\citep{eriksson2021high} due to its prohibitive cost, and MSR~\citep{papenmeier2025understanding} as it does not apply to discretized spaces. 

TuRBO searches within trust-regions for scalability, expanding and contracting trust-region hyperrectangles upon successive improvements and failures respectively, while BAxUS searches in reduced-dimensional subspaces via random projection and applies trust-region filtering within that subspace. TuRBO and BAxUS are originally designed for continuous spaces, particularly in their trust-region methodology. 

We adapted TuRBO and BAxUS to discretized search spaces by representing the trust regions as the $k$ nearest neighbors of the current best solution, where $k$ is doubled after $\tau_\text{succ}$ consecutive improvements and halved after $\tau_\text{fail}$ consecutive non-improvements. The trust region is initialized with $k_\text{init} = \lfloor N/2 \rfloor$ candidates, where $N = 5014$ is the total prompt pool size. Both methods use $\tau_\text{succ} = 3$, dTuRBO uses a fixed $\tau_\text{fail} = 10$, while for dBAxUS, $\tau_\text{fail}$ is set dynamically per subspace expansion stage, calibrated so that the number of failures required to fully contract the trust region matches the per-stage evaluation budget, following the continuous-space contraction rate in BAxUS. Sobol candidate generation and sparse perturbation are omitted in the discretized adaptions.

We show how selected enhancements (batch size, dTuRBO, dBAxUS) affect the simple regret at iteration $200$ in \cref{tab:po_deltas}.

\begin{table}[h]
\centering
\caption{Delta in log simple regret at iteration $200$ vs LogNEI baseline (mean $\pm$ std over $5$ seeds). Negative is better as it indicates lower log regret.}
\label{tab:po_deltas}

\begin{tabular}{lrrrr}
\toprule
\textbf{Enhancement} & \textbf{PO128} & \textbf{PO256} & \textbf{PO512} & \textbf{PO768} \\
\midrule
\texttt{q=5} & $-8.67 \pm 7.75$ & $-8.23 \pm 7.61$ & $-8.40 \pm 7.54$ & $-8.85 \pm 7.67$ \\
\texttt{TuRBO} & $-5.87 \pm 7.55$ & $-10.61 \pm 7.10$ & $-5.12 \pm 7.78$ & $-5.67 \pm 7.67$ \\
\texttt{TuRBO+q=5} & $\mathbf{-14.05 \pm 0.63}$$^\dagger$ & $\mathbf{-13.61 \pm 0.46}$$^\dagger$ & $\mathbf{-13.78 \pm 0.41}$$^\dagger$ & $\mathbf{-14.09 \pm 0.87}$$^\dagger$ \\
\texttt{BAxUS} & $-0.43 \pm 1.09$ & $-2.27 \pm 6.57$ & $-10.82 \pm 6.72$ & $-5.99 \pm 7.38$ \\
\texttt{BAxUS+q=5} & $-8.61 \pm 6.88$ & $\mathbf{-13.61 \pm 0.46}$$^\dagger$ & $\mathbf{-13.78 \pm 0.41}$$^\dagger$ & $\mathbf{-14.09 \pm 0.87}$$^\dagger$ \\
% \vspace{0.5em}
\addlinespace
\multicolumn{5}{l}{\footnotesize $^\dagger$ All 5 seeds hit the log-regret floor ($\approx -18.42$); deltas coincide when methods fully floor out.} \\

\bottomrule
\end{tabular}
\end{table}

\subsection{PCO experiments}
\label{ap:pco_exps}

We evaluated constrained BO methods: (Log)EIC~\citep{pmlr-v32-gardner14}, (Log)NEIC~\citep{letham2019cnei}, UCB-C~\citep{nguyen2024optimistic}, dSCBO~\citep{eriksson2021scalable}, cTS, CMES-IBO~\citep{takeno2022sequential}, cKG~\citep{ungredda2024bayesian},  ADMMBO~\citep{ariafar2019admmbo}, and BBO baselines (TPE, CMA-ES). Log variants for EIC methods are used for numerical stability \citep{ament2023unexpected}. For LogNEIC, UCB-C, cKG, and ADMMBO, we include a variant with GP prior means formulated from parallelism configuration knowledge \citep{oppa}. UCB-C, CMES-IBO, cKG, and ADMMBO were implemented on BoTorch's acquisition interface~\citep{balandat2020botorch}, and cTS via BoTorch's \texttt{ConstrainedMaxPosteriorSampling}.

All acquisition functions are maximized over the discretized candidate set, and previously evaluated configurations remain in the pool. For CMES-IBO, constrained maxima are sampled directly from joint posterior samples over the candidate set rather than via random Fourier features. For cKG, inner maximizations are computed exactly over the candidate set. For TPE and CMA-ES, which search the integer bounding box, each proposal is mapped to the nearest valid configuration.

dSCBO is our discretized adaptation of SCBO, which extends cTS with a trust region. As with dTuRBO in PO experiments (\cref{ap:po_exps}), the trust region is represented by the $k$ nearest neighbors of the incumbent, with $k_\text{init} = \lfloor N/2 \rfloor$, $\tau_\text{succ} = 3$, and $\tau_\text{fail} = 10$. Distances are weighted by the objective GP's ARD lengthscales, following SCBO's trust-region shaping.

For LogNEIC, UCB-C, cKG, and ADMMBO, we include a variant that replaces the GPs' constant prior means with analytic cost models of parallel training~\citep{oppa}. The throughput mean models communication time for DP, TP, and PP (with separate intra- and inter-host latency and bandwidth) plus computation time, and the memory mean models parameter, activation, and overhead memory. Their parameters are learned jointly with the kernel hyperparameters. Following \citet{oppa}, the throughput mean is blended with a constant whose weight increases to one by iteration $20$, so the prior guides early fits and gives way as data accumulates.

Each method gets $10$ random initial points followed by $50$ (PCO-32) or $100$ (PCO-16, PCO-64) iterations, over $10$ seeds. The different number of iterations are due to the different number of data points for each problem. PCO-32 has fewer points ($379$) than PCO-16 ($919$) and PCO-64 ($780$).

\subsection{PCO landscapes}
\label{ap:PCO_landscape}

\begin{figure}
    \centering
    \includegraphics[width=\textwidth]{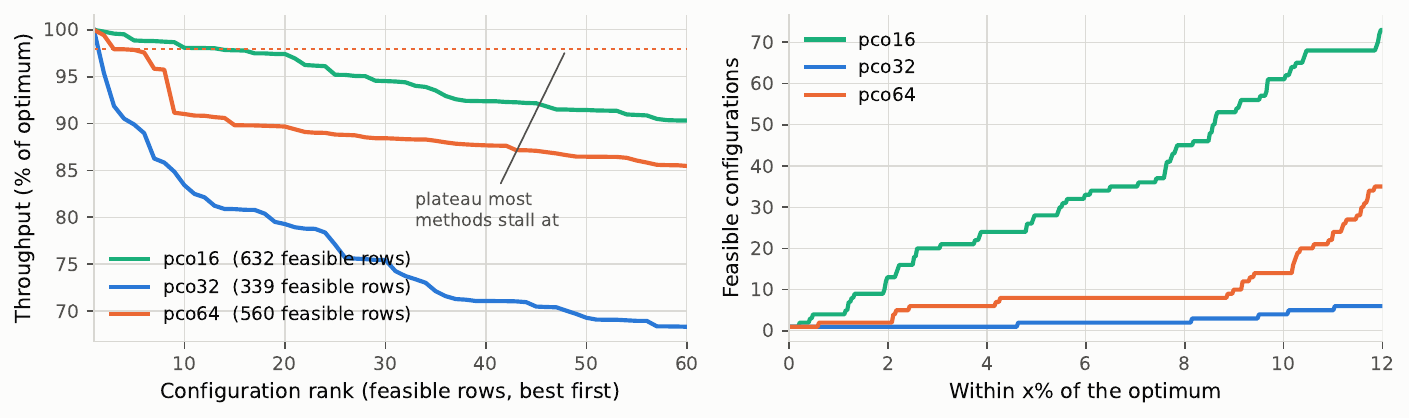}
    \caption{PCO's landscape which shows a plateau near the optimum for PCO-16 and PCO-64.}
    \label{fig:pco_plateau}
\end{figure}

\begin{figure}
    \centering
    \begin{subfigure}[t]{0.8\textwidth}
        \centering
        \includegraphics[width=\textwidth]{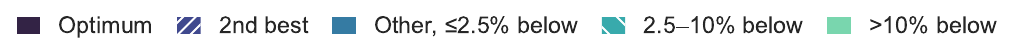}
    \end{subfigure}
    
    \begin{subfigure}[t]{0.32\textwidth}
        \centering
        \includegraphics[width=\textwidth]{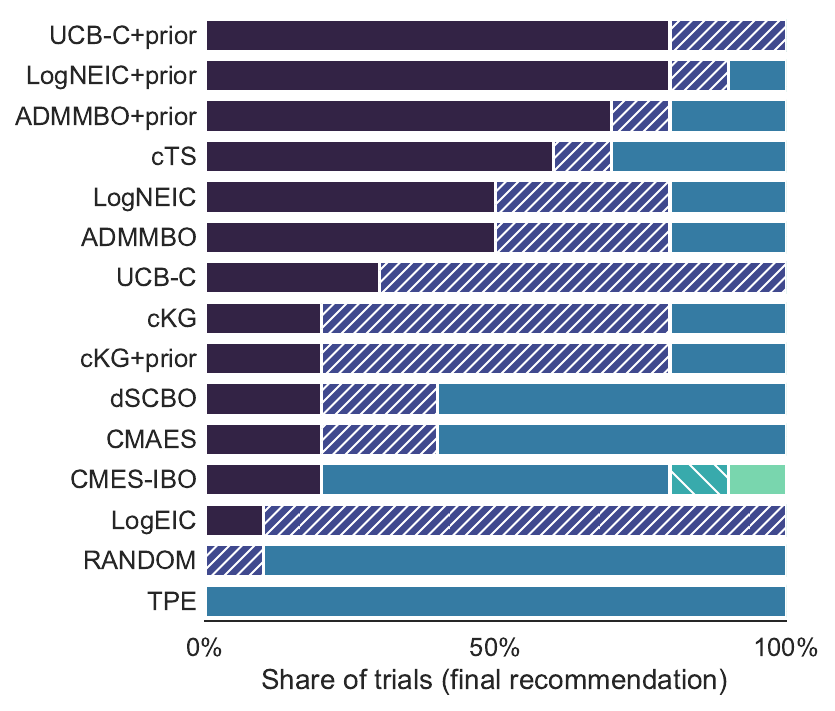}
        \caption{PCO-16 at iteration 100}
    \end{subfigure}
    \hfill
    \begin{subfigure}[t]{0.32\textwidth}
        \centering
        \includegraphics[width=\textwidth]{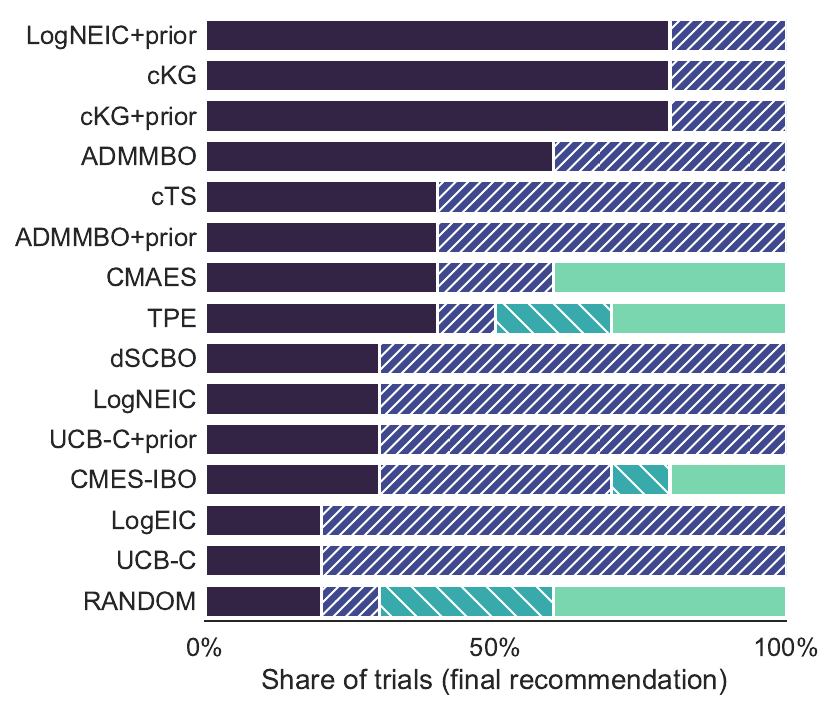}
        \caption{PCO-32 at iteration $50$}
    \end{subfigure}
    \hfill
    \begin{subfigure}[t]{0.32\textwidth}
        \centering
        \includegraphics[width=\textwidth]{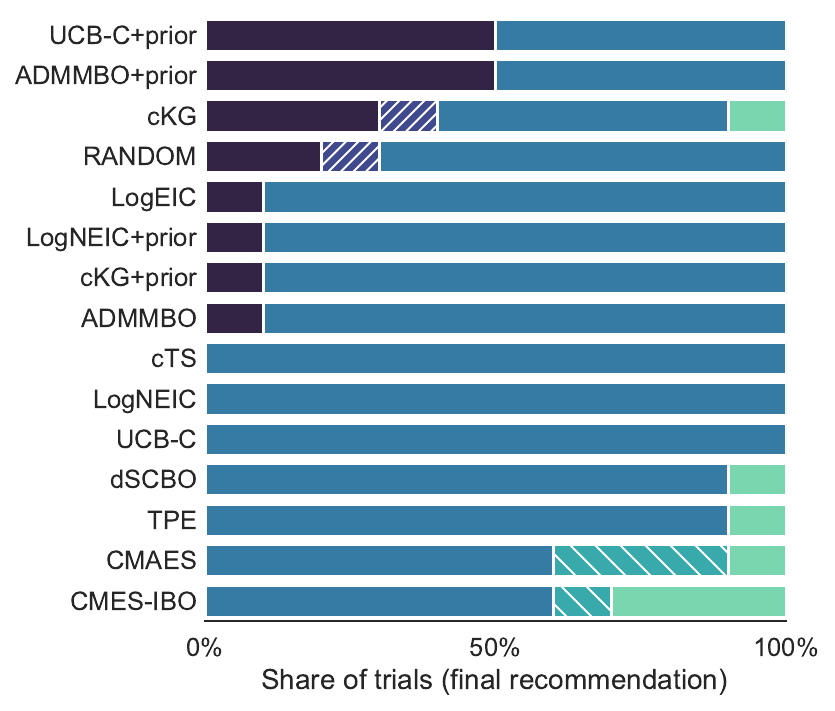}
        \caption{PCO-64 at iteration $100$}
    \end{subfigure}
    \caption{Quality of final recommendations across $10$ seeds, grouped by their gap to the optimum.}
    \label{fig:PCO_exp_outcomes}
\end{figure}

The problem landscapes are shown in \cref{fig:pco_plateau}. PCO-32 has a single sharp optimum, and its second-best configuration is about $4.6\%$ below the optimum. PCO-16 and PCO-64 instead have flat near-optimal regions, with $16$ of $632$ and $6$ of $560$ feasible configurations within $2.5\%$ of the optimum, respectively. On PCO-64, the second-best configuration is only about $0.6\%$ below the optimum. Once a BO method reaches such a plateau, the small throughput differences within it can lead to over-exploitation.

\Cref{fig:PCO_exp_outcomes} supports this. On PCO-32, all BO methods except CMES-IBO recommend the optimum or second-best configuration in every seed, with LogNEIC+prior, cKG, and cKG+prior each finding the optimum in $8$ of $10$ seeds. In contrast, random search, TPE, CMA-ES, and CMES-IBO end more than $10\%$ below the optimum in $2$--$4$ of $10$ seeds. On PCO-16 and PCO-64, nearly all final recommendations lie within $2.5\%$ of the optimum, and many methods reach the plateau but not the peak. On PCO-64, UCB-C+prior and ADMMBO+prior find the optimum in $5$ of $10$ seeds, cKG in $3$, and random search in $2$, while cTS, LogNEIC, UCB-C, dSCBO, TPE, CMA-ES, and CMES-IBO never do. On PCO-16, UCB-C+prior and LogNEIC+prior find it in $8$ of $10$ seeds, while random search and TPE never do.

The prior helps most on the plateau problems, where it improves several methods. We hypothesize that its analytic throughput mean (\cref{ap:pco_exps}) separates near-optimal configurations whose throughputs differ too little for a constant-mean GP to distinguish from few observations. In contrast, cKG, which values queries by their effect on the final recommendation, performs best on the sharp-optimum PCO-32 but gains little from the prior on the plateaus. Overall, no method is robust across all three landscapes, and the plateaus in PCO-16 and PCO-64 expose a failure mode of current methods: they reach near-optimal regions but under-explore within them.

CMES-IBO and dSCBO are less suited to PCO's discretized candidate set and integer search space as they were designed for the continuous settings. CMES-IBO's acquisition function, a lower bound on the information gain, remains positive at observed points. This is harmless in continuous spaces, where exact repeats are unlikely, but over a discretized candidate set it leads to frequent re-querying ($86\%$, $81\%$, and $87\%$ of the budget on PCO-16, PCO-32, and PCO-64). dSCBO performs worse than cTS, which it extends with a trust region, most clearly on PCO-16. With our $k$-NN adaptation (\cref{ap:pco_exps}), the region is always non-empty, but two properties of PCO limit its benefit. First, the candidate set is small enough for cTS to search globally, so localizing search adds little benefit. Second, on a coarse integer lattice, the nearest configurations can still differ substantially in throughput, weakening the locality that trust regions rely on.

\subsection{Simple regret vs normalized regret}
\label{ap:regrets}

We compare the stability of simple regret and normalized regret across seeds on the HPO emulator. Normalized regret was used to compare real and emulated BBO results in \cref{fig:HPO_real_emu,fig:DMO_real_emu} as it is based only on observed values. Formally, normalized regret is defined as $(y_\text{best} - y_\text{min}) / (y_\text{max} - y_\text{min})$, where $y_\text{min}$ and $y_\text{max}$ are the min and max of the best observed values across all iterations and methods. Methods are ranked by $\max_t y_t$, the best \emph{observed} value, where $y_t = f(x_t) + \varepsilon_t$. A maximum over noisy draws selects for large positive $\varepsilon$, so the score carries an optimism bias even after averaging over BO seeds. On our HPO emulator, this overshoot can exceed the true performance gap between methods, and cause noise to influence the ranking.

\textbf{Simple regret is the more stable ranking.} Splitting the seeds into disjoint halves ($5$/$5$ seeds) and ranking the methods within each half, simple regret agrees across halves far more often than normalized regret does, as seen in \cref{tab:regret-metric-quality}. We therefore report simple regret wherever the noiseless objective is available, and normalized regret only on the real runs.

\begin{table}[t]
\centering
\small
\begin{tabular}{lccc}
\toprule
& \multicolumn{3}{c}{split-half rank stability (Spearman $\rho$)} \\
\cmidrule(lr){2-4}
\textbf{Criterion} & mean & median & frac.\ $<0$ \\
\midrule
normalized regret & $0.505$ & $0.518$ & $0.01$ \\
simple regret     & $\mathbf{0.709}$ & $0.718$ & $0.00$\\
\bottomrule
\end{tabular}
\caption{Ranking quality of simple vs.\ normalized regret on the HPO emulator.}
\label{tab:regret-metric-quality}
\end{table}

\subsection{More evaluation metrics}

In \cref{sec:exps}, for single-objective problems, we evaluate performance using simple regret, which is defined as $f^* - \max_{t' \leq t} f(x_{t'})$. While simple regret provides a clean picture of optimization progress over the budget $t$, it requires access to the noiseless $f$ in their $\max$ term, which is computable on benchmarks but unavailable in practice.

Another popular evaluation metric is inference regret, defined as $f^* - f(\hat{x}_t)$, where $\hat{x}_t \triangleq \arg\max_{x \in \mathcal{X}} \mu_t(x)$ is the point that maximizes the posterior mean at iteration $t$. Inference regret is commonly used in entropy search-based methods~\citep{wang2017max}, and it utilizes the surrogate model in estimating the optimal point. We also report the \emph{per-step} variant of simple regret $f^* - f(\tilde{x}_t)$, where $\tilde{x}_t \triangleq \arg\max_{t' \leq t} y_{t'}$ is the best (noisy) observed point. Both metrics are not always monotonically decreasing, but matches practical evaluation conditions more closely.

\begin{figure}
    \centering
    
    % Row 1: two subfigures, second spans remaining width
    \begin{subfigure}[t]{0.32\textwidth}
        \centering
        \includegraphics[width=\textwidth]{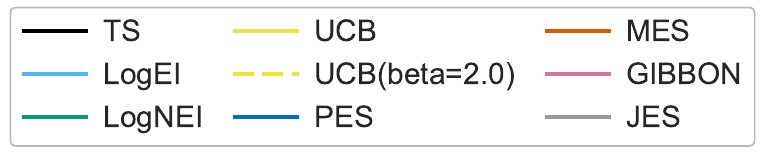}
    \end{subfigure}
    \hfill
    \begin{subfigure}[t]{0.32\textwidth}
        \centering
        \includegraphics[width=\textwidth]{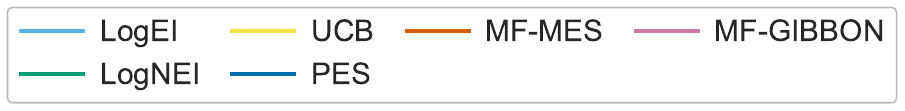}
    \end{subfigure}
    \hfill
    \begin{subfigure}[t]{0.32\textwidth}
        \centering
        \includegraphics[width=\textwidth]{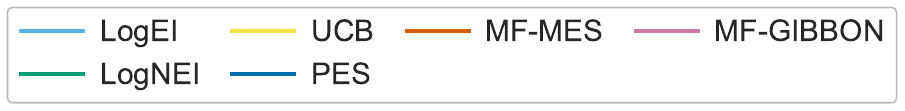}
    \end{subfigure}
    
    % Row 2: three subfigures
    \begin{subfigure}[t]{0.32\textwidth}
        \centering
        \includegraphics[width=\textwidth]{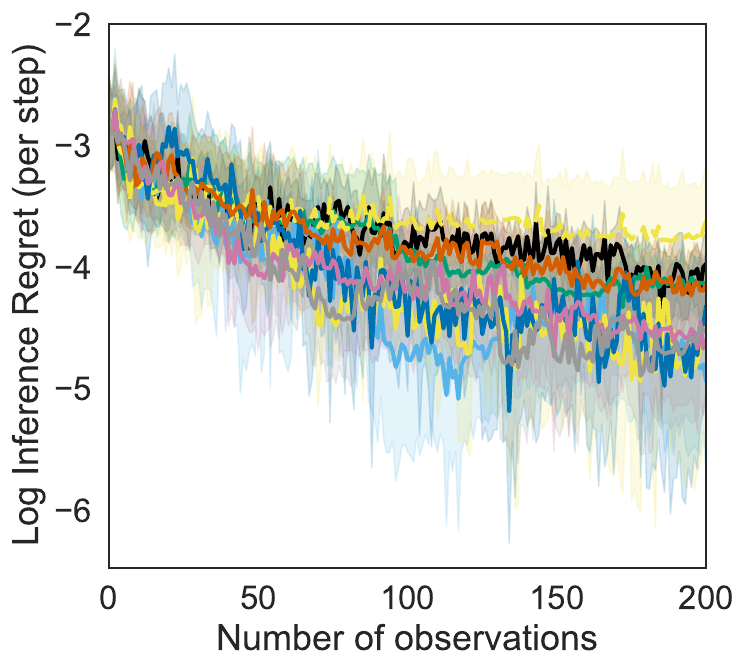}
        \caption{HPO}
    \end{subfigure}
    \hfill
    \begin{subfigure}[t]{0.32\textwidth}
        \centering
        \includegraphics[width=\textwidth]{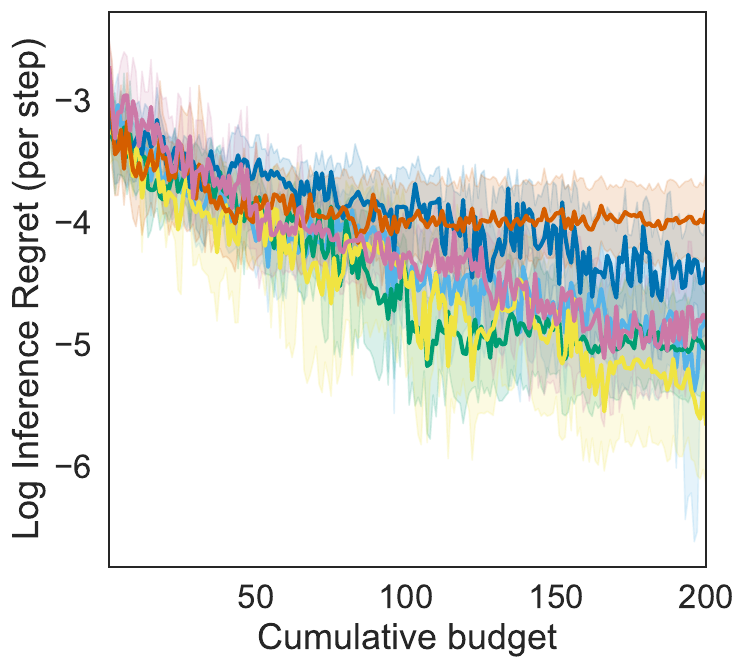}
        \caption{HPO-MF-Cont}
    \end{subfigure}
    \hfill
    \begin{subfigure}[t]{0.32\textwidth}
        \centering
        \includegraphics[width=\textwidth]{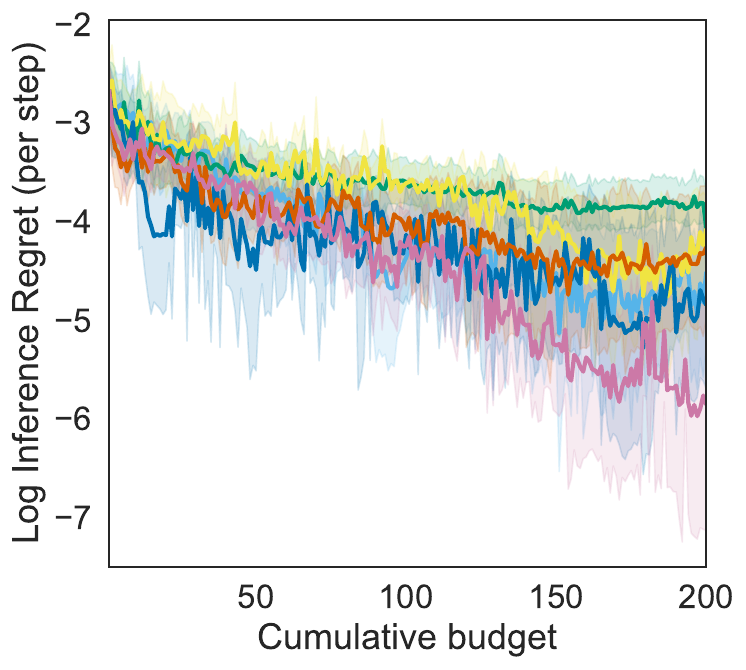}
        \caption{HPO-MF-Disc}
    \end{subfigure}

    % Row 3: 3 subfigures, second spans remaining width
    \begin{subfigure}[t]{0.32\textwidth}
        \centering
        \includegraphics[width=\textwidth]{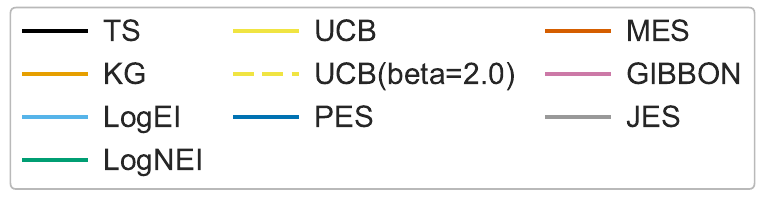}
    \end{subfigure}
    \hfill
    \begin{subfigure}[t]{0.32\textwidth}
        \centering
        \includegraphics[width=\textwidth]{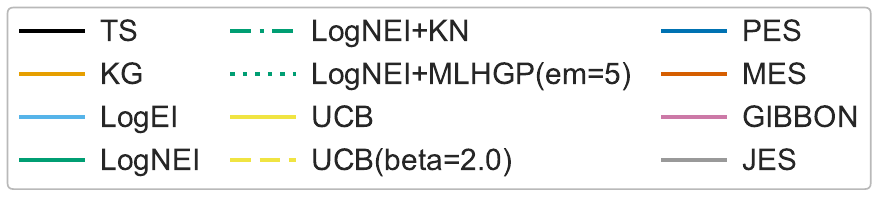}
    \end{subfigure}
    
    % Row 4: three subfigures
    \begin{subfigure}[t]{0.32\textwidth}
        \centering
        \includegraphics[width=\textwidth]{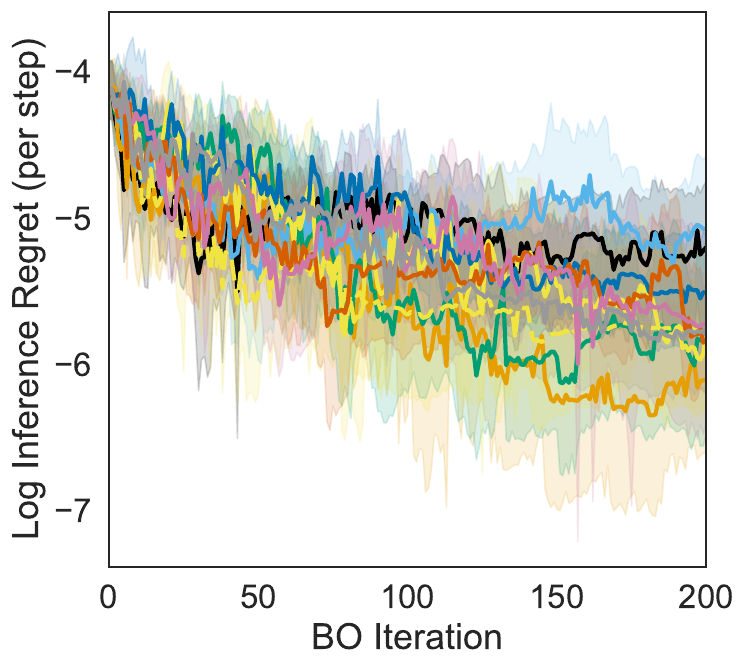}
        \caption{DMO}
    \end{subfigure}
    \hfill
    \begin{subfigure}[t]{0.32\textwidth}
        \centering
        \includegraphics[width=\textwidth]{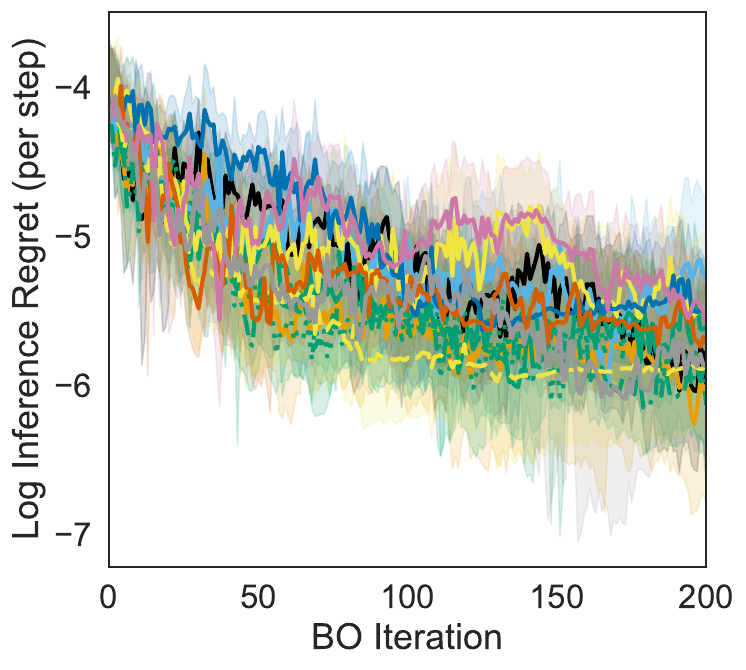}
        \caption{DMO-Het}
    \end{subfigure}

    % Row 5: 1 subfigures, second spans remaining width
    \begin{subfigure}[t]{\textwidth}
        \centering
        \includegraphics[width=\textwidth]{pics/po/po128_long_legend.pdf}
    \end{subfigure}
    
    % Row 6: 4 subfigures
    \begin{subfigure}[t]{0.24\textwidth}
        \centering
        \includegraphics[width=\textwidth]{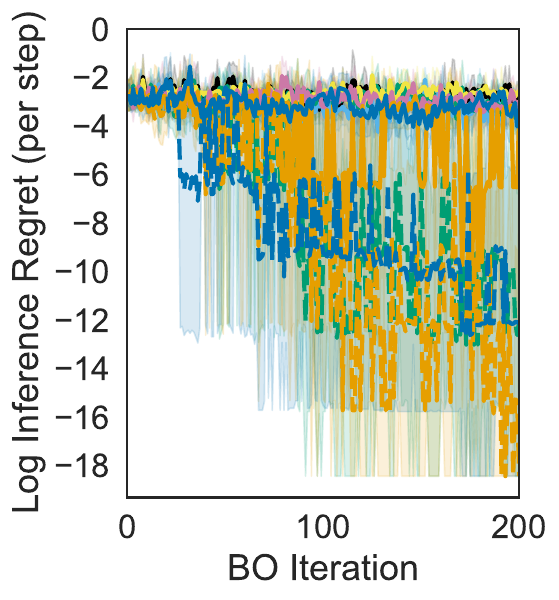}
        \caption{PO-128}
    \end{subfigure}
    \hfill
    \begin{subfigure}[t]{0.24\textwidth}
        \centering
        \includegraphics[width=\textwidth]{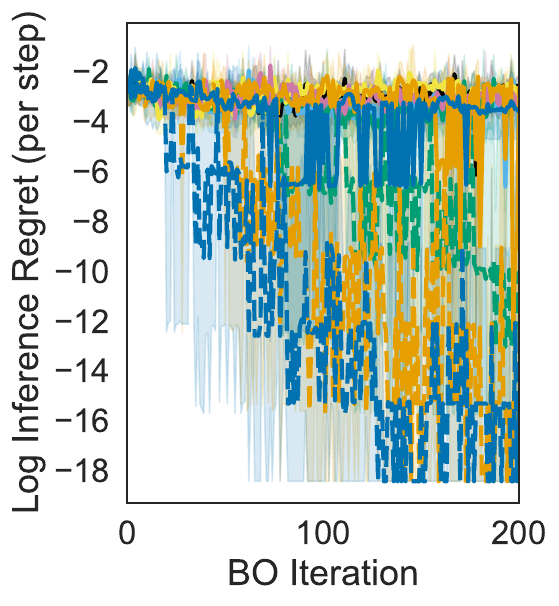}
        \caption{PO-256}
    \end{subfigure}
    \hfill
    \begin{subfigure}[t]{0.24\textwidth}
        \centering
        \includegraphics[width=\textwidth]{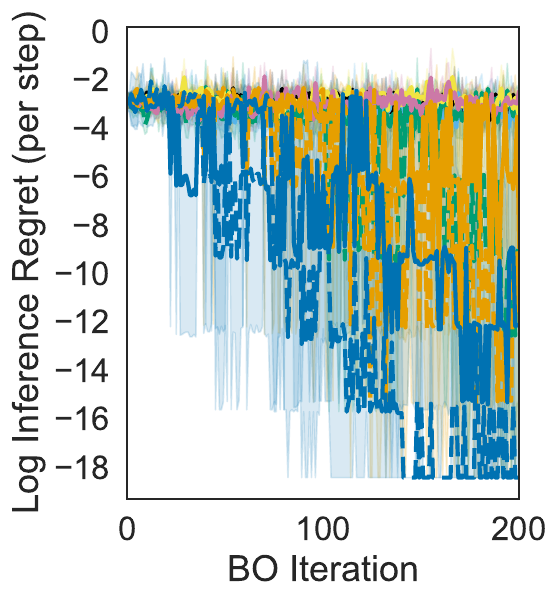}
        \caption{PO-512}
    \end{subfigure}
    \hfill
        \begin{subfigure}[t]{0.24\textwidth}
        \centering
        \includegraphics[width=\textwidth]{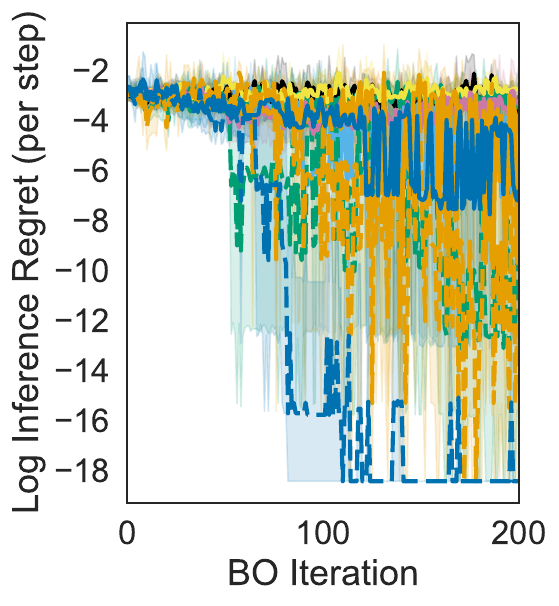}
        \caption{PO-768}
    \end{subfigure}

    % Row 7: 1 subfigures, second spans remaining width
    \begin{subfigure}[t]{\textwidth}
        \centering
        \includegraphics[width=\textwidth]{pics/pco/pco_long_legend.pdf}
    \end{subfigure}
    
    % Row 8: 3 subfigures
    \begin{subfigure}[t]{0.32\textwidth}
        \centering
        \includegraphics[width=\textwidth]{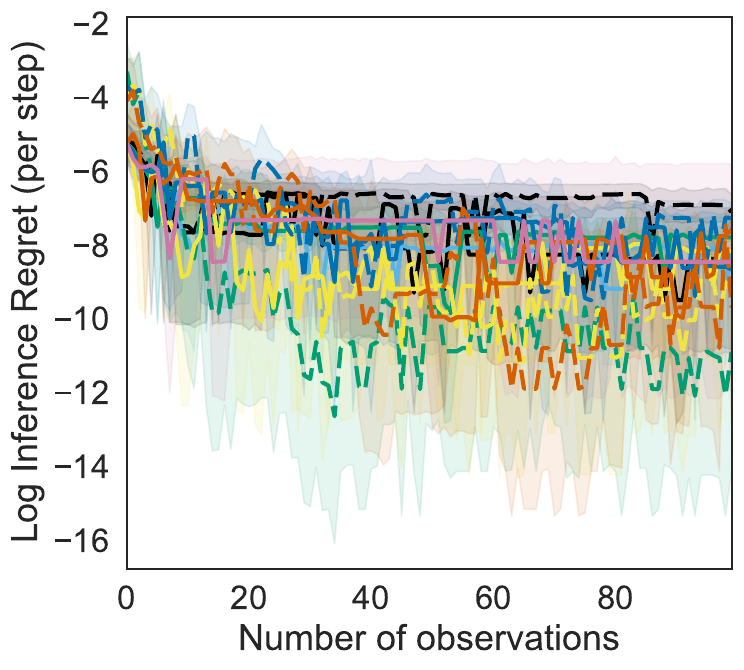}
        \caption{PCO-16}
    \end{subfigure}
    \hfill
    \begin{subfigure}[t]{0.32\textwidth}
        \centering
        \includegraphics[width=\textwidth]{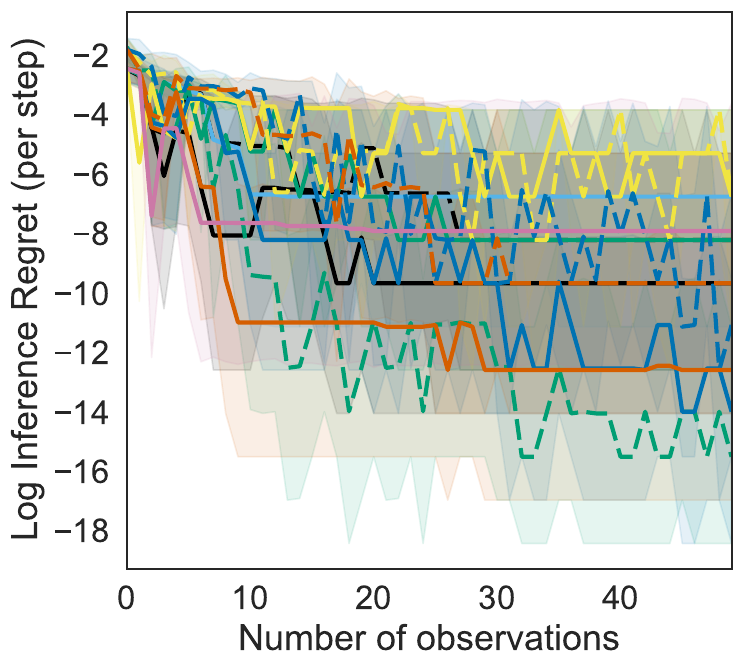}
        \caption{PCO-32}
    \end{subfigure}
    \hfill
    \begin{subfigure}[t]{0.32\textwidth}
        \centering
        \includegraphics[width=\textwidth]{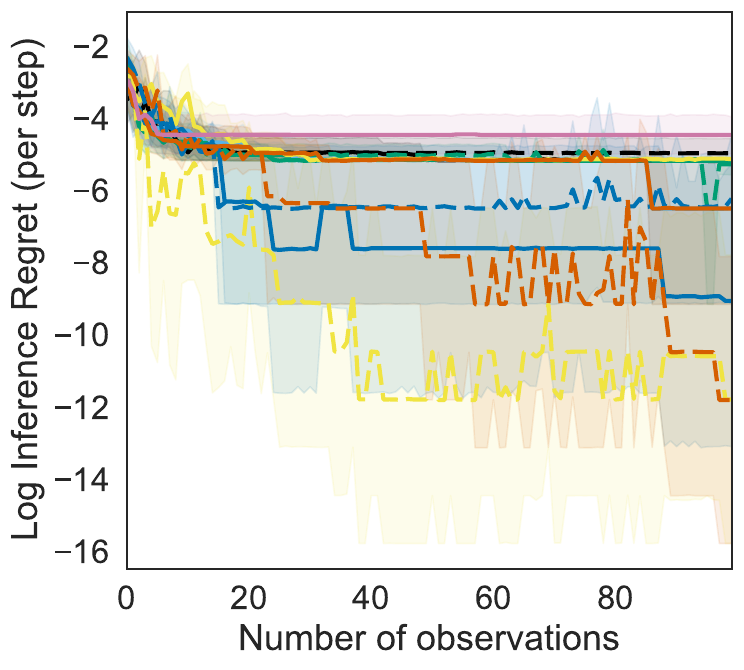}
        \caption{PCO-64}
    \end{subfigure}
    \caption{Results on \bolt's problems using inference regret. Random is not included as random sampling does not maintain a posterior mean.}
    \label{fig:inf_ref}
\end{figure}

\begin{figure}
    \centering
    
    % Row 1: two subfigures, second spans remaining width
    \begin{subfigure}[t]{0.32\textwidth}
        \centering
        \includegraphics[width=\textwidth]{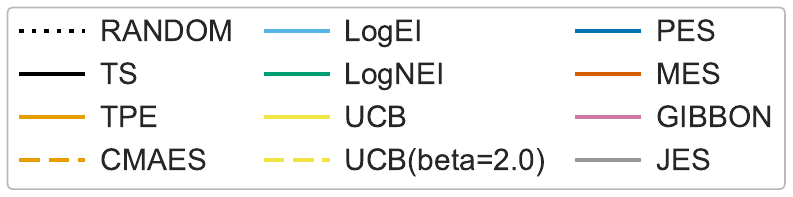}
    \end{subfigure}
    \hfill
    \begin{subfigure}[t]{0.32\textwidth}
        \centering
        \includegraphics[width=\textwidth]{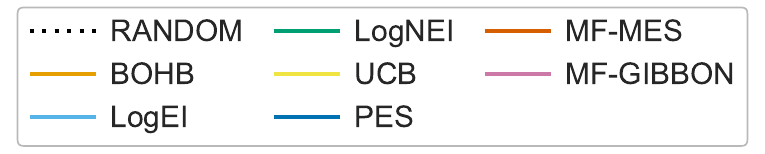}
    \end{subfigure}
    \hfill
    \begin{subfigure}[t]{0.32\textwidth}
        \centering
        \includegraphics[width=\textwidth]{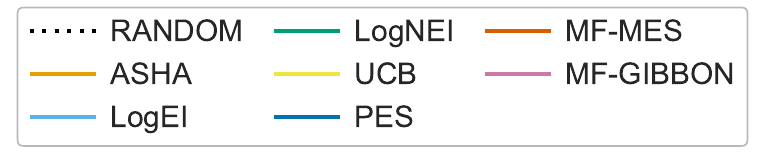}
    \end{subfigure}
    
    % Row 2: three subfigures
    \begin{subfigure}[t]{0.32\textwidth}
        \centering
        \includegraphics[width=\textwidth]{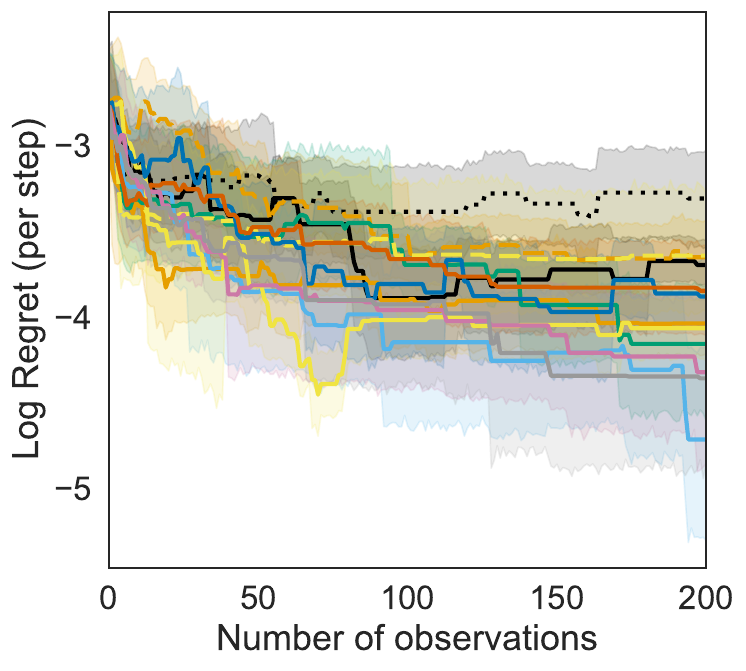}
        \caption{HPO}
    \end{subfigure}
    \hfill
    \begin{subfigure}[t]{0.32\textwidth}
        \centering
        \includegraphics[width=\textwidth]{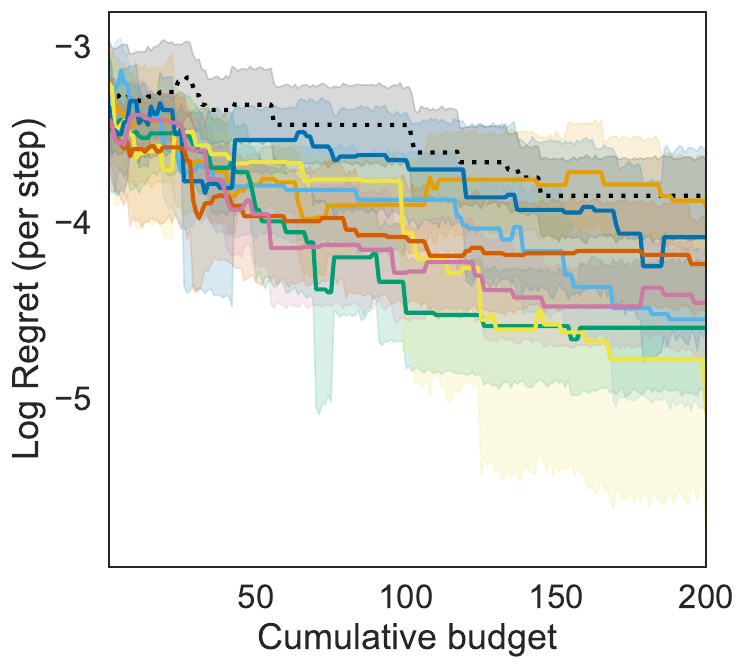}
        \caption{HPO-MF-Cont}
    \end{subfigure}
    \hfill
    \begin{subfigure}[t]{0.32\textwidth}
        \centering
        \includegraphics[width=\textwidth]{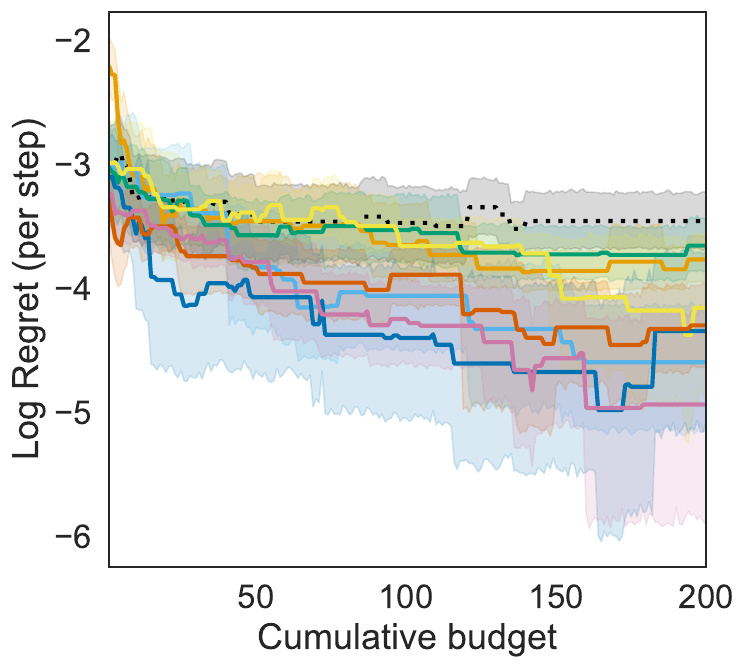}
        \caption{HPO-MF-Disc}
    \end{subfigure}

    % Row 3: 3 subfigures, second spans remaining width
    \begin{subfigure}[t]{0.32\textwidth}
        \centering
        \includegraphics[width=\textwidth]{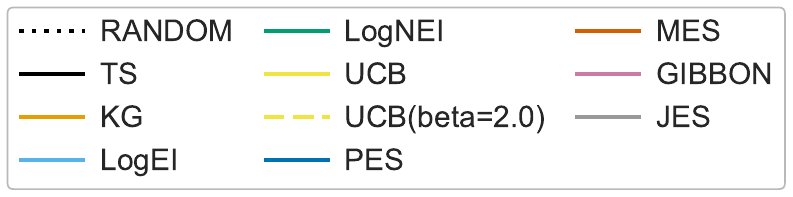}
    \end{subfigure}
    \hfill
    \begin{subfigure}[t]{0.32\textwidth}
        \centering
        \includegraphics[width=\textwidth]{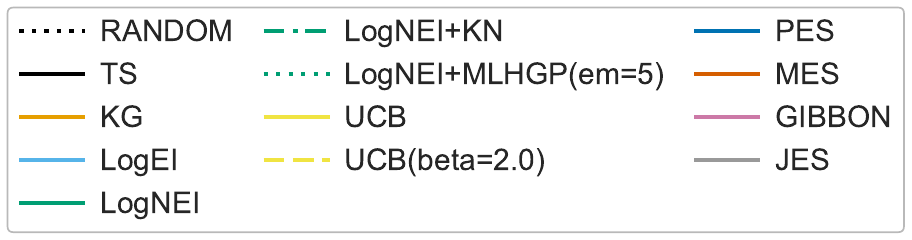}
    \end{subfigure}

    % Row 4: three subfigures
    \begin{subfigure}[t]{0.32\textwidth}
        \centering
        \includegraphics[width=\textwidth]{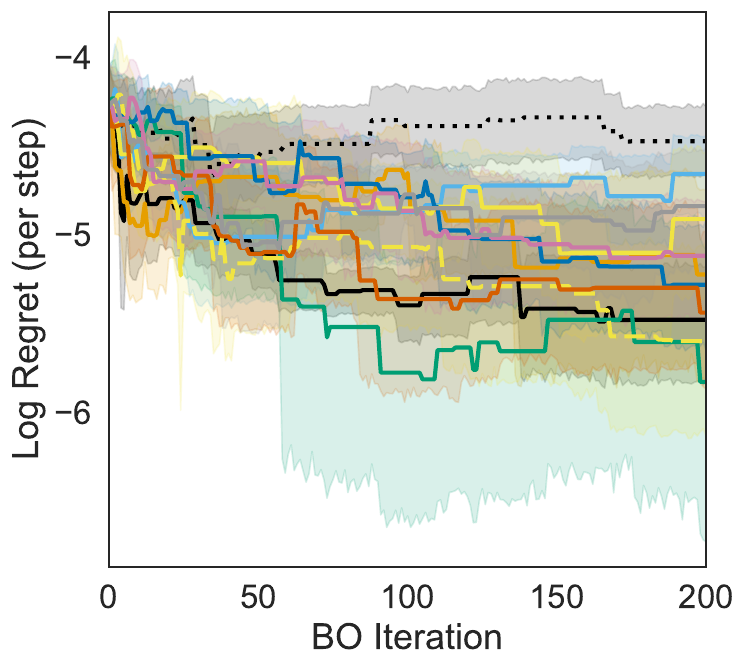}
        \caption{DMO}
    \end{subfigure}
    \hfill
    \begin{subfigure}[t]{0.32\textwidth}
        \centering
        \includegraphics[width=\textwidth]{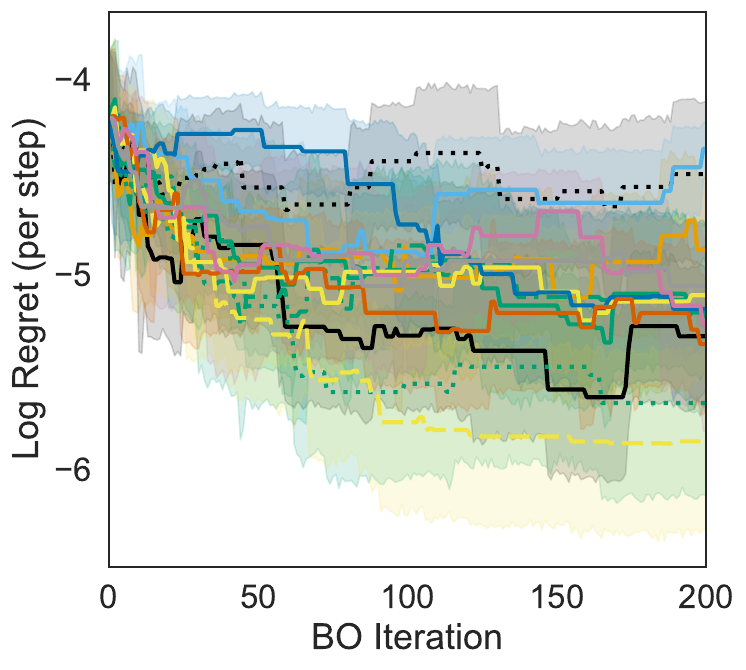}
        \caption{DMO-Het}
    \end{subfigure}

    % Row 5: 1 subfigures, second spans remaining width
    \begin{subfigure}[t]{\textwidth}
        \centering
        \includegraphics[width=\textwidth]{pics/po/po128_long_legend.pdf}
    \end{subfigure}% \vspace{-3pt}
    
    % Row 6: 4 subfigures
    \begin{subfigure}[t]{0.24\textwidth}
        \centering
        \includegraphics[width=\textwidth]{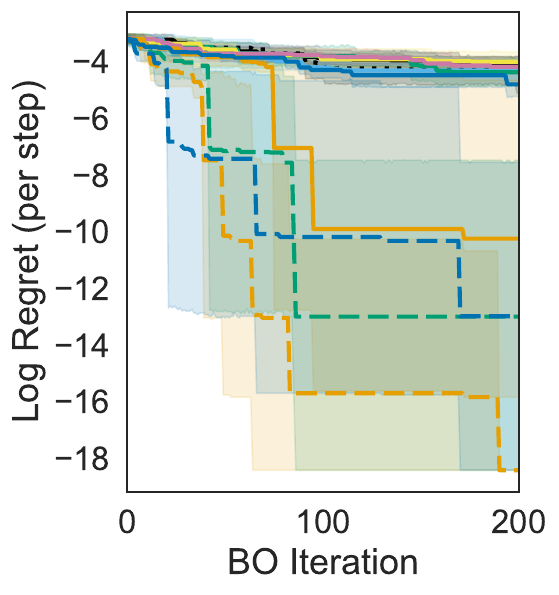}
        \caption{PO-128}
    \end{subfigure}
    \hfill
    \begin{subfigure}[t]{0.24\textwidth}
        \centering
        \includegraphics[width=\textwidth]{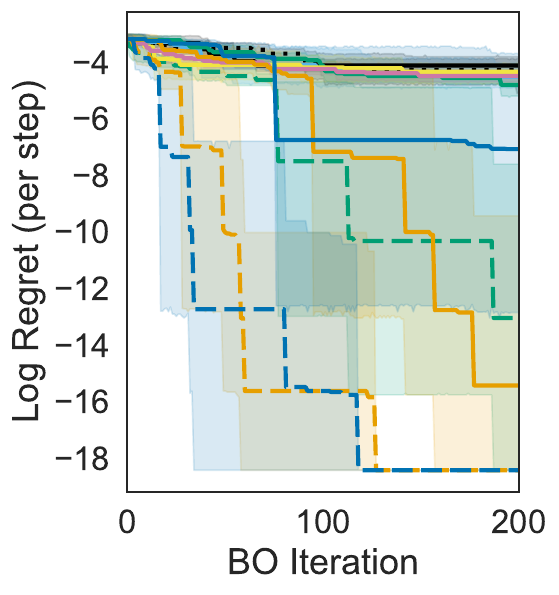}
        \caption{PO-256}
    \end{subfigure}
    \hfill
    \begin{subfigure}[t]{0.24\textwidth}
        \centering
        \includegraphics[width=\textwidth]{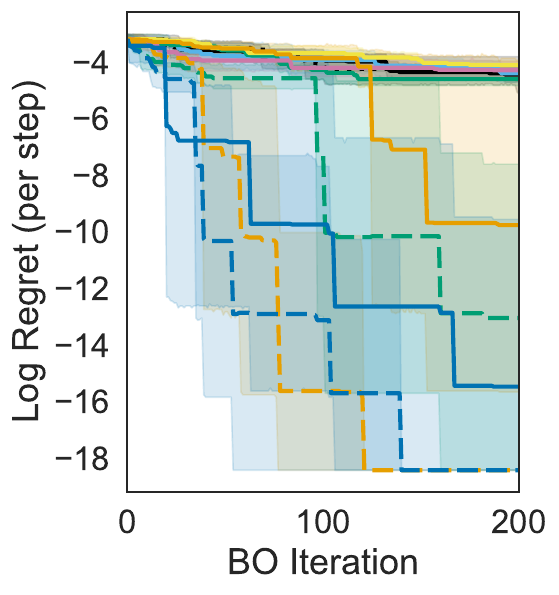}
        \caption{PO-512}
    \end{subfigure}
    \hfill
        \begin{subfigure}[t]{0.24\textwidth}
        \centering
        \includegraphics[width=\textwidth]{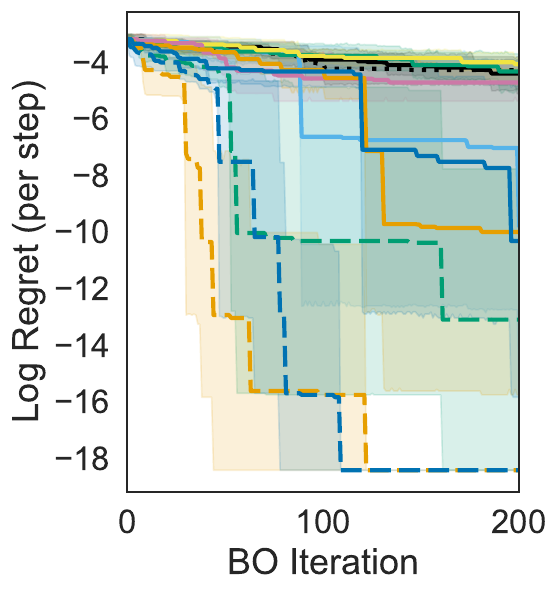}
        \caption{PO-768}
    \end{subfigure}

    % Row 7: 1 subfigures, second spans remaining width
    \begin{subfigure}[t]{\textwidth}
        \centering
        \includegraphics[width=\textwidth]{pics/pco/pco_long_legend.pdf}
    \end{subfigure}
    
    % Row 8: 3 subfigures
    \begin{subfigure}[t]{0.32\textwidth}
        \centering
        \includegraphics[width=\textwidth]{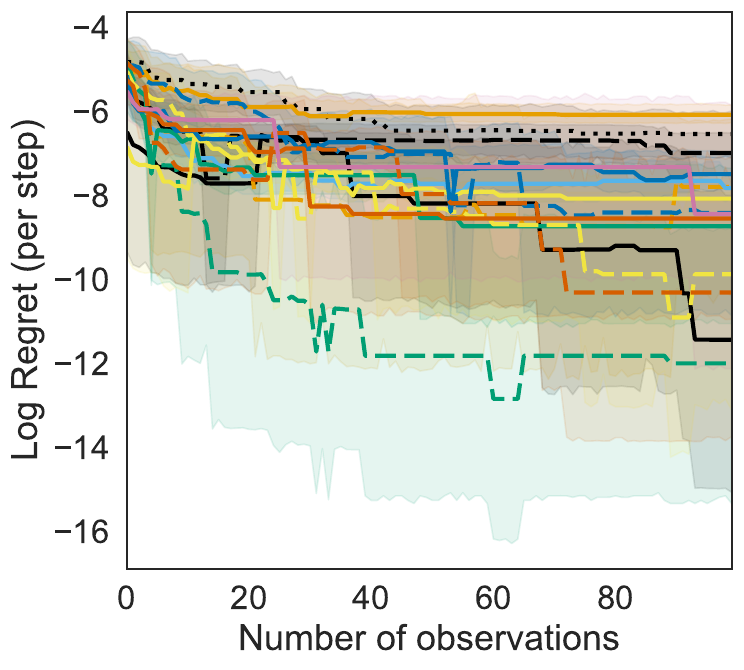}
        \caption{PCO-16}
    \end{subfigure}
    \hfill
    \begin{subfigure}[t]{0.32\textwidth}
        \centering
        \includegraphics[width=\textwidth]{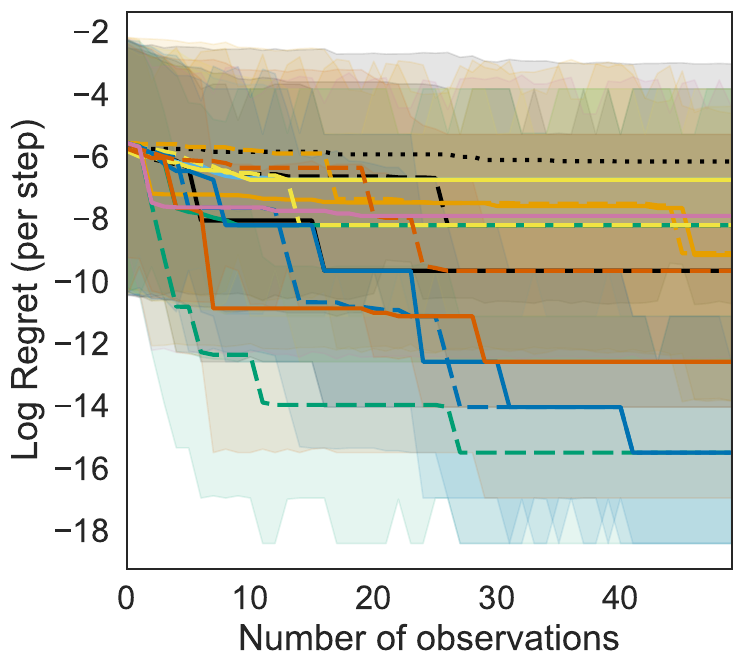}
        \caption{PCO-32}
    \end{subfigure}
    \hfill
    \begin{subfigure}[t]{0.32\textwidth}
        \centering
        \includegraphics[width=\textwidth]{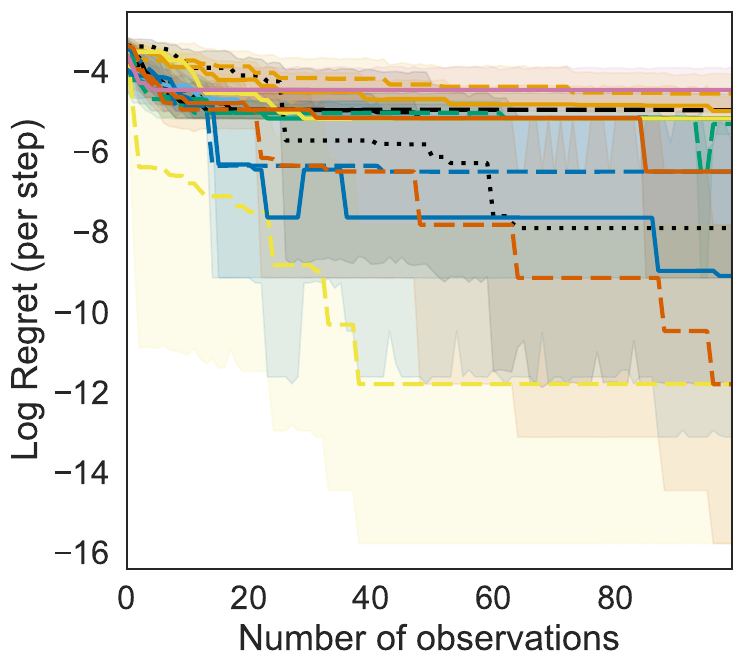}
        \caption{PCO-64}
    \end{subfigure}
    \caption{Results on \bolt's problems using per-step simple regret.}
    \label{fig:per_simple_ref}
\end{figure}

Inference regret plots are \cref{fig:inf_ref} and per-step simple regret plots are in \cref{fig:per_simple_ref}. Per-step simple regret closely follows simple regret across all problems, as observation noise is small relative to differences in objective values.  Under inference regret, methods are less distinguishable on some problems like HPO, suggesting that their posterior means are similar despite different observations. Nevertheless, the main insights from \cref{sec:exps} largely hold under both metrics: MF-GIBBON leads on HPO-MF-Disc, noise-aware LogNEI variants and UCB($\beta=2$) are among the strongest on DMO-Het. Batched, dTuRBO, and dBAxUS methods remain essential for high-dimensional PO, and cKG remains the strongest on PCO-64. One main difference is on PCO-32, where cKG+prior leads under simple regret but not under inference regret, where ADMMBO(+prior) and LogNEIC+prior perform best.

\subsection{Wall-clock time for experiments}
Wall-clock runtimes are reported in \cref{tab:wall_clock_hpo_dmo,tab:po_clock,tab:wall_clock_pco}. These runtimes include emulator query time, which is negligible. Each entry is the total run time divided by the number of iterations ($50$ for \texttt{PCO-32}, $100$ for \texttt{PCO-16}, \texttt{PCO-64}, $200$ for the rest). Per-step cost grows super-linearly with the number of observations, as the surrogate is refitted on more data, so later steps take longer than this average suggests and earlier steps shorter. PCO experiments are ran on CPUs. Other experiments ran on L40 GPUs, with a small proportion using H100 or H200 GPUs to accommodate the higher VRAM demands of the GPU-accelerated implementation, which scales with the number of observations and input dimensionality (e.g., PO-768 has $d=768$ which is memory intensive).

\begin{table}[t]
  \centering
  \caption{Mean wall-clock time per BO step --- HPO and DMO. All ran on L40s.}
  \label{tab:wall_clock_hpo_dmo}
  \begin{subtable}[t]{0.48\textwidth}
    \centering
    \footnotesize
    \caption{\textbf{HPO}}
    \begin{tabular}{lr}
      \toprule
      \textbf{Method} & \textbf{Time\,/\,step} \\
      \midrule
      \multicolumn{2}{l}{\textit{HPO}} \\
      \cmidrule{1-2}
      RANDOM & 0.00184\,s \\
      TS & 7.47\,s \\
      TPE & 0.00987\,s \\
      CMAES & 0.00299\,s \\
      LogEI & 5.73\,s \\
      LogNEI & 6.07\,s \\
      UCB & 7.28\,s \\
      UCB(beta=2.0) & 4.33\,s \\
      PES & 13\,s \\
      MES & 7.1\,s \\
      GIBBON & 7.29\,s \\
      JES & 8.48\,s \\
      \midrule
      \multicolumn{2}{l}{\textit{HPO-MF-Cont}} \\
      \cmidrule{1-2}
      RANDOM & 0.00608\,s \\
      BOHB & 0.151\,s \\
      LogEI & 9.94\,s \\
      LogNEI & 11.7\,s \\
      UCB & 27.6\,s \\
      PES & 18.9\,s \\
      MF-MES & 9.72\,s \\
      MF-GIBBON & 8.99\,s \\
      \midrule
      \multicolumn{2}{l}{\textit{HPO-MF-Disc}} \\
      \cmidrule{1-2}
      RANDOM & 0.0039\,s \\
      ASHA & 0.0111\,s \\
      LogEI & 6.36\,s \\
      LogNEI & 6.48\,s \\
      UCB & 9.18\,s \\
      PES & 16\,s \\
      MF-MES & 7.69\,s \\
      MF-GIBBON & 7.27\,s \\
      \bottomrule
    \end{tabular}
  \end{subtable}
  \hfill
  \begin{subtable}[t]{0.48\textwidth}
    \centering
    \footnotesize
    \caption{\textbf{DMO}}
    \begin{tabular}{lr}
      \toprule
      \textbf{Method} & \textbf{Time\,/\,step} \\
      \midrule
      \multicolumn{2}{l}{\textit{DMO}} \\
      \cmidrule{1-2}
      RANDOM & 0.00561\,s \\
      RANDOM(q=5) & 0.00596\,s \\
      RANDOM(q=10) & 0.00542\,s \\
      TS & 1.9\,s \\
      KG & 21.6\,s \\
      LogEI & 3.17\,s \\
      LogNEI & 4.56\,s \\
      LogNEI(q=5) & 19.2\,s \\
      LogNEI(q=10) & 1.79\,min \\
      UCB & 3.02\,s \\
      UCB(beta=2.0) & 3.01\,s \\
      PES & 4.66\,s \\
      MES & 4.11\,s \\
      MES(q=5) & 9.44\,s \\
      MES(q=10) & 16.1\,s \\
      GIBBON & 3.49\,s \\
      GIBBON(q=5) & 18.2\,s \\
      GIBBON(q=10) & 49.8\,s \\
      JES & 3.75\,s \\
      \midrule
      \multicolumn{2}{l}{\textit{DMO-MO}} \\
      \cmidrule{1-2}
      RANDOM & 0.0202\,s \\
      RANDOM(q=5) & 0.0422\,s \\
      TSEMO & 2.02\,s \\
      NSGA-II(pop\_size=50) & 1.2\,s \\
      ParEGO & 3.15\,s \\
      ParEGO(q=5) & 14.7\,s \\
      NEHVI & 11.2\,s \\
      NEHVI(q=5) & 4.82\,min \\
      PES\_MO & 31\,s \\
      MES\_MO & 4.68\,s \\
      JES\_MO & 6.24\,s \\
      MES\_MO(q=5) & 12.8\,s \\
      \midrule
      \multicolumn{2}{l}{\textit{DMO-Het}} \\
      \cmidrule{1-2}
      RANDOM & 0.00596\,s \\
      TS & 1.79\,s \\
      KG & 21.9\,s \\
      LogEI & 3.26\,s \\
      LogNEI & 4.25\,s \\
      LogNEI+KN & 4.28\,s \\
      LogNEI+MLHGP(em=5) & 5.46\,s \\
      UCB & 2.96\,s \\
      UCB(beta=2.0) & 3\,s \\
      PES & 4.6\,s \\
      MES & 3.72\,s \\
      GIBBON & 3.52\,s \\
      JES & 3.73\,s \\
      \bottomrule
    \end{tabular}
  \end{subtable}
\end{table}

\begin{table}[t]
  \centering
  \footnotesize
  \caption{Mean wall-clock time per BO step --- PO.}
  \label{tab:po_clock}
  \begin{tabular}{lrrrr}
    \toprule
    \textbf{Method} & \textbf{\textit{PO-128}} & \textbf{\textit{PO-256}} & \textbf{\textit{PO-512}} & \textbf{\textit{PO-768}} \\
    \midrule
    RANDOM & 0.00746\,s & 0.00966\,s & 0.0146\,s & 0.0112\,s \\
    TS & 8.19\,s & 9.04\,s & 8.76\,s$^*$ & 4.98\,s$^\dagger$ \\
    LogEI & 10.3\,s & 12\,s & 6.82\,s$^*$ & 1.76\,s$^\dagger$ \\
    LogNEI & 8.55\,s & 10.6\,s & 7.15\,s$^*$ & 6.54\,s$^\dagger$ \\
    LogNEI(q=5) & 30.3\,s & 59.4\,s$^*$ & 1.01\,min$^*$ & 40.7\,s$^\dagger$ \\
    UCB & 7.65\,s & 9.34\,s & 7.22\,s$^*$ & 4.65\,s$^\dagger$ \\
    GIBBON & 9.57\,s & 9.94\,s & 7.34\,s$^*$ & 4.89\,s$^\dagger$ \\
    dTuRBO+LogNEI & 9.8\,s & 10.3\,s & 8.85\,s$^*$ & 3.84\,s $^\dagger$\\
    dTuRBO+LogNEI(q=5) & 14.9\,s & 16.1\,s & 13.5\,s$^*$ & 7.4\,s$^\dagger$ \\
    dBAxUS+LogNEI & 7.77\,s & 7.04\,s$^*$ & 7.13\,s$^*$ & 4.55\,s$^\dagger$ \\
    dBAxUS+LogNEI(q=5) & 8.15\,s & 7.21\,s$^*$ & 9.79\,s$^*$ & 7.68\,s$^\dagger$ \\
    \addlinespace
    \multicolumn{5}{l}{\footnotesize $^*$Ran on H100. $^\dagger$Ran on H200. Otherwise ran on L40} \\
    \bottomrule
  \end{tabular}
\end{table}

\begin{table}[h]
  \centering
  \footnotesize
  \caption{Mean wall-clock time per BO step --- PCO. All ran on CPU.}
  \label{tab:wall_clock_pco}
  \begin{tabular}{lrrr}
    \toprule
    \textbf{Method} & \textbf{\textit{PCO-16}} & \textbf{\textit{PCO-32}} & \textbf{\textit{PCO-64}} \\
    \midrule
    RANDOM & 0.00135\,s & 0.00117\,s & 0.00119\,s \\
    cTS & 3.17\,s & 1.49\,s & 3.23\,s \\
    dSCBO & 1.6\,s & 0.801\,s & 0.88\,s \\
    TPE & 0.016\,s & 0.0108\,s & 0.0112\,s \\
    CMAES & 0.00296\,s & 0.00252\,s & 0.00256\,s \\
    LogEIC & 2.7\,s & 1.77\,s & 3.07\,s \\
    LogNEIC & 2.73\,s & 2.14\,s & 3.25\,s \\
    LogNEIC+prior & 7.79\,s & 7.71\,s & 7.66\,s \\
    UCB-C & 3.6\,s & 2.57\,s & 2.68\,s \\
    UCB-C+prior & 7.68\,s & 6.84\,s & 5.73\,s \\
    cKG & 4.28\,s & 2.52\,s & 3.42\,s \\
    cKG+prior & 8.19\,s & 6.92\,s & 8.46\,s \\
    ADMMBO & 3.13\,s & 1.95\,s & 2.81\,s \\
    ADMMBO+prior & 7.94\,s & 7.09\,s & 7.55\,s \\
    CMES-IBO & 2.94\,s & 1.19\,s & 1.89\,s \\
    \bottomrule
  \end{tabular}
\end{table}

\section{Compute usage}
\label{ap:gpu}

\begin{table}[ht]
\centering
\caption{Compute Usage Summary}
\label{tab:compute}
\begin{tabular}{llrr}
\toprule
\textbf{Stage} & \textbf{Model} & \textbf{Training} & \textbf{Eval } \\
\midrule
\multirow{2}{*}{HPO} 
  & 8B & $ 31$ H200 GPU-days & $ 17$ L40 GPU-days\\
  & 4B & $ 62$ H200 GPU-days & $ 7$ L40 GPU-days\\
\midrule
DMO & 4B & $ 123$ H200 GPU-days & $ 107$ L40 GPU-days \\
\midrule
PO  & 14B  & -- & $21$ H200 GPU-days  \\
\midrule
PCO-16  &  & -- & $57$ H100 GPU-days  \\
PCO-32  &  & -- & $154$ GH200 GPU-days  \\
PCO-64  &  & -- & $277$ GH200 GPU-days  \\
\bottomrule
\end{tabular}
\end{table}

\begin{table}[ht]
\centering
\caption{Compute for real BBO runs in \cref{fig:HPO_real_emu,fig:DMO_real_emu}.}
\label{tab:compute_bbo}
\begin{tabular}{llrr}
\toprule
\textbf{Problem} & \textbf{Training} & \textbf{Eval } \\
\midrule
HPO & $ 53$ H200 GPU-days & $ 15$ L40 GPU-hours\\
DMO & $ 20$ H200 GPU-days & $ 32$ L40 GPU-days \\
\bottomrule
\end{tabular}
\end{table}

We detail the compute used for generation of \bolt's data, and BO experiments in \cref{tab:compute}. Additionally, in \cref{tab:compute_bbo}, we report the compute cost for the BBO evaluations with real LLMs, used in \cref{fig:HPO_real_emu,fig:DMO_real_emu}, to highlight how expensive evaluations are without our emulators and datasets.

\end{document}

%% file: math_commands.tex
\usepackage{amsmath,amsfonts,bm}

\def\eqref#1{equation~\ref{#1}}
\def\1{\bm{1}}

\DeclareMathAlphabet{\mathsfit}{\encodingdefault}{\sfdefault}{m}{sl}
\SetMathAlphabet{\mathsfit}{bold}{\encodingdefault}{\sfdefault}{bx}{n}